\documentclass[review,10pt]{elsarticle}

\usepackage{amssymb}

\usepackage{hyperref}
\usepackage{enumitem}
\usepackage{url}
\usepackage{adjustbox}
\usepackage{multirow}
\usepackage{booktabs}
\usepackage{makecell}
\usepackage{mathtools}
\usepackage{subcaption}
\usepackage{tabularray}
\usepackage{xcolor}

\colorlet{review}{black}

\colorlet{review2}{black}

\usepackage{bm}
\usepackage{amssymb}
\usepackage{amsmath}

\def\p{{\mathrm{p}}}
\def\q{{\mathrm{q}}}
\def\dd{{\mathrm{d}}}

\newcommand*{\softmax}{\operatorname{Softmax}}

\newcommand*{\diag}[1]{\operatorname{Diag} \left( #1 \right)}

\newcommand{\direnergy}[1]{\cE_{D}\left( #1 \right)}

\def\Rbb{{\mathbb{R}}}
\def\Nbb{{\mathbb{N}}}
\def\Ebb{{\mathbb{E}}}

\def\cA{{\mathcal A}}

\def\cE{{\mathcal E}}
\def\cN{{\mathcal N}}

\def\cX{{\mathcal X}}
\def\cY{{\mathcal Y}}
\def\cF{{\mathcal F}}
\def\cf{{\mathcal F}}

\def\ba{{\mathbf a}}

\def\bff{{\mathbf f}}

\def\bh{{\mathbf h}}

\def\bw{{\mathbf w}}
\def\bx{{\mathbf x}}

\def\bz{{\mathbf z}}

\def\bA{{\mathbf A}}

\def\bD{{\mathbf D}}

\def\bH{{\mathbf H}}

\def\bL{{\mathbf L}}

\def\bW{{\mathbf W}}
\def\bX{{\mathbf X}}
\def\bY{{\mathbf Y}}
\def\bZ{{\mathbf Z}}

\newcommand*{\probsmoothatt}{ProbSA\@}
\newcommand{\probsmooth}{\probsmoothatt\@}
\newcommand*{\smoothatt}{SA\@}

\newcommand*{\abmil}{ABMIL\@}
\newcommand*{\clam}{CLAM\@}
\newcommand*{\dsmil}{DSMIL\@}
\newcommand*{\transmil}{TransMIL\@}
\newcommand*{\dtfdmil}{DTFD-MIL\@}
\newcommand*{\gtp}{GTP\@}
\newcommand*{\setmil}{SETMIL\@}
\newcommand*{\camil}{CAMIL\@}
\newcommand*{\iibmil}{IIBMIL\@}
\newcommand*{\pathgcn}{PatchGCN\@}

\newcommand*{\vmil}{VMIL\@}

\newcommand{\transformerabmil}{T-ABMIL\@}

\newcommand{\elbo}{\operatorname{ELBO}}
\newcommand{\cyclical}{\operatorname{cyclical}}

\DeclarePairedDelimiter\floor{\lfloor}{\rfloor}

\journal{Pattern Recognition}

\makeatletter
\def\ps@pprintTitle{%
	\let\@oddhead\@empty
	\let\@evenhead\@empty
	\def\@oddfoot{}%
	\let\@evenfoot\@oddfoot
}
\makeatother

\begin{document}

\begin{frontmatter}



\title{Probabilistic smooth attention for deep multiple instance learning in medical imaging}


\author[inst1,inst3]{Francisco M. Castro-Macías}
\author[inst2,inst3]{Pablo Morales-Álvarez}
\author[inst4,inst5]{Yunan Wu}
\author[inst1,inst3]{Rafael Molina}
\author[inst4,inst5]{Aggelos K. Katsaggelos}

\affiliation[inst1]{organization={Department of Computer Science and AI, University of Granada}, 
            country={Spain}}

\affiliation[inst2]{organization={Department of Statistics and Operations Research, University of Granada}, 
            country={Spain}}

\affiliation[inst3]{organization={Research Centre for Information and Communication Technologies (CITIC), University of Granada}, 
            country={Spain}}

\affiliation[inst4]{organization={Department of Electrical and Computer Engineering, Northwestern University}, 
            country={USA}}

\affiliation[inst5]{organization={Center for Computational Imaging and Signal Analytics in Medicine, Northwestern University}, 
            country={USA}}

\begin{abstract}
The Multiple Instance Learning (MIL) paradigm is attracting plenty of attention in medical imaging classification, where labeled data is scarce. 
MIL methods cast medical images as bags of instances (e.g. patches in whole slide images, or slices in CT scans), and only bag labels are required for training.  
Deep MIL approaches have obtained promising results by aggregating instance-level representations via an attention mechanism to compute the bag-level prediction.
{\color{review}
These methods typically capture both local interactions among adjacent instances and global, long-range dependencies through various mechanisms. However, they treat attention values deterministically, potentially overlooking uncertainty in the contribution of individual instances.
}
In this work we propose a novel probabilistic framework that estimates a probability distribution over the attention values, and accounts for both global and local interactions. 
In a comprehensive evaluation involving {\color{review} eleven} state-of-the-art baselines and three medical datasets, we show that our approach achieves top predictive performance in different metrics. Moreover, the probabilistic treatment of the attention provides uncertainty maps that are interpretable in terms of illness localization. 
\end{abstract}

\begin{keyword}
Multiple instance learning
\sep
probabilistic machine learning
\sep
Bayesian methods
\sep
Whole Slide Images
\sep
CT scans
\end{keyword}

\end{frontmatter}



\section{Introduction}
\label{sec:introduction}

\begin{figure}[!t]
    \centering
    \begin{subfigure}[b]{1.0\textwidth}
        \centering
        \includegraphics[trim={1.0cm 0.5cm 0cm 0cm},clip,width=0.4\linewidth]{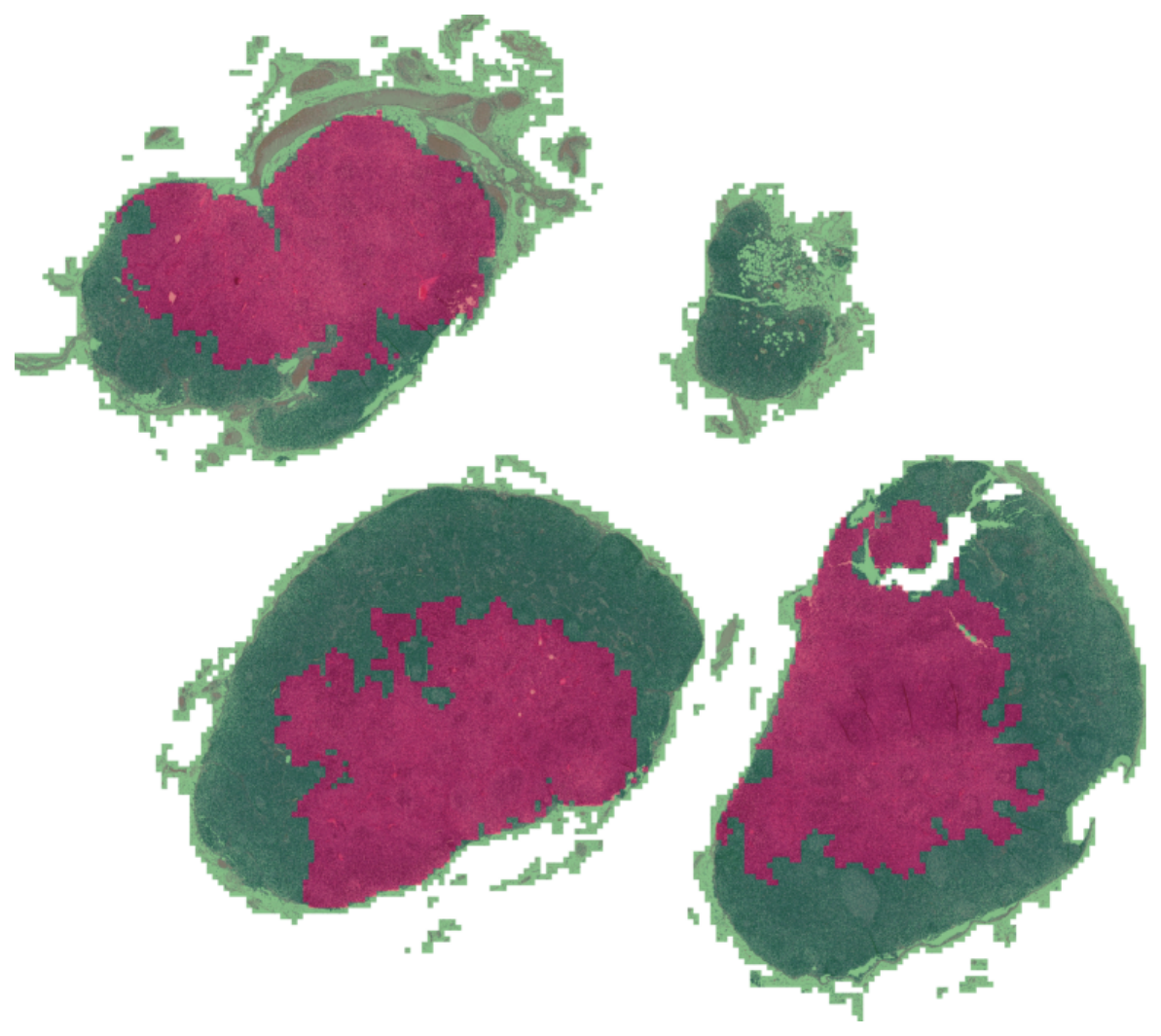}
        \caption{Labeled patches in a WSI.}
        \label{fig:wsi_labels}
    \end{subfigure}

    \begin{subfigure}[b]{1.0\textwidth}
        \centering
        \includegraphics[width=0.98\linewidth]{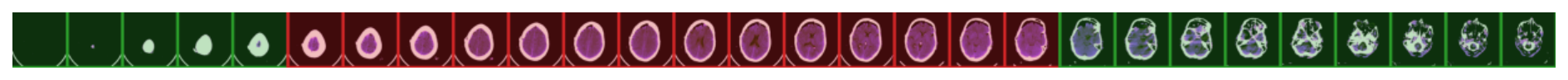}
        \caption{Labeled slices in a CT scan.}
        \label{fig:ctscan_labels}
    \end{subfigure}
    \caption{
         Two modalities of MIL medical images: 
         (a) a whole slide image (WSI) por tumour detection and (b) a CT scan for hemorrhage detection. 
         The red color indicates malignant/hemorrhage patches/slices, respectively. During training, instance labels are not known, and only a global bag-level label is available.
         Note that these labels show spatial dependencies: a patch/slice is usually surrounded by patches/slices with the same label. 
    }
    \vspace{-0.3cm}
    \label{fig:instance_labels}
\end{figure}

Multiple Instance Learning (MIL) is a popular machine learning framework in which the model is trained on sets of instances, called bags, rather than individual instances \citep{dietterich1997solving}. 
Unlike the traditional supervised learning framework, which needs the label of each instance, in MIL only the label of the bag is required. 
This approach is particularly beneficial in scenarios where labelling many individual instances is impractical or prohibitively expensive.
This is especially relevant in the context of medical imaging, where annotations are extremely costly \citep{song2023artificial}. 
\autoref{fig:instance_labels} shows two illustrative examples of MIL problems within this context: (i) cancer detection from Whole Slide Images (WSIs), where the WSI represents the bag, and the patches are the instances; and (ii) intracranial hemorrhage detection from Computerized Tomographic (CT) scans, where the full scan represents the bag and the slices at different heights are the instances.

Several methods have been proposed for learning in the MIL scenario, with Deep Learning (DL) approaches achieving remarkable results \citep{song2023artificial}.
A seminal work in the area was the one by \citet{ilse2018attention}, which proposes the attention-based MIL (\abmil) method. \abmil\ features an attention mechanism with the ability to focus on the most relevant instances. 
The attention assigned to each instance can serve as a proxy to find instances with a positive label, i.e., those showing evidence of the disease or injury. 
With this in mind, the attention mechanism has been improved in various ways \citep{lu2021data,zhang2022dtfd,li2021dual}.
However, these methods have a major limitation: they are unable to capture interactions between instances because the attention mechanism treats them independently.
Ultimately, this negatively affects the discriminative performance of these approaches.
Based on the existing literature, two types of interactions among instances have been considered: global and local.

\textit{Global interactions} are those that may exist between any pair of instances in a bag. 
\citet{shao2021transmil} showed that they can contain important information and proposed the \transmil\ model, which includes a Transformer encoder to capture them.
Note that the self-attention mechanism used in Transformers is a natural choice to model these interactions since it computes a similarity score between each pair of instances.
For this reason, different variations of this mechanism were included to boost the performance of existing MIL models \citep{li2021dt,fourkioti2023camil,yang2024variational}.
\textit{Local interactions} are those that exist between neighboring instances. 
A natural way of modeling them is to treat each bag as a graph, where the instances are the nodes and the edges between them represent these local interactions.
This is the approach followed by recent works, which have exploited them in combination with global interactions \citep{fourkioti2023camil,zheng2022graph,zhao2022setmil,castro2024sm} and alone using Graph Neural Networks \citep{li2018graph,chen2021whole}.
Note that an important difference between global and local interactions is that global interactions are \textit{learned} from the data, while local interactions are known beforehand and encoded into the model.

Recently, we proposed an interesting way of leveraging local interactions, called Smooth Attention (\smoothatt) \citep{wu2023smooth}.
It is based on the idea that, in CT scans, a slice is usually adjacent to slices with the same label, see \autoref{fig:ctscan_labels}.
Since the attention values act as a proxy to estimate these labels, they should inherit this property.
To enforce this spatial constraint, \smoothatt\ proposes a novel regularization term based on the Dirichlet energy \citep{zhou2003learning}, which is used to train the \abmil\ model. 
It obtained very promising results in the intracranial hemorrhage (ICH) detection task, outperforming the vanilla ABMIL and other related approaches.
However, SA does not account for global correlations, and it estimates attention values deterministically.

{\color{review}
The deterministic nature of \smoothatt\ and the other previously mentioned methods prevents them from expressing uncertainty in their predictions. However, the ability to model uncertainty is a crucial requirement in medical diagnosis. To address this limitation, we propose the Probabilistic Smooth Attention (\probsmoothatt) framework -- a probabilistic generalization of \smoothatt\ -- which introduces the following key contributions:
}
\begin{enumerate}
    \item \probsmoothatt\ provides a general probabilistic formulation. When a deterministic procedure is used for inference, we recover \smoothatt\ as a particular case.
    When Bayesian inference is leveraged instead, we obtain a new method that estimates attention values through a full probability distribution. As we will see, this yields enhanced predictive performance and allows for uncertainty estimation in the predictions. 
    \item \probsmoothatt\ handles both local and global interactions. We explain how the local interactions in \cite{wu2023smooth} can be combined with global interactions to yield a new method superior to the current state-of-the-art (SOTA) approaches, which usually leverage both types of interactions.
    \item We provide a comprehensive evaluation as required for archival work. \probsmoothatt\ is evaluated using three different datasets (two more than \cite{wu2023smooth}) covering two different medical imaging problems: cancer detection in WSIs and hemorrhage detection in CT scans. Also, \probsmoothatt\ is compared against ten SOTA methods in MIL (six more than those used in \citep{wu2023smooth}).
\end{enumerate}

The rest of the paper is organized as follows. 
\autoref{sec:background} reviews the foundations of our work (\abmil\ and \smoothatt).
\autoref{sec:prob_smooth_att} presents the novel \probsmoothatt\ framework, discussing the probabilistic model and inference as well as the combination with global interactions. 
\autoref{sec:experiments} discusses the empirical evaluation of the model, including dataset description, experimental setup, ablation study to understand \probsmoothatt, and comparison against SOTA methods. 
\autoref{sec:conclusion} provides the main conclusions and limitations. 

\section{Background}
\label{sec:background}

\autoref{subsec:deep_mil} describes the MIL problem and reviews the attention-based MIL model (\abmil)~\citep{ilse2018attention}, which is at the basis of the proposed approach.
\autoref{subsec:deterministic_smooth} describes our deterministic approach to introduce smoothness in the attention values, previously published as Smooth Attention (\smoothatt). 

\subsection{Attention-based Multiple Instance Learning}
\label{subsec:deep_mil}

In MIL, the training set consists of pairs of the form $\left( \bX, Y\right)$, where $\bX \in \Rbb^{N \times P}$ is a bag of instances and $Y \in \left\{ 0,1 \right\}$ is the bag label.
Each bag is composed of a variable number of instances, i.e., $\bX = \left[ \bx_1, \ldots, \bx_{N}\right]^T $, where $\bx_n \in \Rbb^P$ are the instances. 
Each instance $\bx_n$ has an \textit{unobserved} label $y_n \in \{0,1\}$ which is not available during training. 
It is assumed that $Y = \max \left\{ y_1, \ldots, y_{N}\right\}$, that is, a bag $\bX$ is considered positive if and only if there is at least one positive instance in the bag.
The goal is to learn a function that predicts the label of previously unseen bags. 

Attention-based MIL (\abmil) was proposed by \citet{ilse2018attention} to learn in this scenario. 
This method maps the input bag $\bX$ to a bag representation $\bz \in \Rbb^D$, which is later used to predict the bag label. 
To compute this representation they use the attention-based pooling, which assigns an \emph{attention value} $f_n$ to each instance $\bx_n$. This value indicates its importance within the bag and can be used as a proxy to estimate the instance label $y_n$.
Formally,
\begin{equation}
    \bz = \bH^\top \softmax\left( \bff \right), \label{eq:abmil_z}
\end{equation}
where $\bH = \left[ h\left( \bx_1 \right), \ldots, h\left( \bx_N \right) \right]^\top$, $h \colon \Rbb^P \to \Rbb^D$ is a neural network applied to each instance, $\bff = \left[ f_1, \ldots, f_N \right]^\top$, and $f_n = \bw^\top \tanh\left( \bW \bh_n \right)$, with $\bW \in \Rbb^{D_{f} \times D}$ and $\bw \in \Rbb^{D_{f}}$ being trainable weights. 

Note that in \abmil\ both the intermediate embeddings $\bh_n$ and the attention values $\bff$ are obtained by applying a transformation \emph{independently} to each instance. 
As a result, they will be the same regardless of what other instances are in the bag or what structure the bag has. 
In other words, the dependencies between the instances are neglected. 

\subsection{Smooth Attention (\smoothatt): deterministic attention smoothing}
\label{subsec:deterministic_smooth}

In order to take into account the dependencies between the instances during training, Smooth Attention (\smoothatt)~\cite{wu2023smooth} proposes to leverage the following property of medical images: if an instance shows evidence of a lesion, neighboring instances likely contain it too, see \autoref{fig:instance_labels}.
Therefore, for the attention maps to estimate the instance labels accurately, they should also exhibit this \emph{smoothing} property: the attention value assigned to an instance should be similar to the values assigned to surrounding instances. 

\smoothatt\ restricts the solution space by discarding solutions with highly variable attention maps. 
This is achieved through the \textit{Dirichlet energy}, a well-known functional that measures the variability of a function defined on a graph \citep{zhou2003learning}.
For a function $\bff \in \Rbb^N$ defined on a graph with adjacency matrix $\bA = \left[ A_{ij} \right] \in \Rbb^{N \times N}$, the Dirichlet energy of $\bff$ is given by, 
\begin{equation}\label{eq:DE_def}
    \direnergy{\bff, \bA} = \frac{1}{2}\sum_{i=1}^N \sum_{j=1}^N A_{ij} \left( f_i - f_j \right)^2 = \bff^\top \bL \bff,
\end{equation}
where $\bL$ is the graph Laplacian matrix $\bL = \bD - \bA$, and $\bD \in \Rbb^{N \times N}$ is the degree matrix, $\bD = \diag{D_1, \ldots, D_N }$, $D_n = \sum_i A_{ni}$.
\smoothatt\ treats the attention values as a function defined on the bag graph, which is defined by $A_{ij} > 0$ if instances $\bx_i$ and $\bx_j$ are neighbors, and $A_{ij} = 0$ otherwise. 

The Dirichlet energy is used to define a regularization term that penalizes solutions with highly variable attention maps. 
Formally, given a dataset of bags $\left\{ \bX_1, \ldots, \bX_B \right\}$, with corresponding bag labels $\left\{ Y_1, \ldots, Y_B \right\}$ and adjacency matrices $\left\{ \bA_1, \ldots, \bA_B \right\}$, \smoothatt\ uses the architecture proposed in \abmil\ and seeks to minimize the following objective, 
\begin{equation}\label{eq:smoothatt_objective}
    \sum_{b=1}^B \left\{ - \log \p \left( Y_b \mid \bX_b \right) + \lambda \direnergy{ \bff_b, \bA_b } \right\},
\end{equation}
where $\p \left( Y_b \mid \bX_b \right) = \operatorname{Bernouilli}\left(Y_b \mid  \psi\left( {\bH_b}^\top \softmax\left( \bff_b \right) \right) \right) $, $\lambda > 0$ is an hyperparameter, $\psi \colon \Rbb^{D} \to \left[ 0,1 \right]$ is the bag classifier, $\bH_b$ and $\bff_b$ are as in Eq. \eqref{eq:abmil_z}.

\smoothatt\ achieved very promising results in the task of hemorrhage detection, and here we extend it in two ways.
First, since its architecture is based on \abmil, it does not include global interactions, which has proven successful in previous works. 
Second, like the rest of current SOTA deep MIL methods, it is a deterministic approach. 
By estimating a probability distribution over the attention values, we will achieve enhanced discriminative performance as well as interpretable uncertainty estimations.  
In the following section, we propose the Probabilistic Smooth Attention (\probsmoothatt) framework, a generalization of \smoothatt\ that addresses these two aspects. 

\section{Probabilistic Smooth Attention (\probsmoothatt)}
\label{sec:prob_smooth_att}

In this section, we describe the novel Probabilistic Smooth Attention framework.
\autoref{subsec:model_inference} describes the Bayesian formulation of the model.
In \autoref{subsec:global_interactions} we describe how to include global interactions in the architecture of the proposed methodology.
Finally, \autoref{subsec:summary} provides an overview of the different model variants proposed in this work.

\subsection{Model formulation}
\label{subsec:model_inference}

We assume we are given a training set of bags $\cX = \left\{ \bX_1, \ldots, \bX_B \right\}$, with corresponding bag labels $ \cY = \left\{ Y_1, \ldots, Y_B \right\}$ and adjacency matrices $\cA = \left\{ \bA_1, \ldots, \bA_B \right\}$.
Recall that $\bX_b \in \Rbb^{N_b \times D}, Y_b \in \left\{ 0,1 \right\}$, and $ \bA_b \in \Rbb^{N_b \times N_b}$.
Since we will be dealing with bags of different sizes, in the following we denote $\Rbb^{\Nbb \times D} =  \cup_{m \in \Nbb} \Rbb^{m \times D}, \Rbb^{\Nbb} = \cup_{m \in \Nbb} \Rbb^{m}$, and $\left( 0, +\infty \right)^{\Nbb} = \cup_{m \in \Nbb} \left( 0, +\infty \right)^{m}$.

\noindent
\textbf{Probabilistic model and inference.}  We model the attention values of each bag $\bX_b$ as an unobserved latent variable $\bff_b$.
The smoothing property is encoded as a prior distribution given by
\begin{equation}
    \p\left(\bff_b \mid \bA_b\right) \propto \exp\left( - \direnergy{\bff_b, \bA_b} \right).
\end{equation}
This corresponds to the conditional and simultaneous autoregressive models in the statistics literature \cite{ripley1981spatial}. 
Following \abmil\ \cite{ilse2018attention}, the attention values are used to compute the bag label likelihood, 
\begin{equation}
    \p(Y_b \mid \bX_b, \bff_b) = \operatorname{Bernouilli}\left(Y_b \mid  \psi\left( \bH_b^\top \softmax\left( \bff_b \right) \right) \right), 
\end{equation}
where $\bH_b = H\left( \bX_b \right)$, $H\colon \Rbb^{\Nbb \times P} \to \Rbb^{\Nbb \times D}$ is a bag transformation, and $\psi \colon \Rbb^{D} \to \left[ 0,1 \right]$ is the bag classifier. 
Note that in the original \abmil\ and in \smoothatt\ the bag transformation $H$ is implemented by applying a neural network independently to every instance.
We will explain how to include global interactions in $H$ in \autoref{subsec:global_interactions}.
Finally, the joint probabilistic model is obtained assuming independence across bags, 
\begin{equation}
    \p( \cY, \cf \mid \cX, \cA ) =\prod_{b=1}^{B} \p(Y_b \mid \bX_b, \bff_b) \p(\bff_b \mid \bA_b),
\end{equation}
where $\cF = \left\{ \bff_1, \ldots, \bff_B \right\}$.
Making predictions in this model requires computing the posterior of the bag label given the observations. 
More precisely, given a new bag $\bX^*$, the corresponding bag label $Y^*$ is obtained from, 
\begin{equation}
	\p(Y^* \mid \bX^*, \cY) = \int \p(Y^* \mid \bX^*, \bff^*) \p(\bff^* \mid \bX^*, \cY) \dd \bff^*.
\end{equation}
However, computing the posterior $\p(\bff \mid \bX, \cY)$ in closed form is not possible. 
To address this, we follow the Variational Inference (VI) \cite{bishop2006pattern} approach: we approximate it using a variational distribution $\q(\bff \mid \bX, \cY)$, which we write as $\q(\bff \mid \bX)$ in what follows. 
Using the variational distribution, we can make predictions using
\begin{align}
	\p(Y^* \mid \bX^*, \cY) & \approx \int \p(Y^* \mid \bX^*, \bff^*) \q(\bff^* \mid \bX^*) \dd \bff^* = \label{eq:exp_posterior_exact}\\
    & = \Ebb_{\q(\bff^* \mid \bX^*)} \left[ \p(Y^* \mid \bX^*, \bff^*) \right]. \label{eq:exp_posterior}
\end{align}
The optimal choice of $\q(\bff \mid \bX)$ is obtained by minimizing the Kullback-Leibler (KL) divergence between the true posterior and the variational distribution \cite{bishop2006pattern}. This is equivalent to maximizing the following Evidence Lower BOund (ELBO),
\begin{equation}\label{eq:elbo}
    \operatorname{ELBO} = \sum_{b=1}^{B}\Ebb_{\q(\bff_b \mid \bX_b)} \left[  \log \frac{\p(Y_b \mid \bX_b, \bff_b) \p(\bff_b \mid \bA_b )}{\q(\bff_b \mid \bX_b)} \right].
\end{equation}
Unfortunately, it is not possible to obtain the optimal variational distribution in closed form. 
Instead, we proceed as in \cite{kingma2013auto}: we restrict $\q(\bff \mid \bX)$ to belong to some parameterized family of distributions and adjust its parameters by maximizing the ELBO in Eq.~\eqref{eq:elbo}. 
In the following, we consider two choices for this family: as a Gaussian and as a Dirac delta. 

\noindent
\textbf{Modelling $\q(\bff \mid \bX)$ as a Gaussian.}
In this case, we parameterize the variational distribution as a multivariate Gaussian distribution with a diagonal covariance matrix, 
\begin{equation}\label{eq:bayes_gaussian}
    \q(\bff \mid \bX) = \cN \left( \bff \mid \mu \left( \bX \right), \Sigma \left( \bX \right) \right), \quad \Sigma \left( \bX \right) = \diag{\sigma \left( \bX \right) },
\end{equation}
where $\mu \colon \Rbb^{\Nbb \times D} \to \Rbb^{\Nbb}$ and $\sigma \colon \Rbb^{\Nbb \times D} \to \left( 0, +\infty \right)^{\Nbb}$ are implemented using neural networks. 
With this choice, the ELBO in Eq.~\eqref{eq:elbo} can be written as
\begin{equation}\label{eq:elbo_gaussian}
    \hspace{-5pt}
    \operatorname{ELBO} = \sum_{b=1}^{B} \left\{ \Ebb_{\q(\bff_b \mid \bX_b)} \left[  \log \p(Y_b \mid \bX_b, \bff_b) \right] -  \operatorname{KL}\left[ \q(\bff_b \mid \bX_b),  \p(\bff_b \mid \bA_b) \right] \right\}.
\end{equation}
Since sampling from this distribution is very efficient, we can approximate the expectations in the first term and Eq.~\eqref{eq:exp_posterior} -- Eq.~\eqref{eq:elbo} using Monte Carlo sampling and the reparameterization trick \citep{kingma2013auto}. 
Moreover, the KL divergence in the second term admits the following closed-form expression, 
\begin{align}
    \operatorname{KL}\left[ \q(\bff_b \mid \bX_b),  \p(\bff_b \mid \bA_b) \right] = & \ \direnergy{\mu(\bX_b), \bA_b}  + \operatorname{Tr}\left(\bL_b \Sigma(\bX_b) \right) + \label{eq:kl_gaussian_1}\\
    & - \frac{1}{2} \left( \log \left( \frac{\lvert \Sigma(\bX_b) \rvert}{2 \lvert \bL_b \rvert} \right) - N_b \right) \label{eq:kl_gaussian_2},
\end{align}
which can be computed efficiently. 
Note that the KL divergence contains the Dirichlet energy used by \smoothatt\ plus a new term that regularizes the covariance matrix. 
Informally speaking, if we make the covariance matrix very small, $\Sigma(\cdot) \to 0$, the KL divergence becomes the regularization term used by \smoothatt, see Eq. \eqref{eq:smoothatt_objective}.
This can be formalized by modeling $\q(\bff \mid \bX)$ as a Dirac delta, as follows. 

\noindent
\textbf{Recovering \smoothatt: $\q(\bff \mid \bX)$ as a Dirac delta.}
We define the variational distribution as a \textit{deterministic distribution}, i.e., a distribution with zero variance. 
To do so, we resort to the notion of Dirac delta \cite{arfken2011mathematical}. 
However, since it is not a proper function we cannot use it to define our variational density. 
Fortunately, there exists mathematical machinery to use this concept rigorously \cite{arfken2011mathematical,folland2009fourier}, see \ref{app:dirac_delta} for a precise justification of the following statements. 
We define
\begin{equation}\label{eq:varposterior_diracdelta}
    \q(\bff \mid \bX) = \delta \left( \bff - \mu\left(\bX\right) \right) = \begin{cases}
        \infty & \text{if $\bff = \mu(\bX)$},\\
        0 & \text{otherwise},
    \end{cases}
\end{equation}
where $\delta\left( \cdot \right)$ is the Dirac delta and $\mu \colon \Rbb^{\Nbb \times D} \to \Rbb^{\Nbb}$ is implemented using a neural network.
Informally speaking, $\q(\bff \mid \bX)$ assigns all the probability to one point which is determined by the output of $\mu$. 
The ELBO becomes
\begin{equation}\label{eq:elbo_diracdelta}
    \operatorname{ELBO} = \sum_{b=1}^{B} \left\{ \log \p(Y_b \mid \bX_b, \bff_b = \mu(\bX_b)) - \direnergy{ \mu(\bX_b), \bA_b } \right\},
\end{equation}
which is the negative of the objective used by \smoothatt\ with $\lambda = 1$, recall Eq.~\eqref{eq:smoothatt_objective}.
Therefore, the probabilistic model \probsmoothatt\ presented in this section has our previous \smoothatt\ \cite{wu2023smooth} as a particular case. 

\noindent
\textbf{The loss objective.}
Whatever choice we make for $\q(\bff \mid \bX)$ (as a Gaussian or as a Dirac delta), we must maximize the corresponding ELBO with respect to the parameters of the neural network.
In both cases, the negative of the ELBO leads to the following minimization objective, 
\begin{equation}
    \mathcal{L}_{\elbo} = - \operatorname{ELBO} = \mathcal{L}_{\operatorname{LL}} + \mathcal{L}_{\operatorname{KL}},
\end{equation}
where $\mathcal{L}_{\operatorname{LL}}$ is the negative log-likelihood and $\mathcal{L}_{\operatorname{KL}}$ is a KL regularization term, see \autoref{tab:methods_summary}.
Previous works have introduced a balancing hyperparameter $\lambda \in \left[ 0,1 \right]$ to control the strength of the regularization and alleviate KL collapse \cite{higgins2017beta}. Following this idea, we propose to minimize the following objective, 
\begin{equation}\label{eq:lambda_loss}
    \mathcal{L} = \mathcal{L}_{\operatorname{LL}} + \lambda \mathcal{L}_{\operatorname{KL}}.
\end{equation}
Note that by setting $\lambda=1$ we recover the negative of the ELBO, and by setting $\lambda=0$ we drop the KL regularization. 
To avoid the KL vanishing problem, some works have proposed to vary the value of $\lambda$ during training, following different schedules \cite{fu2019cyclical}. 
We will use the cyclical annealing schedule proposed by \citet{fu2019cyclical}, and analyze its impact on the performance in \autoref{subsec:ablation}.

\begin{figure}
    \centering
    \includegraphics[trim={1.6cm 0cm 10.3cm 0cm},clip,width=0.98\textwidth]{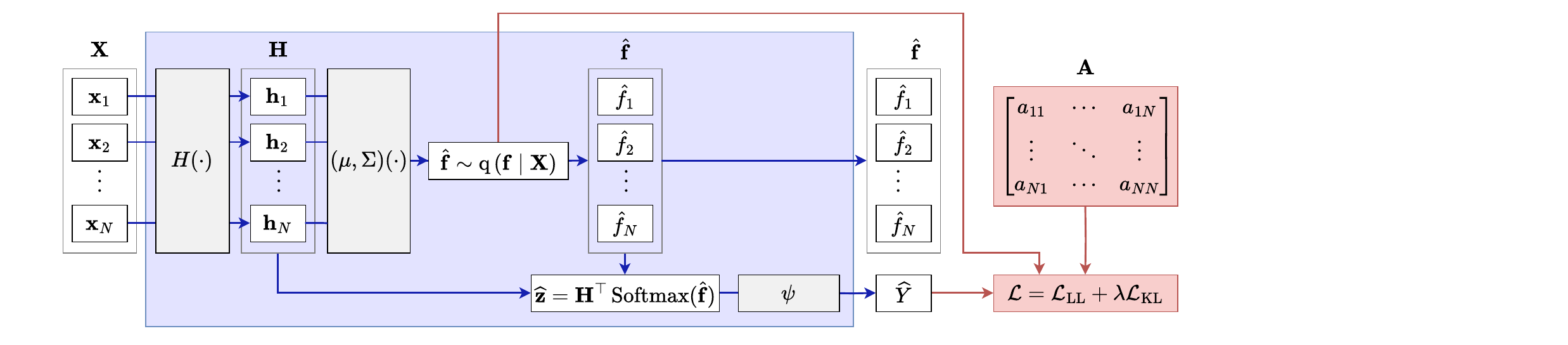}
    \vspace{-0.3cm}
    \caption{
    	\color{review}
        Diagram of the proposed \probsmoothatt. 
        In the forward pass (\textcolor{blue}{blue}), a sample $\hat{\mathbf{f}}$ is drawn from the variational attention distribution $\q \left( \bff \mid \bX \right)$ and then used to compute the bag representation $\widehat{\mathbf{z}}$ and the predicted bag label $\widehat{Y}$. In the loss computation (\textcolor{red}{red}), spatial interactions are incorporated via the bag adjacency matrix into de Kullback-Leibler divergence term.  
    }
    \label{fig:model_diagram}
    \vspace{-0.3cm}
\end{figure}

\noindent
\textbf{Making predictions.} 
To predict the bag label $Y^*$ of a new bag $\bX^*$, we must approximate the integral in \autoref{eq:exp_posterior}. 
The variational posterior choices we have presented facilitate efficient sampling.
Thus, we draw $S$ samples $\left\{ \bff^*_1, \ldots, \bff^*_S  \right\}$ from $\q(\bff^* \mid \bX^*)$ and compute
\begin{equation}\label{eq:predictions}
    \p(Y^* \mid \bX^*, \cY) \approx \frac{1}{S} \sum_{s=1}^S \p(Y^* \mid \bX^*, \bff_s^*).
\end{equation}
{\color{review}This process is illustrated in ~\autoref{fig:model_diagram}. 
For a given input bag, one or more samples are drawn from the variational attention distribution and then used to compute the predicted bag label according to ~\autoref{eq:predictions}. 
}

\subsection{Incorporating global interactions}
\label{subsec:global_interactions}

Both \smoothatt\ and the proposed \probsmoothatt\ take into account local interactions through a regularization term in the training objective, see Eq.~\eqref{eq:lambda_loss}.
However, previous works in deep MIL have demonstrated the benefits of including both global and local interactions \citep{fourkioti2023camil,zheng2022graph}. 
Note that local interactions provide information about the environment of each instance but do not account for long-range dependencies. 
For example, detecting some types of tumors requires finding certain structures in distant regions within the WSI \cite{shao2021transmil,fourkioti2023camil}.
Next, we explain how we can include global interactions in the proposed \probsmoothatt.

{\color{review} As evidenced in ~\autoref{fig:model_diagram}}, one of the main advantages of \probsmoothatt\ is that it is \textit{architecture agnostic}.
This means that we can choose any architecture to implement the bag transformation $H$, the bag classifier $\psi$, or the attention distribution networks $(\mu, \Sigma)$, and still use the same training objective.
Therefore, global interactions can be included by implementing $H$ as a Transformer encoder, 
\begin{equation}
    H\left( \bX \right) = \operatorname{TransformerEnc}(\bX).
\end{equation}
At the core of the Transformer encoder module is the self-attention mechanism, which processes the input by calculating similarity scores between all pairs of instances in the bag. 
These scores determine how much attention each instance should pay to others, allowing the model to capture (global) contextual relationships effectively.
Each layer of the Transformer encoder leverages multiple heads in the self-attention mechanism, along with skip-connections and pre-layer normalization \citep{bishop2024deep}. More details can be found in \ref{app:transformer_encoder}. 


\subsection{Summary of the proposed models}
\label{subsec:summary}

In the previous subsections, we introduced key design choices within the \probsmoothatt\ framework that define different model variants. Specifically, there are two options for the variational posterior $\q\left( \bff \mid \bX \right)$ and two for the bag transformation $H$. These choices can be naturally combined, resulting in four distinct model variants. 


\autoref{tab:methods_summary} summarizes these variants and the fundamental differences between them.
\abmil\ and \transformerabmil\ stand for the two choices for the bag transformation $H$. Similarly, $\Sigma=0$ and $\Sigma=\operatorname{Diag}$ refer to the two choices for the variational posterior $\q\left( \bff \mid \bX \right)$.
Note that we write $\Sigma=0$ for the Dirac delta variational posterior since we recover it by making the variance infinitely small in the Gaussian, $\Sigma(\cdot) \rightarrow 0$.
The variants with $\Sigma=0$ can be regarded as an extension of the deterministic method proposed in our previous work \citep{wu2023smooth}.
They are deterministic because they learn a degenerate distribution for the attention values.
In contrast, the variants with $\Sigma=\operatorname{Diag}$ learn a Gaussian distribution, from which we can sample and study its moments, see \autoref{subsec:inst_pred}.
In the following section, we conduct extensive experimentation to evaluate these four variants.

\begin{table}
\begin{adjustbox}{width=\textwidth}
    \begin{tabular}{@{}ccccc@{}}
        \toprule
        Notation & $H(\bX)$ & $\q(\bff \mid \bX)$ & $\mathcal{L}_{\operatorname{LL}}$ & $\mathcal{L}_{\operatorname{KL}}$ \\ \midrule
        \makecell{\abmil+\probsmooth\\$\Sigma = 0$} & \makecell{$\left[ h(\bx_1), \ldots, h(\bx_N) \right]^\top$\\ $h$ is a MLP} & $\delta \left( \bff - \mu\left(\bX\right) \right)$ & $- \sum_b \log \p(Y_b \mid \bX_b, \bff_b = \mu(\bX_b))$ & $\sum_b \direnergy{\mu(\bX_b), \bA_b}$  \\
        \midrule
        \makecell{\abmil+\probsmooth\\$\Sigma = \operatorname{Diag}$} & \makecell{$\left[ h(\bx_1), \ldots, h(\bx_N) \right]^\top$\\ $h$ is a MLP} & \makecell{$\cN \left( \bff \mid \mu \left( \bX \right), \Sigma \left( \bX \right) \right)$} & $- \sum_b \Ebb_{\q(\bff_b \mid \bX_b)} \left[  \log \p(Y_b \mid \bX_b, \bff_b) \right]$ & \makecell{$\sum_b\operatorname{KL}\left[ \q(\bff_b \mid \bX_b),  \p(\bff_b \mid \bA_b) \right]$\\Eq.\eqref{eq:kl_gaussian_1}-\eqref{eq:kl_gaussian_2}} \\ 
        \midrule
        \makecell{\transformerabmil+\probsmooth\\$\Sigma = 0$} & \makecell{$\operatorname{TransformerEnc}(\bX)$} & $\delta \left( \bff - \mu\left(\bX\right) \right)$ & $- \sum_b \log \p(Y_b \mid \bX_b, \bff_b = \mu(\bX_b))$ & $\sum_b \direnergy{\mu(\bX_b), \bA_b}$ \\
        \midrule
        \makecell{\transformerabmil+\probsmooth\\$\Sigma = \operatorname{Diag}$} & \makecell{$\operatorname{TransformerEnc}(\bX)$} & $\cN \left( \bff \mid \mu \left( \bX \right), \Sigma \left( \bX \right) \right)$ & $- \sum_b \Ebb_{\q(\bff_b \mid \bX_b)} \left[  \log \p(Y_b \mid \bX_b, \bff_b) \right]$ & \makecell{$\sum_b\operatorname{KL}\left[ \q(\bff_b \mid \bX_b),  \p(\bff_b \mid \bA_b) \right]$\\Eq.\eqref{eq:kl_gaussian_1}-\eqref{eq:kl_gaussian_2}} \\ 
        \bottomrule
    \end{tabular}%
\end{adjustbox}
\caption{
    Different variants of the proposed methodology. 
    Each variant is determined by the bag transformation $H$ and the variational posterior $\q(\bff \mid \bX)$. 
    The latter yields different forms of the objective terms $\mathcal{L}_{\operatorname{LL}}$ and $\mathcal{L}_{\operatorname{KL}}$.
    See~\autoref{fig:model_diagram} for a graphical illustration.
}
\label{tab:methods_summary}
\vspace{-0.3cm}
\end{table}



\section{Experiments}
\label{sec:experiments}

In this section, we evaluate the proposed method on three real medical image classification problems, comparing it against ten SOTA MIL methods.
Our code will be made publicly available upon the acceptance of the paper at the following Github repository: \url{https://github.com/Franblueee/ProbSA-MIL}. 
In \autoref{subsec:exp_framework} we describe the experimental framework. 
In \autoref{subsec:ablation} we perform an ablation study to better understand the proposed method.
In \autoref{subsec:comparison_sota} we compare our method against ten SOTA methods in MIL.
Finally, \autoref{subsec:inst_pred} shows how the probabilistic treatment can help in by pointing out regions of the attention maps where the method may not be confident enough.

\subsection{Experimental framework}
\label{subsec:exp_framework}

\noindent
\textbf{Datasets.}
We evaluate the proposed method on three medical MIL datasets: RSNA \citep{flanders2020construction}, PANDA \citep{bulten2022artificial}, and CAMELYON16 \citep{bejnordi2017diagnostic}. 
In RSNA the goal is to detect acute intracranial hemorrhage from CT scans. There are 1150 CT scans, each of them having from 24 to 57 slices. They are labeled as having hemorrhage (at least one slice shows evidence of hemorrhage) or not hemorrhage (no slice shows evidence of hemorrhage). 
In PANDA the goal is to detect prostate cancer from microscopy scans of prostate biopsy samples. It is composed of 10616 WSIs at $10\times$ magnification. 
In CAMELYON16 the goal is to detect breast cancer metastasis. It consists of 400 WSIs at $20\times$ magnification.
In both PANDA and CAMELYON16, we extract patches of size $512 \times 512$ using the method proposed by \citet{lu2021data}. 
Each WSI is labeled as tumorous (at least one patch shows evidence of tumor) and non-tumorous (no patch shows evidence of tumor).
For RSNA and CAMELYON16, we use the standard train/test partition. 
For PANDA, since the labels for the test set have not been released, we use the train/test split proposed by \citet{silva2021self}.
We split the initial train data into five different train/validation splits. 
Every model is trained on each of these splits and then evaluated on the test set.
We report the average performance on this test set. 

\noindent
\textbf{Metrics.}
Medical imaging datasets are typically imbalanced, containing more negative examples (those that do not show evidence of disease or lesion) than positive examples (those that show evidence of disease or lesion). 
For this reason, we analyze the performance of each method using the area under the ROC curve (AUROC) and the F1 score.
We also report the average rank: we sort the different methods according to their performance on each tuple (metric, dataset), and report the average position.

\noindent
\textbf{Feature extraction.}
Neither the proposed method nor the methods we compare with can be trained end-to-end alongside a feature extractor. 
This limitation arises because, for the datasets under consideration, almost no single bag can fit within the memory of commercially available GPUs.
For this reason, we extract features from each instance using a pre-trained model. 
To choose this model, we note that previous works have pointed out the importance of using large-scale domain-aligned feature extractors \citep{kang2023benchmarking,wang2024pathology}, with several improvements in MIL \citep{shao2021transmil,li2021dual,fourkioti2023camil}.
For pathology data (such as PANDA and CAMELYON16), the mentioned studies provide the weights of these models. 
However, we have not been able to find them for CT scans.
For this reason, for the RSNA dataset, we use a ResNet50 pre-trained on Imagenet. 
For PANDA and CAMELYON16, we use a ResNet50 pre-trained using the Barlow Twins self-supervised learning method on a huge dataset of WSIs patches \citep{ kang2023benchmarking}, whose weights can be found online\footnote{Weights available \href{https://github.com/lunit-io/benchmark-ssl-pathology}{here}.}.
Both choices transform the input instance into a vector of $P=2048$ components.
Finally, note that we use the same features for every MIL method, which ensures a fair comparison. 

\noindent
\textbf{MIL methods and architectures.}
We compare the proposed \probsmoothatt\ with SOTA MIL methods. 
To ensure a fair comparison, we divide them into two groups, depending on whether or not they take into account global interactions through the self-attention mechanism. 
In the first group, we include those methods that do not account for these interactions: the proposed \abmil+\probsmoothatt, \abmil ~\citep{ilse2018attention}, \clam ~\citep{lu2021data}, \dsmil ~\citep{li2021dual}, \pathgcn ~\citep{chen2021whole}, and \dtfdmil ~\citep{zhang2022dtfd}.
In the second group we include those methods that do account for these interactions: the proposed \transformerabmil+\probsmoothatt, \transmil ~\citep{shao2021transmil}, \setmil ~\citep{zhao2022setmil}, \gtp ~\citep{zheng2022graph}, \iibmil ~\citep{ren2023iib}, \camil ~\citep{fourkioti2023camil}, {\color{review} and \vmil ~\citep{yang2024variational}}. 
Note that the proposed \probsmoothatt\ outperforms the related \cite{castro2024sm}, which uses the same datasets and metrics.

For every method, we use the original implementation, which is publicly available on their GitHub repositories.
We modify the number of layers and their dimensions to make the comparison under the same parameter budget.
Every method first transforms the input instances using one fully connected layer with 512 units ($D=512$).
The different variants of the attention pooling used by \abmil, \clam, \dsmil, \pathgcn, and \dtfdmil\ are implemented with an inner dimension of $D_f=128$.
The different transformer encoders used by \transmil, \setmil, \gtp, \iibmil, and \camil\ have 2 transformer layers; key, query, and value dimensions of $512$, and $8$ attention heads.
The bag-embedding classifier is implemented using one fully connected layer.
The attention pooling used by \abmil+\probsmoothatt\ and \transformerabmil+\probsmoothatt\ is the same as in \abmil. 
The transformer encoder in \transformerabmil+\probsmoothatt\ uses Pytorch's implementation of dot product attention\footnote{Pytorch's implementation of dot product attention is available \href{https://pytorch.org/docs/stable/generated/torch.nn.functional.scaled_dot_product_attention.html}{here}.}.
{\color{review}
The bag adjacency matrices (see Eq.~\eqref{eq:DE_def}) are constructed at the beginning of training and remain static throughout the process—that is, they are not learned or updated during optimization. Following the approach of ~\citep{fourkioti2023camil}, we compute edge weights based on the inverse of the distance between corresponding instances in the feature space.
This ensures that, if the feature space is appropriately structured, the weight between a positive and a negative instance will be close to zero, which suppresses attention smoothing across boundaries. 
}

\noindent
\textbf{Training setup.} 
To ensure fair and reproducible conditions, we trained each method under the same setup.
{\color{review2}
Since the authors of \iibmil\ do not share their training code, we implemented the training procedure following the description in the original manuscript.
}
The number of epochs was set to $100$. 
We used the Adam optimizer with the default Pytorch configuration. 
The base learning rate was set to $10^{-4}$. 
We adopted a slow training start using Pytorch's \texttt{LinearLR} scheduler with \texttt{start\_factor=0.1} and \texttt{total\_iters=10}. 
During training, we monitored the bag AUROC in the validation set and kept the weights that obtained the best results in terms of bag AUROC. 
In RSNA and PANDA, the batch size was set to 32.
In CAMELYON16, it was set to 4 for non-transformer methods, and to 1 for transformer-based methods. 
For SETMIL, however, we had to set it to 1 in PANDA and CAMELYON16 due to its high GPU memory requirements. 
In RSNA and PANDA, we weighted the loss function to account for the imbalance between positive and negative bags. 
All experiments were performed on an NVIDIA GeForce RTX 3090. 

\subsection{Ablation study: analyzing the novel \probsmoothatt}
\label{subsec:ablation}

The aim of this subsection is to study the effect of three key components of our method: the baseline method on top of which it is implemented (\abmil\ vs \transformerabmil), the hyperparameter $\lambda$, and the variational posterior $\q(\bff \mid \bX)$. 


To analyze this, we ran the four variants described in \autoref{tab:methods_summary} with $\lambda \in \left\{ 0, 0.1, 0.5, 1.0, \textnormal{cyclical} \right\}$.  
Here, $\lambda=0$ corresponds to the baseline models \abmil\ and \transformerabmil, on which we build. 
Choosing $\lambda=0.1$ removes most of the influence of the KL term, while setting $\lambda=1.0$ corresponds to using the actual ELBO. 
When $\lambda=\textnormal{cyclical}$, out method adopts the cyclical schedule, mentioned above. Namely, the training process is divided into $M=5$ cycles of equal number of training steps $L$ ($L$ is obtained as the total number of training steps divided by $M$).
In each cycle, the hyperparameter starts at $\lambda=0$ and increases linearly for $\floor*{0.8L}$ training steps until it reaches $\lambda=1$, and stays like that for the rest of the cycle.

\begin{figure}[t!]
    \begin{center}
         \includegraphics[trim={2.5cm 8.5cm 2.5cm 0.5cm},clip,width=0.3\textwidth]{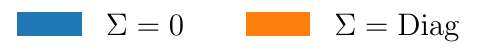}
         \vspace{-0.4cm}
    \end{center}
    \begin{subfigure}[b]{0.45\textwidth}
        \centering
        \includegraphics[width=\textwidth]{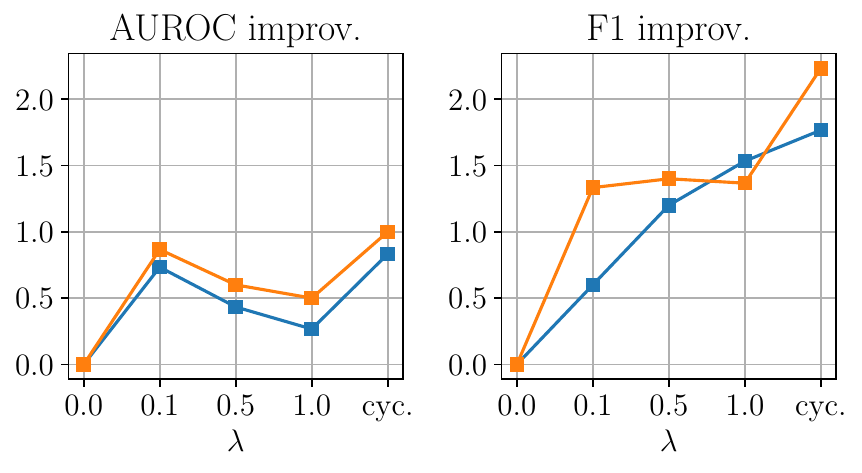}
        \caption{\abmil+\probsmoothatt.}
    \end{subfigure}
    \quad
    \begin{subfigure}[b]{0.45\textwidth}
        \centering
        \includegraphics[width=\textwidth]{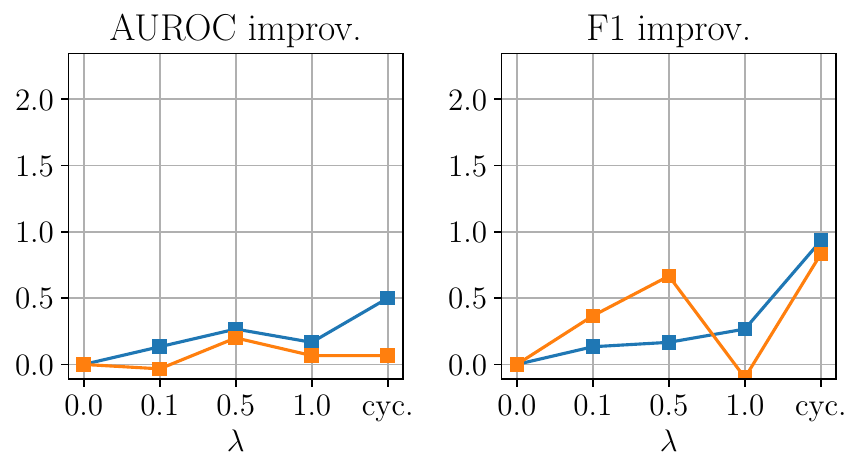}
        \caption{\transformerabmil+\probsmoothatt.}
    \end{subfigure}
    \vspace{-0.2cm}
    \caption{
        Average improvement over the baseline ($\lambda=0)$ for different values of the balancing hyperparameter.
        For most values of this hyperparameter, the improvement introduced by \probsmoothatt\ is positive, regardless of the posterior choice $\Sigma=0$ or $\Sigma=\operatorname{Diag}$.
        As theoretically expected, the simpler model \abmil\ benefits from a richer posterior ($\Sigma=\operatorname{Diag}$) more than \transformerabmil\ does.
    }
    \label{fig:ablation_mean_improvement}
    \vspace{-0.3cm}
\end{figure}

The results are shown in \autoref{fig:ablation_mean_improvement}, where we have plotted the average improvement over the baselines ($\lambda=0$) for different values of $\lambda$. 
The breakdown of these results for each dataset is in \autoref{tab:ablation_results}.
We analyze them in the following. 

\noindent
\textbf{Improvement over the baselines.} First, we study whether the proposed \probsmoothatt\ ($\lambda>0$) improves the results with respect to the baselines on top of which it is implemented ($\lambda=0$).
{\color{review}
In \autoref{fig:ablation_mean_improvement}, most values of $\lambda$ correspond to points above zero, indicating that the proposed \probsmoothatt\  improves performance compared to the baseline approaches. 
Notably, the improvement is negative for only two $\lambda$ values in the \transformerabmil\ variant, with a value very close to zero. 
This suggests the proposed method does not significantly degrade performance in these cases.}
We also note that the improvement is greater for the \abmil\ architecture than for \transformerabmil.
This is reasonable, as simpler architectures tend to benefit more from a probabilistic formulation. 
When looking at \autoref{tab:ablation_results} the baselines ($\lambda=0$) obtain the worst rank in the first three variants. 
{
\color{review}
The last variant (\transformerabmil+\probsmoothatt\ with $\Sigma=\operatorname{Diag}$) does not outperform the baseline in RSNA and PANDA, but does so in CAMELYON16 by a wide margin (98.486 vs 97.673 and 94.657 vs 91.826).
}
 

\noindent
\textbf{Optimal value of $\lambda$.} 
Next, we analyze the best choice for the balancing hyperparameter. 
We observe that $\lambda=\cyclical$ provides the best performance on average: it achieves the highest improvement in \autoref{fig:ablation_mean_improvement} and the best rank in \autoref{tab:ablation_results}. 
Note that the optimal value of $\lambda$ is highly dependent on the dataset and architecture.
\autoref{tab:ablation_results} suggests that for \abmil+\probsmoothatt\ higher values of $\lambda$ yield better results on RSNA, while lower values are preferred for PANDA. 
This suggests that the cyclical scheduler will deliver a good performance in most cases, although it can be outperformed by conducting a grid search for the given dataset. 
This is consistent with what was observed in \citep{fu2019cyclical}: the cyclical schedule allows the progressive learning of more meaningful attention values. 
For this reason, we report the results corresponding to $\lambda=\cyclical$ in the next section.

\noindent
\textbf{Choice of $\q(\bff \mid \bX)$, i.e, $\Sigma = 0$ vs $\Sigma = \operatorname{Diag}$.} 
Finally, we compare the two options considered for the variational posterior: the Gaussian ($\Sigma=\operatorname{Diag}$) and the Dirac delta ($\Sigma=0$). 
From \autoref{fig:ablation_mean_improvement} it is clear that the Gaussian provides a wider improvement when paired with the \abmil\ architecture.
For \transformerabmil, one could argue that the Dirac delta is a better choice in terms of AUROC, but it is not as clear for the F1 metric. 
However, the Gaussian posterior has an advantage over the Dirac delta: it naturally produces instance uncertainty maps, as will be discussed in Section \ref{subsec:inst_pred}.

\subsection{Comparison against existing methods}
\label{subsec:comparison_sota}

The goal of this subsection is to compare the proposed \probsmoothatt\ with SOTA deep MIL methods. 
As we described in \autoref{subsec:exp_framework}, we categorize these methods into two families to ensure a fair comparison in terms of their architectural capabilities. 
The first group comprises methods that do not model global interactions within their architecture. 
This group includes our \abmil+\probsmoothatt. 
The second group consists of methods that incorporate global interactions, typically leveraging the self-attention mechanism. Our \transformerabmil+\probsmoothatt\ falls into this category. 

\begin{figure}[t!]
    \centering
    \begin{subfigure}[b]{0.48\textwidth}
        \centering
        \includegraphics[width=\linewidth]{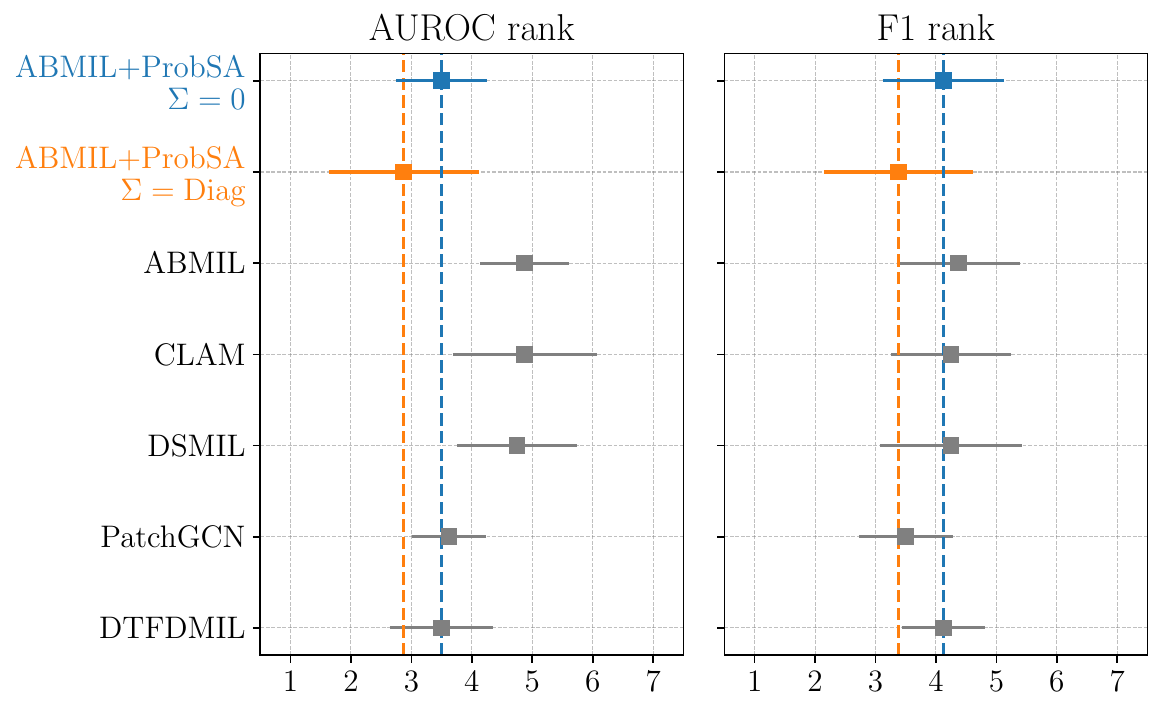}
        \caption{\abmil+\probsmoothatt.}
        \label{fig:sota_rank_no_transformers}
    \end{subfigure}
    \quad
    \begin{subfigure}[b]{0.48\textwidth}
        \centering
        \includegraphics[width=\linewidth]{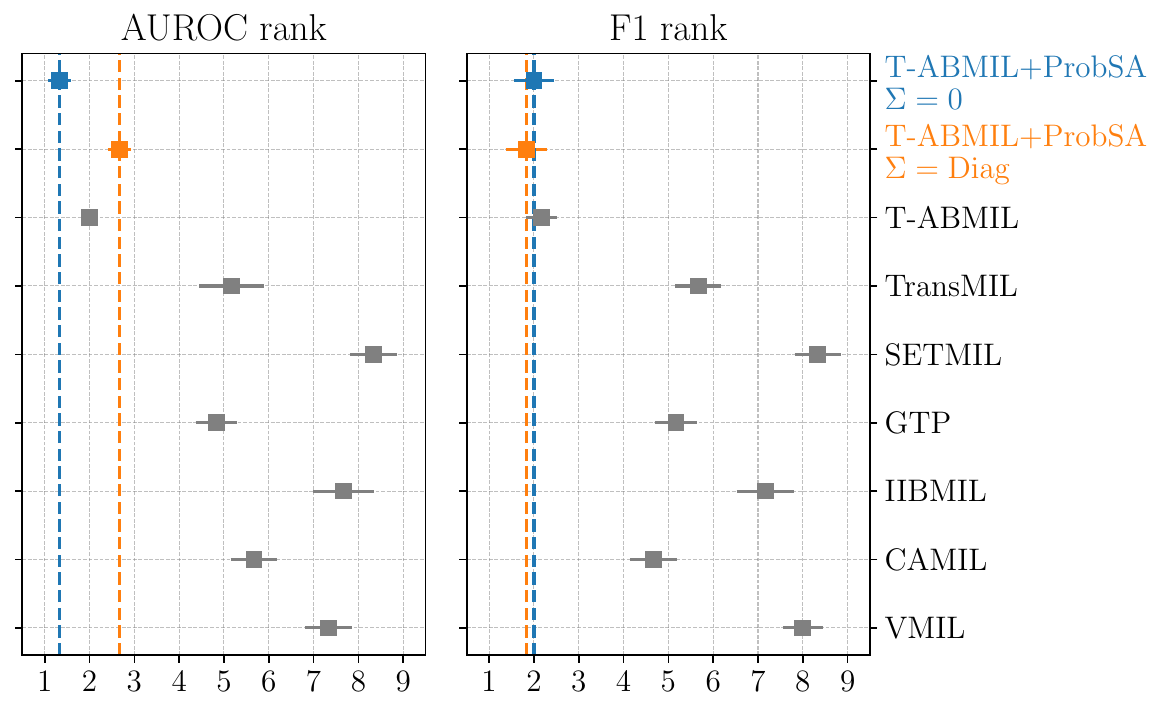}
        \caption{\transformerabmil+\probsmoothatt.}
        \label{fig:sota_rank_transformers}
    \end{subfigure}
    \caption{
        Average AUROC and F1 rank for each family of methods. Note that lower rank indicates better performance. The proposed \probsmoothatt\ attains the best rank in each classification metric. The Gaussian variant ($\Sigma = \operatorname{Diag}$) places first in three out of four (metric, family) pairs. 
    }
    \label{fig:sota_rank}
\end{figure}

\begin{table}[t!]
	\centering
	\begin{adjustbox}{width=\textwidth}
		\begin{tabular}{@{}ccccccccc@{}}
			\toprule
			& \multicolumn{2}{c}{\textbf{RSNA}} & \multicolumn{2}{c}{\textbf{PANDA}} & \multicolumn{2}{c}{\textbf{CAMELYON16}} \\ \midrule
			Model & AUROC $(\uparrow)$ & F1 $(\uparrow)$ & AUROC $(\uparrow)$ & F1 $(\uparrow)$ & AUROC $(\uparrow)$ & F1 $(\uparrow)$ & Rank $(\downarrow)$ \\ \midrule
			\abmil+\probsmooth($\Sigma=0$) & $\underline{90.189}_{0.482}$ & $\underline{83.168}_{1.259}$ & $97.896_{0.090}$ & $94.973_{0.140}$ & $97.633_{0.618}$ & $91.307_{1.053}$ & $4.000_{2.098}$ \\
			\abmil+\probsmooth($\Sigma=\operatorname{Diag}$) & $\mathbf{90.231}_{0.401}$ & $\mathbf{83.230}_{1.719}$ & $98.094_{0.069}$ & $95.274_{0.236}$ & $\mathbf{97.885}_{1.259}$ & $\mathbf{92.351}_{2.385}$ & $\mathbf{1.833}_{1.329}$ \\
			\abmil & $87.924_{1.179}$ & $78.317_{2.531}$ & $97.843_{0.201}$ & $95.060_{0.435}$ & $97.454_{1.074}$ & $90.756_{2.483}$ & $5.667_{0.816}$ \\
			\clam & $82.764_{6.162}$ & $52.941_{32.008}$ & $97.915_{0.191}$ & $95.169_{0.176}$ & $96.985_{1.561}$ & $\underline{91.656}_{1.103}$ & $5.333_{1.862}$ \\
			\dsmil & $87.539_{1.973}$ & $76.164_{5.715}$ & $98.043_{0.135}$ & $\mathbf{95.398}_{0.269}$ & $93.301_{2.042}$ & $85.801_{3.320}$ & $5.167_{2.317}$ \\
			\pathgcn & $89.598_{1.036}$ & $79.748_{2.472}$ & $\mathbf{98.103}_{0.238}$ & $\underline{95.365}_{0.255}$ & $97.469_{0.873}$ & $91.457_{2.430}$ & $\underline{2.667}_{1.033}$ \\
			\dtfdmil & $88.529_{0.360}$ & $79.348_{1.034}$ & $\underline{98.096}_{0.192}$ & $95.361_{0.298}$ & $\underline{97.821}_{0.532}$ & $90.762_{1.988}$ & $3.333_{1.211}$ \\
			\bottomrule
		\end{tabular}
	\end{adjustbox}
	\caption{
		Classification results (mean and standard deviation from five independent runs) for methods that do not model global interactions.
		The best is in bold and the second-best is underlined.
		$(\downarrow)$/$(\uparrow)$ means lower/higher is better.
	}
	\label{tab:classification_results_no_transformers}
    \vspace{-0.3cm}
\end{table}

\noindent
\textbf{Methods without global interactions.}
\autoref{tab:classification_results_no_transformers} shows the results in terms of AUROC and F1 for methods that do not account for global interactions.
The average rank for each metric is shown in \autoref{fig:sota_rank_no_transformers}.
The proposed \probsmoothatt\ with a Gaussian variational posterior ($\Sigma=\operatorname{Diag}$) achieves the highest rank and the best result in four out of six (dataset, metric) pairs. 
{\color{review2}
We observe that \pathgcn\ achieves the best performance in PANDA and the second-best rank overall. 
This can be attributed to the fact that its architecture is the only one that incorporates local interactions through the graph convolutional layer.
}

\noindent
\textbf{Methods with global interactions.}
\autoref{tab:classification_results_transformers} shows the results in terms of AUROC and F1 for methods that do not account for global interactions. 
The average rank for each metric is shown in \autoref{fig:sota_rank_transformers}.
{\color{review}
The two variants of the proposed \probsmoothatt\ are in the top 3.}
The Dirac delta variant ($\Sigma=0$) yields the best AUROC rank, while the Gaussian variant ($\Sigma=\operatorname{Diag}$) yields the best F1 rank, see \autoref{fig:sota_rank_transformers}. 
Interestingly, the Gaussian variant performs best in PANDA and is very competitive in CAMELYON16, where it obtains the highest F1, but falls short in RSNA. 
{\color{review2}
We observe that \iibmil\ and \vmil\ perform notably worse on the CAMELYON16 dataset, despite achieving competitive results on RSNA and PANDA. This discrepancy can be attributed to the challenging nature of CAMELYON16. In this dataset, bags are significantly larger -- containing, on average,  approximately five times more instances per bag than PANDA -- and the proportion of positive instances is very low. Since both \iibmil\ and \vmil\ rely on an instance-level classifier that requires the identification of positive instances during training, the difficulty in locating positive instances in CAMELYON16 likely hampers the training process, leading to inaccurate predictions.
}

\noindent
{\color{review2}
\textbf{Performance in RSNA.}
Finally, we observe that all methods perform significantly worse on the RSNA dataset compared to the other two. This discrepancy can be attributed to the feature extractor used for RSNA, which was not pretrained on data from this domain, resulting in a feature space with reduced representational capacity.
}

\begin{table}
	\centering
	\begin{adjustbox}{width=\textwidth}
		\begin{tabular}{@{}ccccccccc@{}}
			\toprule
			& \multicolumn{2}{c}{\textbf{RSNA}} & \multicolumn{2}{c}{\textbf{PANDA}} & \multicolumn{2}{c}{\textbf{CAMELYON16}} \\ \midrule
			Model & AUROC $(\uparrow)$ & F1 $(\uparrow)$ & AUROC $(\uparrow)$ & F1 $(\uparrow)$ & AUROC $(\uparrow)$ & F1 $(\uparrow)$ & Rank $(\downarrow)$ \\ \midrule
            \transformerabmil+\probsmooth($\Sigma=0$) & $\mathbf{91.781}_{1.200}$ & $\mathbf{84.591}_{2.535}$ & $97.974_{0.156}$ & $95.213_{0.072}$ & $\mathbf{98.418}_{1.104}$ & $\underline{94.220}_{2.168}$ & $\mathbf{1.833}_{0.983}$ \\
            \transformerabmil+\probsmooth($\Sigma=\operatorname{Diag}$) & $90.814_{1.683}$ & $83.683_{3.669}$ & $\underline{97.988}_{0.117}$ & $\mathbf{95.339}_{0.249}$ & $98.133_{0.828}$ & $\mathbf{94.732}_{3.071}$ & $2.167_{0.983}$ \\
            \transformerabmil & $\underline{91.083}_{0.978}$ & $\underline{84.083}_{1.962}$ & $\mathbf{98.014}_{0.226}$ & $\underline{95.289}_{0.322}$ & $\underline{98.209}_{0.695}$ & $92.967_{3.638}$ & $\underline{2.000}_{0.632}$ \\
            \transmil & $90.271_{0.534}$ & $81.258_{2.541}$ & $96.921_{0.239}$ & $93.918_{0.337}$ & $97.811_{2.136}$ & $91.773_{2.276}$ & $5.500_{1.225}$ \\
            \setmil & $61.588_{0.811}$ & $12.061_{16.515}$ & $86.974_{1.203}$ & $79.236_{1.305}$ & $75.340_{0.900}$ & $64.151_{2.001}$ & $8.333_{1.033}$ \\
            \gtp & $90.296_{1.464}$ & $82.324_{4.364}$ & $97.768_{0.199}$ & $94.590_{0.694}$ & $95.102_{1.034}$ & $90.096_{1.045}$ & $4.833_{0.983}$ \\
            \iibmil & $86.877_{1.152}$ & $68.290_{6.579}$ & $97.417_{0.064}$ & $94.578_{0.182}$ & $45.342_{5.937}$ & $10.971_{24.533}$ & $7.500_{1.225}$ \\
            \camil & $88.686_{2.245}$ & $80.409_{2.846}$ & $97.489_{0.111}$ & $94.676_{0.314}$ & $96.811_{1.510}$ & $91.949_{2.328}$ & $5.167_{1.169}$ \\
            \vmil & $89.560_{1.153}$ & $69.622_{1.964}$ & $93.763_{0.357}$ & $83.921_{0.281}$ & $50.000_{0.000}$ & $00.000_{0.000}$ & $7.667_{1.033}$ \\
			\bottomrule
		\end{tabular}
	\end{adjustbox}
	\caption{
		Classification results (mean and standard deviation from five independent runs) for methods that model global interactions.
		The best is in bold and the second-best is underlined.
		$(\downarrow)$/$(\uparrow)$ means lower/higher is better.
	}
	\label{tab:classification_results_transformers}
    \vspace{-0.3cm}
\end{table}

\subsection{Exploring instance predictions}
\label{subsec:inst_pred}

We conclude the experimental section by exploring the localization capabilities of each method.
Recall that deep MIL methods usually assign a scalar to each instance that reflects its importance within the bag.
For simplicity, we will refer to these scalars as \textit{attention values}, although they can be obtained in different ways (e.g., using GradCam as in \gtp~\citep{zheng2022graph}). 
The proposed \probsmoothatt, instead of assigning an attention value to each instance, outputs a probability distribution, denoted as $\q\left( \bff \mid \bX \right)$.  
In this section, we show that the first and second-order moments of the Gaussian variational posterior $\cN \left( \mu \left( \bX \right), \Sigma \left( \bX \right) \right)$ provide important information for illness localization.

%

\begin{figure}[t!]
	\begin{center}
		\begin{adjustbox}{width=\textwidth}
			\centering
			\begin{tblr}{
					colspec = {X[c,h,2.5cm]X[c]},
					cells   = {font = \fontsize{8pt}{8pt}\selectfont},
                    rows={rowsep=0.2pt}
				}
				& \makecell{\includegraphics[trim={0cm 0cm 0cm 0cm},clip,width=4.5cm]{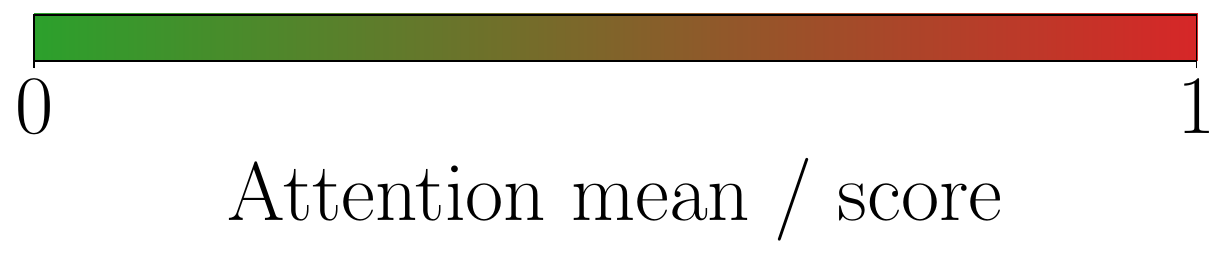} \hfill \includegraphics[trim={0cm 0cm 0cm 0cm},clip,width=4.5cm]{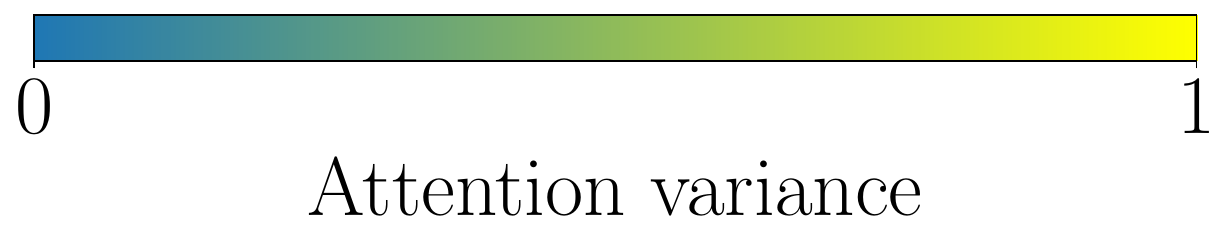}} \\
                \vspace{-10pt}
				CT scan & \includegraphics[trim={0cm 0cm 0cm 0cm},clip,width=0.75\textwidth]
				{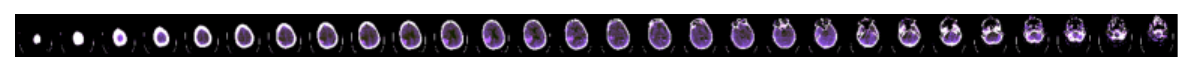} \\
				Slice labels & \includegraphics[trim={0cm 0cm 0cm 0cm},clip,width=0.75\textwidth]
				{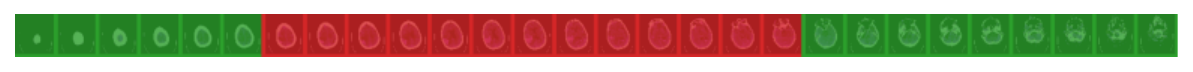} \\
                \vspace{-15pt}
				\makecell{\abmil+\probsmoothatt \\ Mean} & \includegraphics[trim={0cm 0cm 0cm 0cm},clip,width=0.75\textwidth]{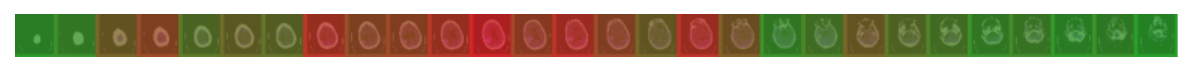} \\
                \vspace{-15pt}
				\makecell{\abmil+\probsmoothatt \\ Variance} & \includegraphics[trim={0cm 0cm 0cm 0cm},clip,width=0.75\textwidth]{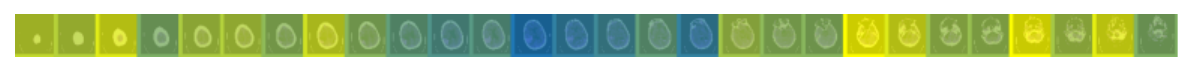} \\
                \vspace{-10pt}
				\pathgcn & \includegraphics[trim={0cm 0cm 0cm 0cm},clip,width=0.75\textwidth]{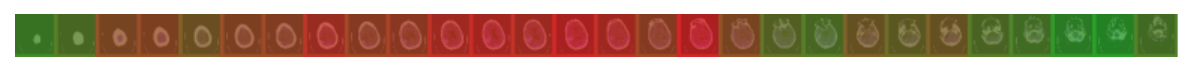} \\
                \vspace{-10pt}
				\dtfdmil & \includegraphics[trim={0cm 0cm 0cm 0cm},clip,width=0.75\textwidth]{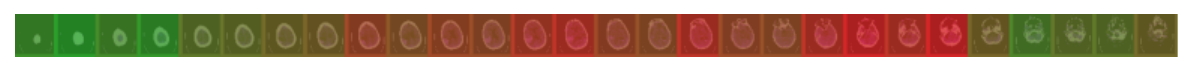} \\
                \vspace{-10pt}
				\gtp & \includegraphics[trim={0cm 0cm 0cm 0cm},clip,width=0.75\textwidth]{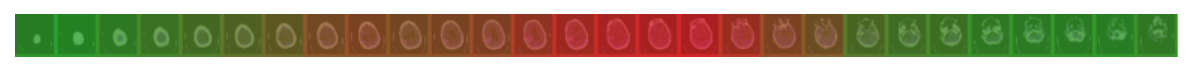} \\
                \vspace{-10pt}
				\camil & \includegraphics[trim={0cm 0cm 0cm 0cm},clip,width=0.75\textwidth]{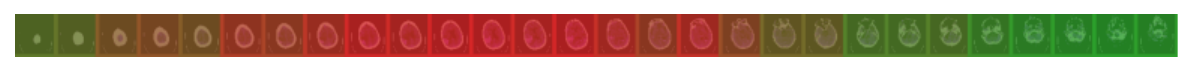}
			\end{tblr}
		\end{adjustbox}
        \vspace{-0.3cm}
		\caption{
			Attention maps in a CT scan from RSNA. The attention values have been normalized to ease visualization.
            The proposed \probsmoothatt\ produces a very accurate attention map and flags false positives with high variance. 
		}
		\label{fig:attmaps-rsna-main}
        \vspace{-0.3cm}
	\end{center}
    
\end{figure}

\begin{figure}[t!]
    \footnotesize
    \begin{center}
        \centering
        \begin{adjustbox}{width=0.7\textwidth}
        \begin{tabular}{cc}
             \includegraphics[trim={0cm 0cm 0cm 0cm},clip,width=5cm]{img/att_map-bar_horizontal.pdf} 
             &
             \includegraphics[trim={0cm 0cm 0cm 0cm},clip,width=5cm]{img/uncert_map-bar_horizontal.pdf}
        \end{tabular}
        \end{adjustbox}
        \begin{adjustbox}{width=1.0\textwidth}
        \begin{tabular}{ccccc}
            \includegraphics[trim={0cm 0cm 0cm 0cm},clip,width=0.22\textwidth]{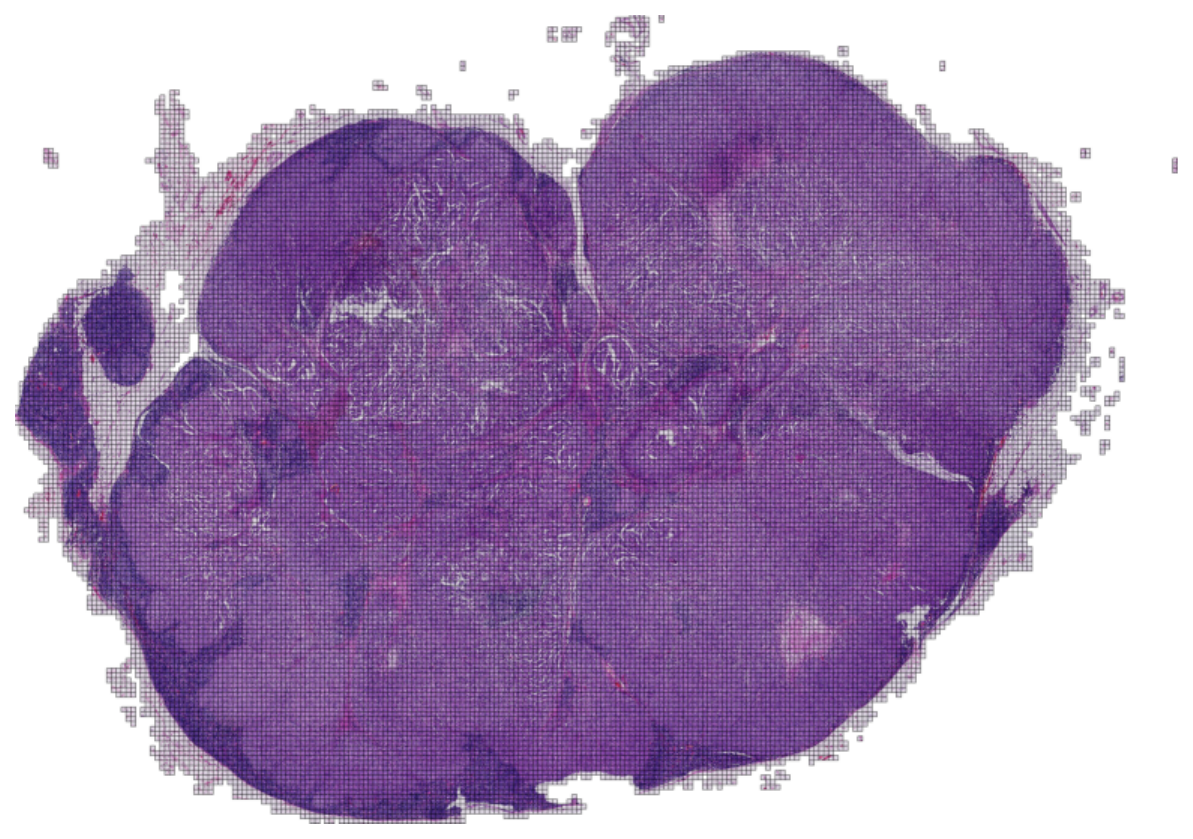}
            & 
            \includegraphics[trim={0cm 0cm 0cm 0cm},clip,width=0.22\textwidth]
            {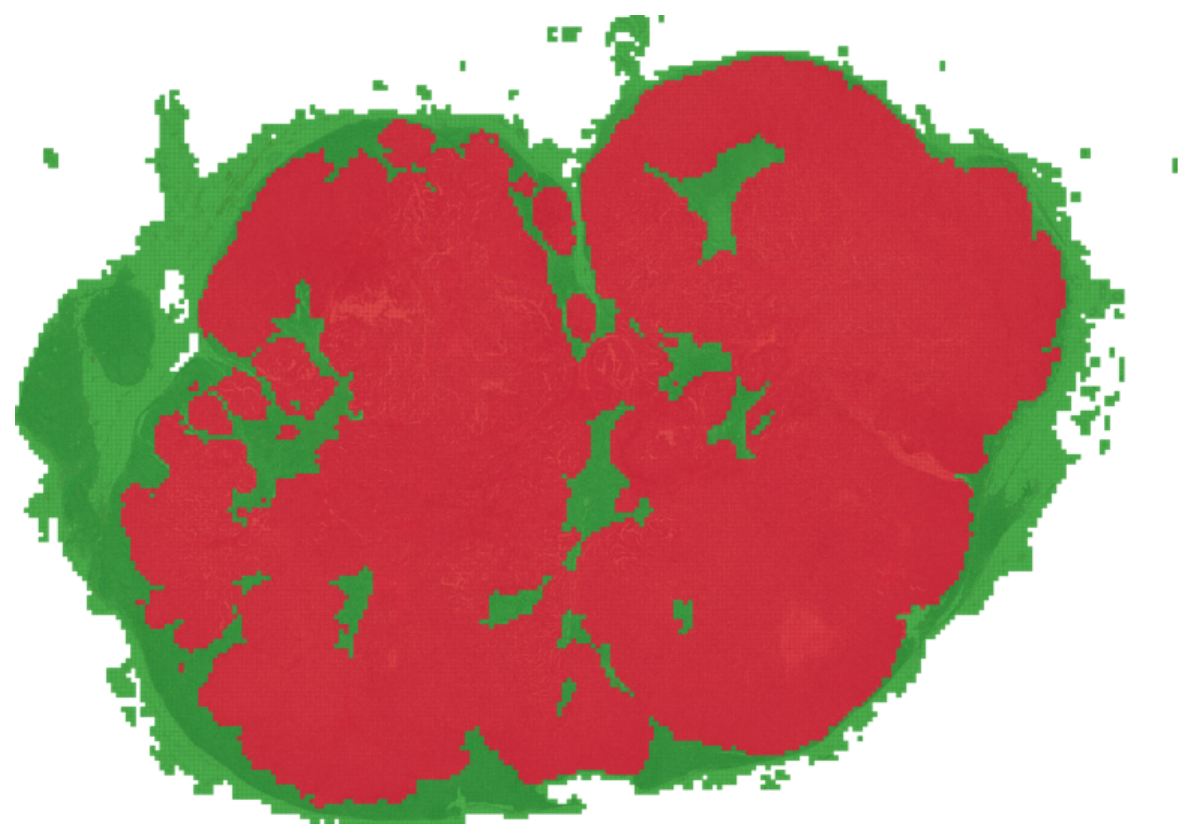}
            & 
            \includegraphics[trim={0cm 0cm 0cm 0cm},clip,width=0.22\textwidth]
            {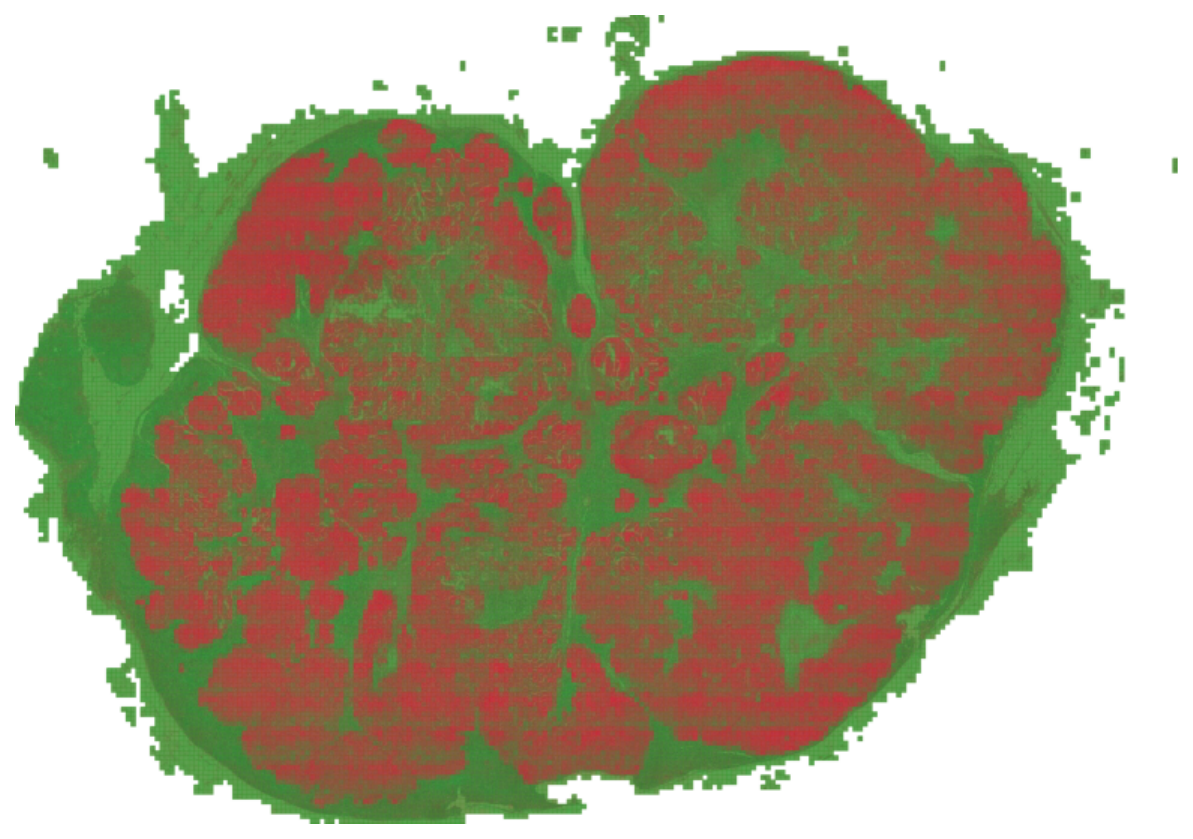}
            & 
            \includegraphics[trim={0cm 0cm 0cm 0cm},clip,width=0.22\textwidth]
            {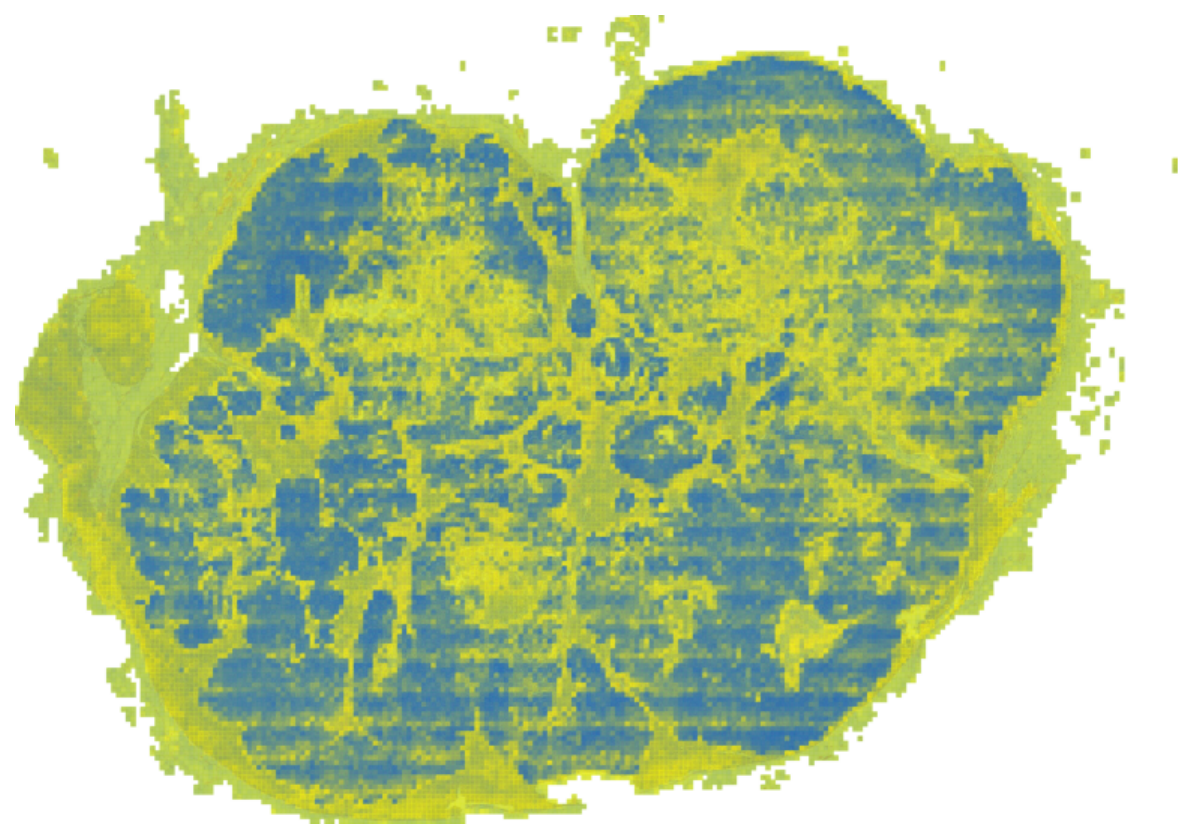}
            \\
            WSI &  Patch labels & \makecell{\abmil+\probsmoothatt \\ Mean} & \makecell{\abmil+\probsmoothatt \\ Variance} \\
            \includegraphics[trim={0cm 0cm 0cm 0cm},clip,width=0.22\textwidth]
            {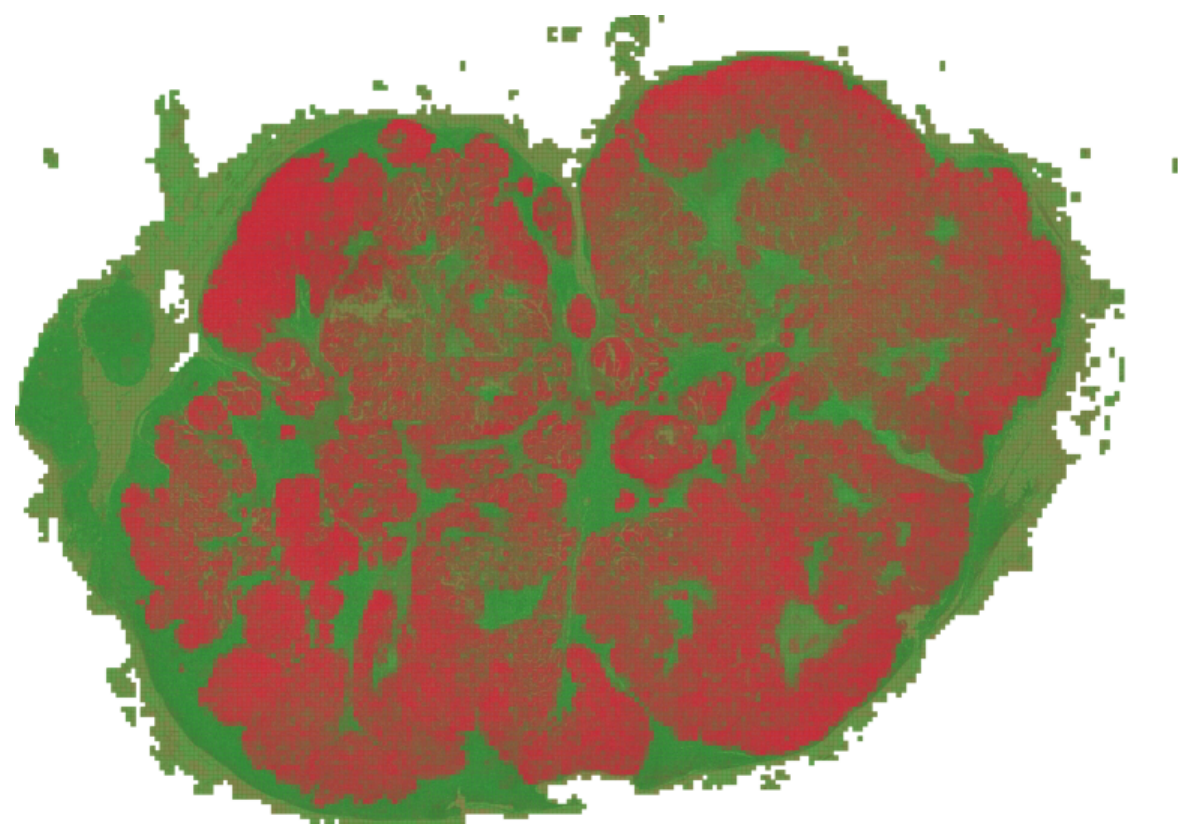}
            & 
            \includegraphics[trim={0cm 0cm 0cm 0cm},clip,width=0.22\textwidth]
            {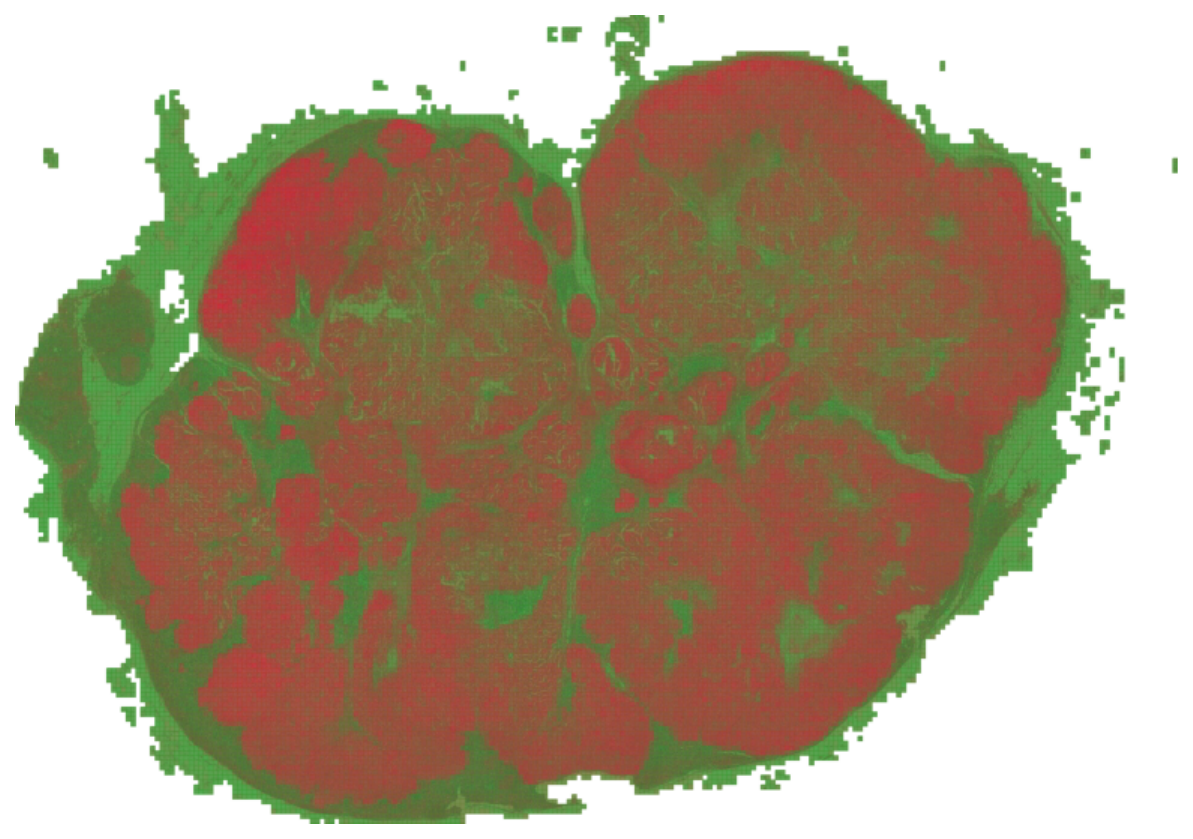}
            & 
            \includegraphics[trim={0cm 0cm 0cm 0cm},clip,width=0.22\textwidth]{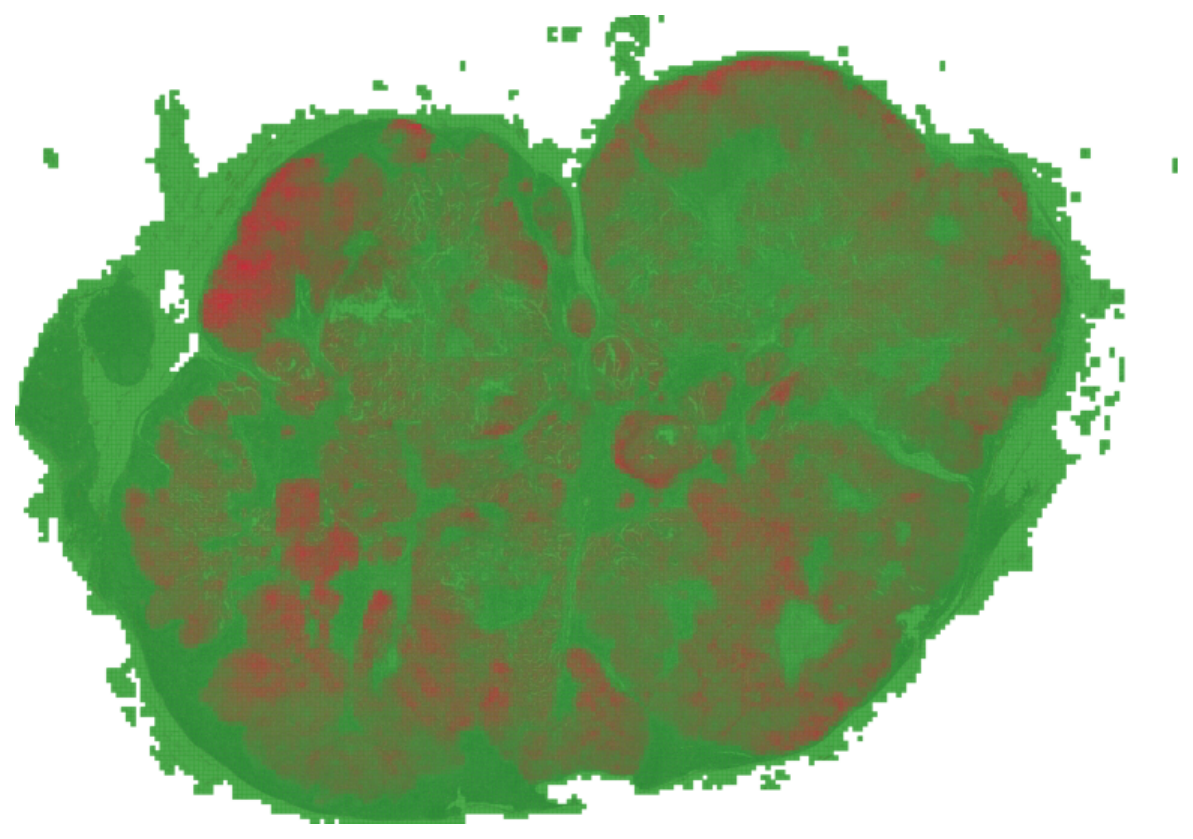}
            & 
            \includegraphics[trim={0cm 0cm 0cm 0cm},clip,width=0.22\textwidth]
            {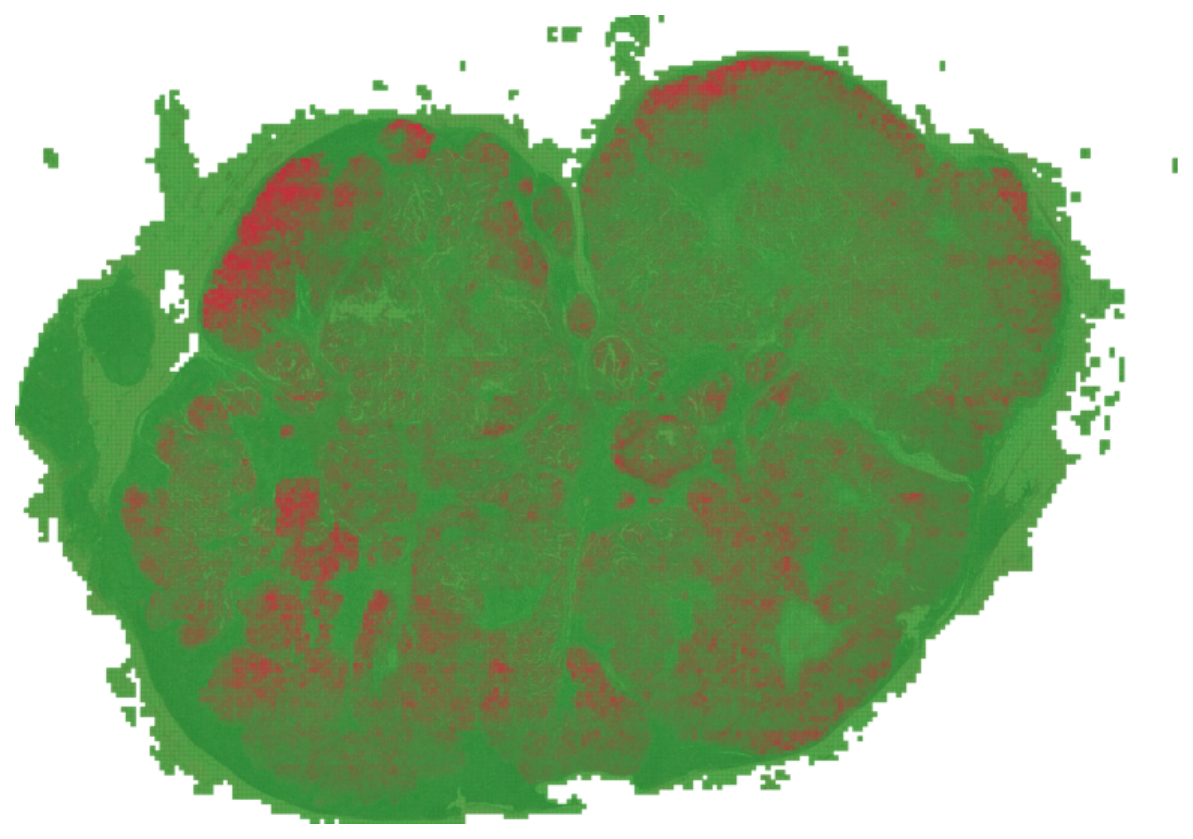} \\
            \pathgcn & \dtfdmil & \gtp & \camil
        \end{tabular}
        \end{adjustbox}
        \caption{
        Attention maps in a WSI from CAMELYON16. The attention values have been normalized to ease visualization.
        While none of the methods is able to fully localize the cancerous area, the proposed \probsmoothatt\ is the only capable of flagging its wrong predictions. 
        }
        \label{fig:attmaps-camelyon-main}
    \end{center}
    \vspace{-0.7cm}
\end{figure}

\begin{figure}[t!]
\footnotesize
	\begin{center}
		\centering
		\begin{adjustbox}{width=0.7\textwidth}
			\begin{tabular}{cc}
				\includegraphics[trim={0cm 0cm 0cm 0cm},clip,width=5cm]{img/att_map-bar_horizontal.pdf} 
				&
				\includegraphics[trim={0cm 0cm 0cm 0cm},clip,width=5cm]{img/uncert_map-bar_horizontal.pdf}
			\end{tabular}
		\end{adjustbox}
		\begin{adjustbox}{width=1.0\textwidth}
			\begin{tabular}{ccccc}
				\includegraphics[trim={0.9cm 0cm 0cm 0cm},clip,angle=90,width=0.22\textwidth]{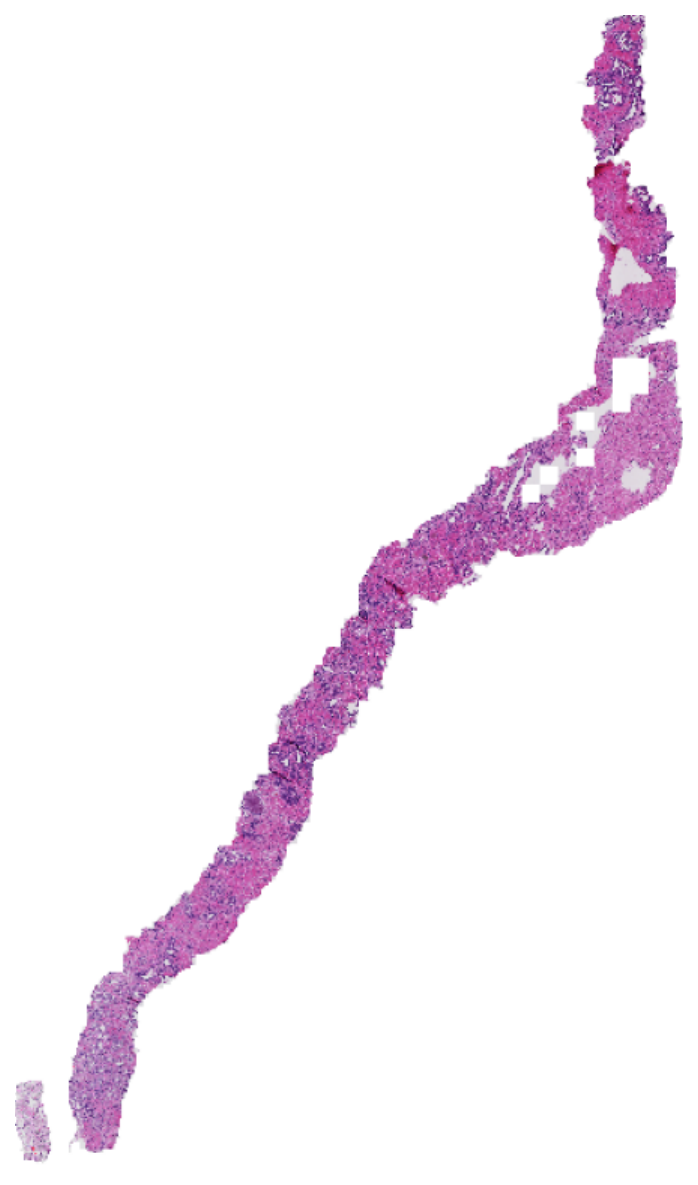}
				& 
				\includegraphics[trim={0.9cm 0cm 0cm 0cm},clip,angle=90,width=0.22\textwidth]
				{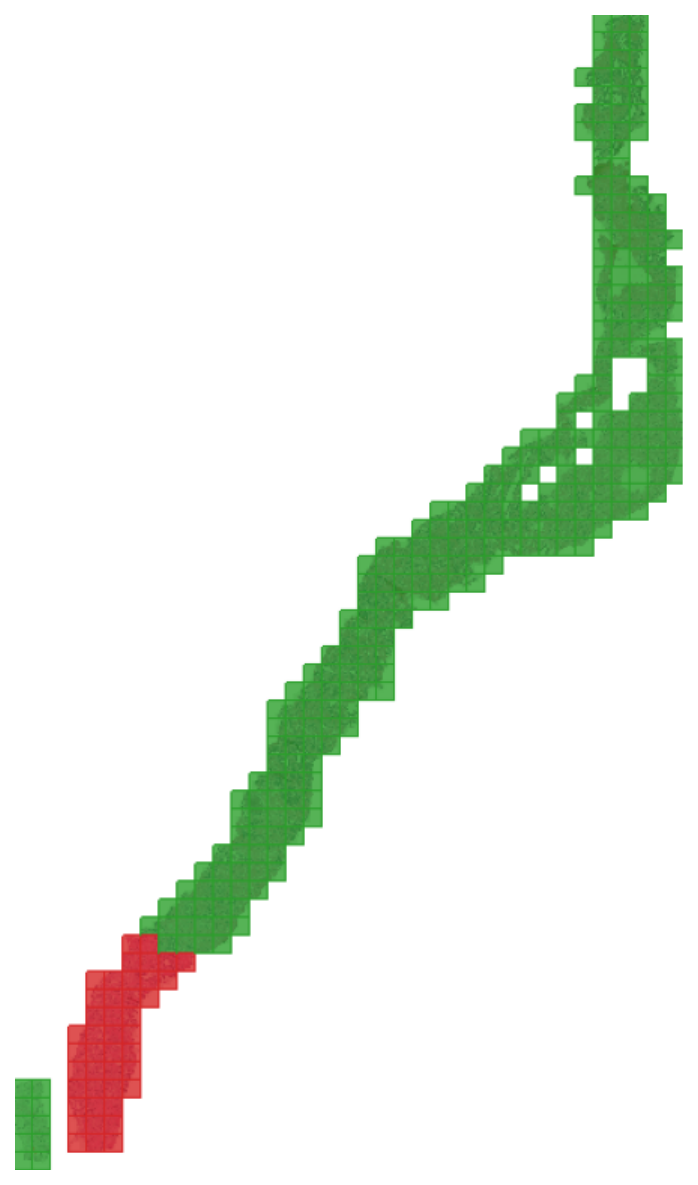}
				& 
				\includegraphics[trim={0.9cm 0cm 0cm 0cm},clip,angle=90,width=0.22\textwidth]
				{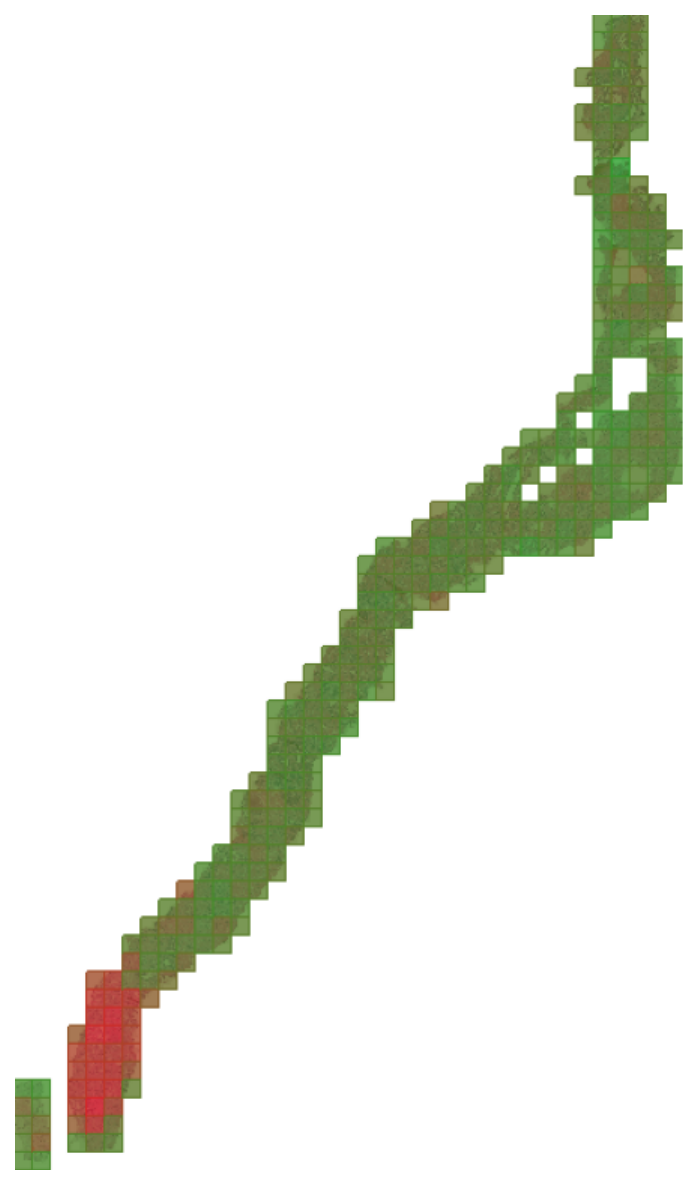}
				& 
				\includegraphics[trim={0.9cm 0cm 0cm 0cm},clip,angle=90,width=0.22\textwidth]
				{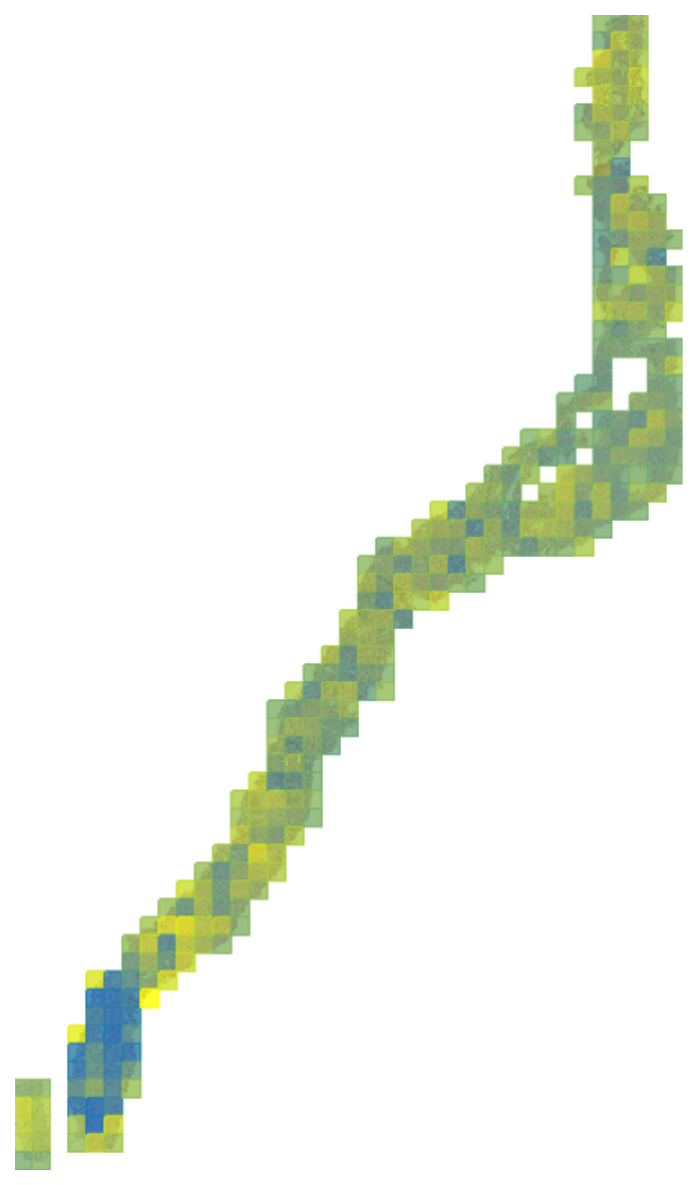}
				\\
				WSI &  Patch labels & \makecell{\abmil+\probsmoothatt \\ Mean} & \makecell{\abmil+\probsmoothatt \\ Variance} \\
				\includegraphics[trim={0.9cm 0cm 0cm 0cm},clip,angle=90,width=0.22\textwidth]
				{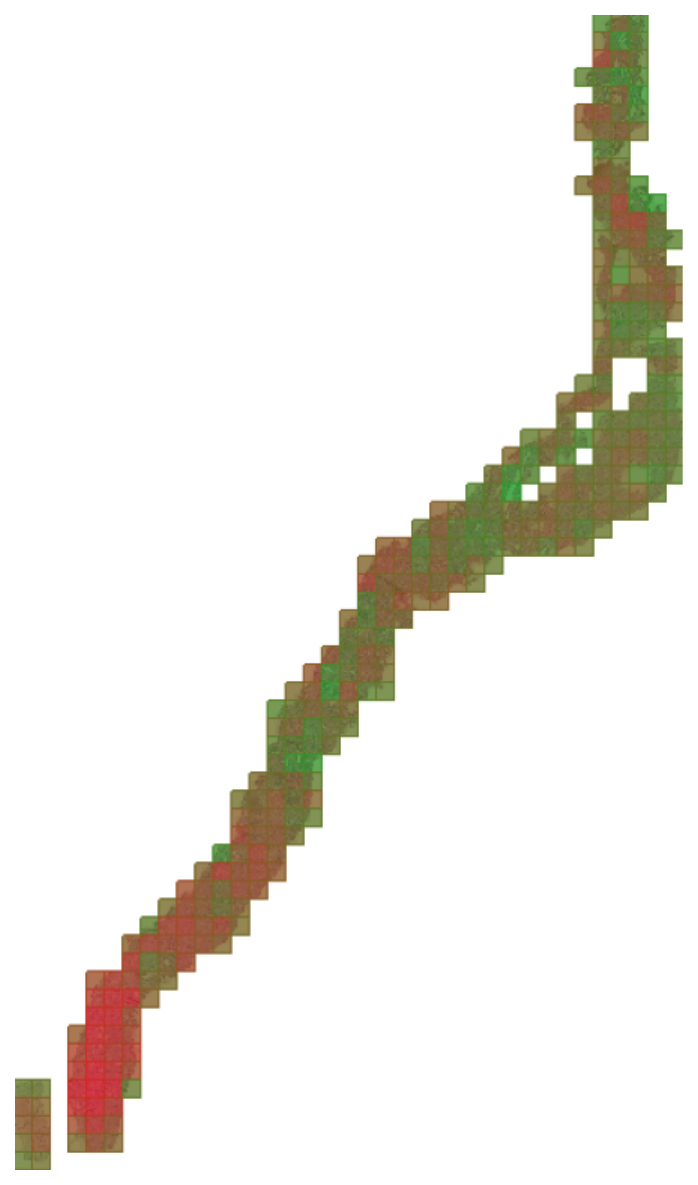}
				& 
				\includegraphics[trim={0.9cm 0cm 0cm 0cm},clip,angle=90,width=0.22\textwidth]
				{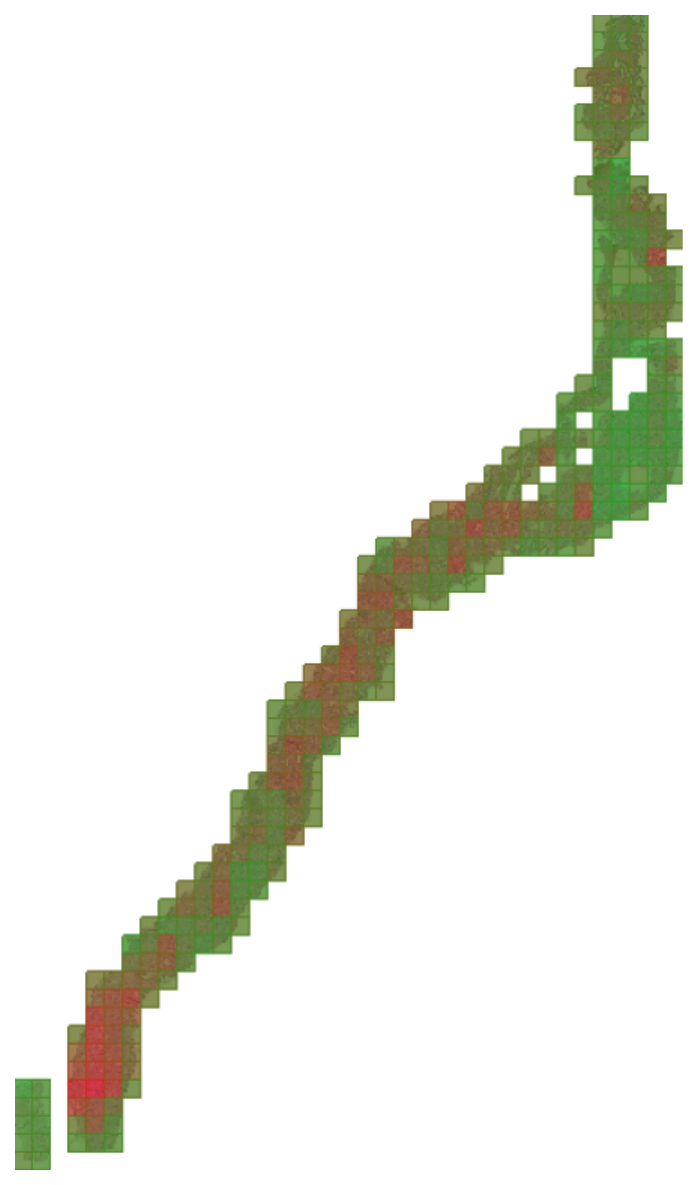}
				& 
				\includegraphics[trim={0.9cm 0cm 0cm 0cm},clip,angle=90,width=0.22\textwidth]{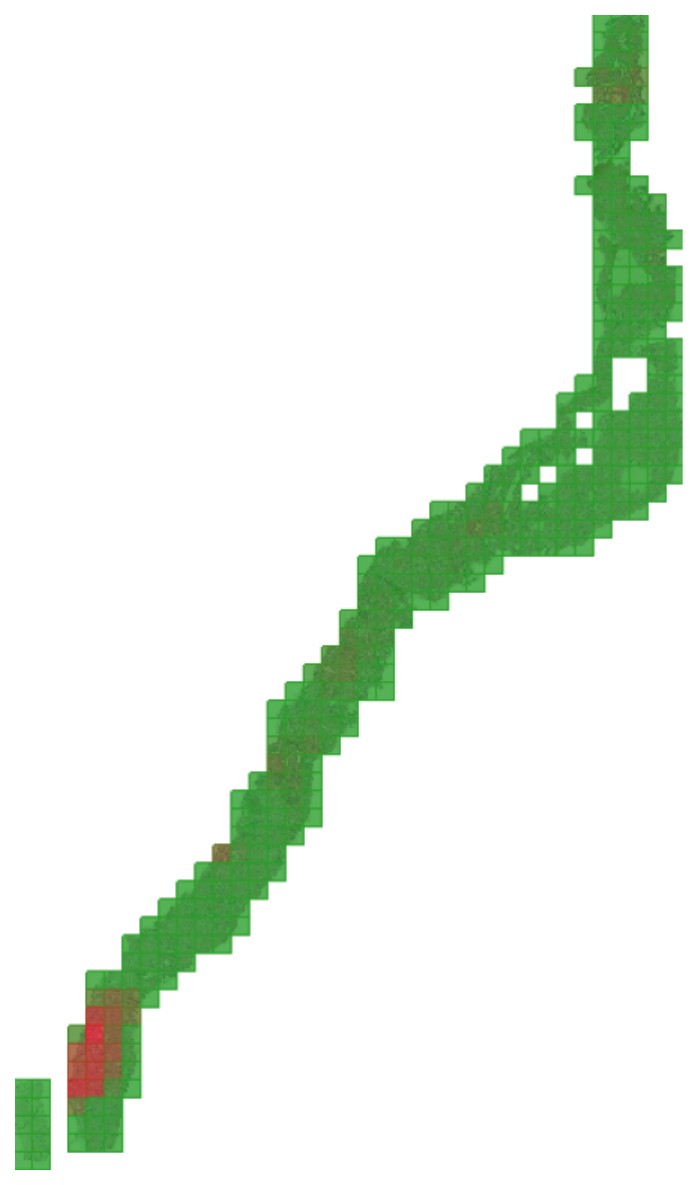}
				& 
				\includegraphics[trim={0.9cm 0cm 0cm 0cm},clip,angle=90,width=0.22\textwidth]
				{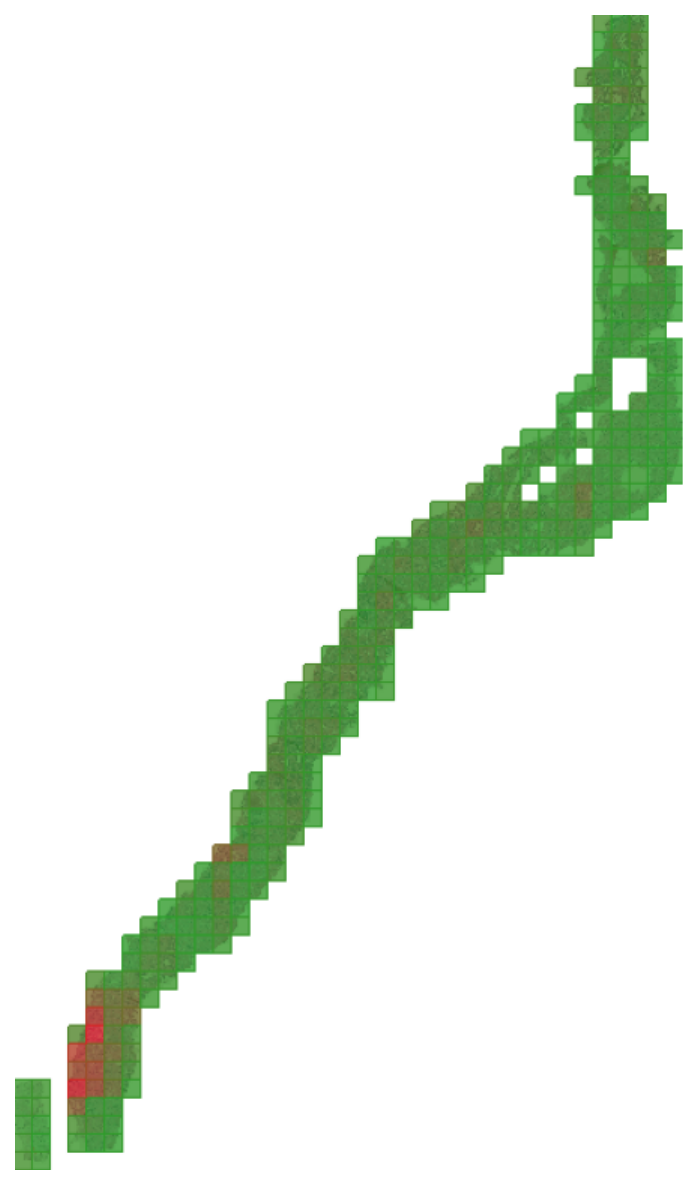} \\
            	\pathgcn & \dtfdmil & \gtp & \camil
			\end{tabular}
		\end{adjustbox}
		\caption{
			Attention maps in a WSI from PANDA. The attention values have been normalized to ease visualization. 
            The proposed \probsmoothatt\ produces fewer false positives than \abmil, \pathgcn, and \dtfdmil.
            Also, positive patches with low attention are flagged with high variance (see bottom of the WSI). 
            }
		\label{fig:attmaps-panda-main}
        \vspace{-0.5cm}
	\end{center}
    
\end{figure}

To show this, we analyse the attention maps produced by \probsmoothatt\ and compare them to those from other methods.
We display the attention maps of a CT scan from RSNA in \autoref{fig:attmaps-rsna-main}, a WSI from PANDA in \autoref{fig:attmaps-panda-main}, and a WSI from CAMELYON16 in \autoref{fig:attmaps-camelyon-main}.
Additional attention maps, including two more CT scans, two WSIs from PANDA, and two WSIs from CAMELYON16, are provided in \ref{app:additional_tables_figures}.
In total, we present attention maps for methods from each family across three representative positive bags from each dataset.
Notably, the instance labels displayed are used strictly for evaluation purposes and were not available during training.

\noindent 
\textbf{\probsmoothatt\ is aligned with previous works.} 
Previous works have shown that positive instances, i.e., those that are unhealthy, typically receive high attention values.
This enables one to locate areas containing hemorrhage (in the case of RSNA) or cancerous regions (in the case of PANDA and CAMELYON16).
\autoref{fig:attmaps-rsna-main}, \autoref{fig:attmaps-camelyon-main}, and \autoref{fig:attmaps-panda-main} show that the attention maps generated by \probsmoothatt\ also exhibit this desired pattern, demonstrating performance comparable to other leading methods, including \dtfdmil, \pathgcn, \gtp, and \camil. 

\noindent 
\textbf{\probsmoothatt\ reduces false positives.} 
{\color{review2}
Other methods tends to produce undesirable medium-to-high intensity isolated red spots in healthy areas, falsely indicating unhealthy instances.
A clear example of this phenomena is in \autoref{fig:attmaps-panda-main}, where \pathgcn\ and \dtfdmil\ assign high attention to healthy patches.
Other examples are \autoref{fig:attmaps-rsna-main}, \autoref{fig:attmaps-panda-a04310d441e8d2c7a5066627baeec9b6}, \autoref{fig:attmaps-camelyon-test_040}, and \autoref{fig:attmaps-camelyon-test_105}.
}
Note that these spots do not appear on the maps generated by \probsmoothatt. 
We attribute this behaviour to the training objective, which encourages the model to produce \textit{smooth} attention maps, minimizing isolated high-attention regions in healthy areas.

\noindent
\textbf{Attention variance flags wrong predictions.}
In practice, there are situations where the attention maps are not accurate. 
For example, in \autoref{fig:attmaps-camelyon-main} none of the considered methods completely detects the cancerous area. 
Unlike the rest of the methods, the novel probabilistic \probsmoothatt\ includes a mechanism to measure uncertainty, and one would expect that these wrong predictions are flagged with high variance.
Indeed, in \autoref{fig:attmaps-camelyon-main} we observe that \probsmoothatt\ assigns high variance to these incorrectly predicted instances. 
\autoref{fig:attmaps-panda-4d048dd585259614c27e11431ec2c860} and \autoref{fig:attmaps-rsna-ID_fdf596e74f} show a similar behaviour: as the other competing methods, \probsmoothatt\ wrongly assigns high attention values to healthy instances but, unlike those baselines, it flags out these instances through a high variance value.

\section{Conclusion and limitations}
\label{sec:conclusion}
In this work, we have introduced \probsmoothatt, a new attention-based deep MIL approach for medical imaging that accounts for both local and global interactions among instances in the bag. 
Local interactions, which refer to dependencies among neighbouring instances, are introduced by favouring smooth attention values within the bag.
Global interactions, which are longer-range dependencies, are modelled using a Transformer encoder. 
Moreover, we propose a probabilistic formulation that estimates attention values through a probability distribution, instead of deterministically through a single numeric value. 
To the best of our knowledge, this is the first method that tackles both types of interactions in a probabilistic manner.
We show enhanced predictive performance against current SOTA deep MIL approaches, as well as interpretability of the attention uncertainty maps provided by the probabilistic treatment.

There exist two main limitations to the proposed approach.
First, we have only analyzed two possibilities for the variational attention distribution: a Dirac delta, which leads to a deterministic treatment, and a Gaussian distribution with a diagonal covariance matrix, which can be treated through the re-parametrization trick and is computationally tractable.
However, more expressive variational distributions, possibly accounting for different modes, could be leveraged at a higher computational cost. 
Second, the proposed method does not solve the localization issues of current attention-based deep MIL approaches. 
Whereas the uncertainty maps are useful to highlight low-confidence regions, the mean of the attention does not accurately estimate the instance labels in every case, similar to existing methods.

{\color{review2}
In future work, alongside addressing the aforementioned limitations, we will explore alternative priors to enforce smoothness in the attention maps. 
However, other approaches different than the one used in this study may render an intractable KL divergence term, and thus they would require careful handling.
}

\section*{Acknowledgments}

This work was supported by project PID2022-140189OB-C22 funded by MCIN / AEI / 10.13039 / 501100011033. 
Francisco M. Castro-Macías acknowledges FPU contract FPU21/01874 funded by Ministerio de Universidades. 
Pablo Morales-Álvarez acknowledges grant C-EXP-153-UGR23 funded by Consejería de Universidad, Investigación e Innovación and by the European Union (EU) ERDF Andalusia Program 2021-2027.
Funding for open access charge: Universidad de Granada / CBUA.

\bibliographystyle{elsarticle-num-names} 
\bibliography{references}

\begin{thebibliography}{32}
\expandafter\ifx\csname natexlab\endcsname\relax\def\natexlab#1{#1}\fi
\providecommand{\url}[1]{\texttt{#1}}
\providecommand{\href}[2]{#2}
\providecommand{\path}[1]{#1}
\providecommand{\DOIprefix}{doi:}
\providecommand{\ArXivprefix}{arXiv:}
\providecommand{\URLprefix}{URL: }
\providecommand{\Pubmedprefix}{pmid:}
\providecommand{\doi}[1]{\href{http://dx.doi.org/#1}{\path{#1}}}
\providecommand{\Pubmed}[1]{\href{pmid:#1}{\path{#1}}}
\providecommand{\bibinfo}[2]{#2}
\ifx\xfnm\relax \def\xfnm[#1]{\unskip,\space#1}\fi
\bibitem[{Dietterich et~al.(1997)Dietterich, Lathrop, and
  Lozano-P{\'e}rez}]{dietterich1997solving}
\bibinfo{author}{T.~G. Dietterich}, \bibinfo{author}{R.~H. Lathrop},
  \bibinfo{author}{T.~Lozano-P{\'e}rez},
\newblock \bibinfo{title}{Solving the multiple instance problem with
  axis-parallel rectangles},
\newblock \bibinfo{journal}{Artificial intelligence}  (\bibinfo{year}{1997}).
\bibitem[{Song et~al.(2023)Song, Jaume, Williamson, Lu, Vaidya, Miller, and
  Mahmood}]{song2023artificial}
\bibinfo{author}{A.~H. Song}, \bibinfo{author}{G.~Jaume},
  \bibinfo{author}{D.~F. Williamson}, \bibinfo{author}{M.~Y. Lu},
  \bibinfo{author}{A.~Vaidya}, \bibinfo{author}{T.~R. Miller},
  \bibinfo{author}{F.~Mahmood},
\newblock \bibinfo{title}{Artificial intelligence for digital and computational
  pathology},
\newblock \bibinfo{journal}{Nature Reviews Bioengineering} \bibinfo{volume}{1}
  (\bibinfo{year}{2023}) \bibinfo{pages}{930--949}.
\bibitem[{Ilse et~al.(2018)Ilse, Tomczak, and Welling}]{ilse2018attention}
\bibinfo{author}{M.~Ilse}, \bibinfo{author}{J.~Tomczak},
  \bibinfo{author}{M.~Welling},
\newblock \bibinfo{title}{Attention-based deep multiple instance learning},
\newblock in: \bibinfo{booktitle}{International Conference on Machine
  Learning}, \bibinfo{year}{2018}.
\bibitem[{Lu et~al.(2021)Lu, Williamson, Chen, Chen, Barbieri, and
  Mahmood}]{lu2021data}
\bibinfo{author}{M.~Y. Lu}, \bibinfo{author}{D.~F. Williamson},
  \bibinfo{author}{T.~Y. Chen}, \bibinfo{author}{R.~J. Chen},
  \bibinfo{author}{M.~Barbieri}, \bibinfo{author}{F.~Mahmood},
\newblock \bibinfo{title}{Data-efficient and weakly supervised computational
  pathology on whole-slide images},
\newblock \bibinfo{journal}{Nature biomedical engineering}
  (\bibinfo{year}{2021}).
\bibitem[{Zhang et~al.(2022)Zhang, Meng, Zhao, Qiao, Yang, Coupland, and
  Zheng}]{zhang2022dtfd}
\bibinfo{author}{H.~Zhang}, \bibinfo{author}{Y.~Meng},
  \bibinfo{author}{Y.~Zhao}, \bibinfo{author}{Y.~Qiao},
  \bibinfo{author}{X.~Yang}, \bibinfo{author}{S.~E. Coupland},
  \bibinfo{author}{Y.~Zheng},
\newblock \bibinfo{title}{Dtfd-mil: Double-tier feature distillation multiple
  instance learning for histopathology whole slide image classification},
\newblock in: \bibinfo{booktitle}{IEEE/CVF Conference on Computer Vision and
  Pattern Recognition}, \bibinfo{year}{2022}, pp.
  \bibinfo{pages}{18802--18812}.
\bibitem[{Li et~al.(2021)Li, Li, and Eliceiri}]{li2021dual}
\bibinfo{author}{B.~Li}, \bibinfo{author}{Y.~Li}, \bibinfo{author}{K.~W.
  Eliceiri},
\newblock \bibinfo{title}{Dual-stream multiple instance learning network for
  whole slide image classification with self-supervised contrastive learning},
\newblock in: \bibinfo{booktitle}{Proceedings of the IEEE/CVF conference on
  computer vision and pattern recognition}, \bibinfo{year}{2021}, pp.
  \bibinfo{pages}{14318--14328}.
\bibitem[{Shao et~al.(2021)Shao, Bian, Chen, Wang, Zhang
  et~al.}]{shao2021transmil}
\bibinfo{author}{Z.~Shao}, \bibinfo{author}{H.~Bian},
  \bibinfo{author}{Y.~Chen}, \bibinfo{author}{Y.~Wang},
  \bibinfo{author}{J.~Zhang}, et~al.,
\newblock \bibinfo{title}{Transmil: Transformer based correlated multiple
  instance learning for whole slide image classification},
\newblock \bibinfo{journal}{Advances in neural information processing systems}
  (\bibinfo{year}{2021}).
\bibitem[{Li et~al.(2021)Li, Yang, Zhao, Xing, Zhang, Gao, Huang, Wang, and
  Yao}]{li2021dt}
\bibinfo{author}{H.~Li}, \bibinfo{author}{F.~Yang}, \bibinfo{author}{Y.~Zhao},
  \bibinfo{author}{X.~Xing}, \bibinfo{author}{J.~Zhang},
  \bibinfo{author}{M.~Gao}, \bibinfo{author}{J.~Huang},
  \bibinfo{author}{L.~Wang}, \bibinfo{author}{J.~Yao},
\newblock \bibinfo{title}{Dt-mil: deformable transformer for multi-instance
  learning on histopathological image},
\newblock in: \bibinfo{booktitle}{International Conference on Medical Image
  Computing and Computer Assisted Intervention},
  \bibinfo{organization}{Springer}, \bibinfo{year}{2021}.
\bibitem[{Fourkioti et~al.(2024)Fourkioti, Vries, and
  Bakal}]{fourkioti2023camil}
\bibinfo{author}{O.~Fourkioti}, \bibinfo{author}{M.~D. Vries},
  \bibinfo{author}{C.~Bakal},
\newblock \bibinfo{title}{{CAMIL}: Context-aware multiple instance learning for
  cancer detection and subtyping in whole slide images},
\newblock in: \bibinfo{booktitle}{International Conference on Learning
  Representations}, \bibinfo{year}{2024}.
\bibitem[{Yang et~al.(2024)Yang, Jiao, Wu, and Li}]{yang2024variational}
\bibinfo{author}{B.~Yang}, \bibinfo{author}{C.~Jiao}, \bibinfo{author}{J.~Wu},
  \bibinfo{author}{L.~Li},
\newblock \bibinfo{title}{Variational multiple-instance learning with embedding
  correlation modeling for hyperspectral target detection},
\newblock \bibinfo{journal}{IEEE Transactions on Neural Networks and Learning
  Systems}  (\bibinfo{year}{2024}).
\bibitem[{Zheng et~al.(2022)Zheng, Gindra, Green, Burks, Betke, Beane, and
  Kolachalama}]{zheng2022graph}
\bibinfo{author}{Y.~Zheng}, \bibinfo{author}{R.~H. Gindra},
  \bibinfo{author}{E.~J. Green}, \bibinfo{author}{E.~J. Burks},
  \bibinfo{author}{M.~Betke}, \bibinfo{author}{J.~E. Beane},
  \bibinfo{author}{V.~B. Kolachalama},
\newblock \bibinfo{title}{A graph-transformer for whole slide image
  classification},
\newblock \bibinfo{journal}{IEEE transactions on medical imaging}
  \bibinfo{volume}{41} (\bibinfo{year}{2022}).
\bibitem[{Zhao et~al.(2022)Zhao, Lin, Sun, Zhang, Huang, Wang, and
  Yao}]{zhao2022setmil}
\bibinfo{author}{Y.~Zhao}, \bibinfo{author}{Z.~Lin}, \bibinfo{author}{K.~Sun},
  \bibinfo{author}{Y.~Zhang}, \bibinfo{author}{J.~Huang},
  \bibinfo{author}{L.~Wang}, \bibinfo{author}{J.~Yao},
\newblock \bibinfo{title}{Setmil: spatial encoding transformer-based multiple
  instance learning for pathological image analysis},
\newblock in: \bibinfo{booktitle}{International Conference on Medical Image
  Computing and Computer-Assisted Intervention},
  \bibinfo{organization}{Springer}, \bibinfo{year}{2022}.
\bibitem[{Castro-Mac{\'\i}as et~al.(2024)Castro-Mac{\'\i}as, Morales-Alvarez,
  Wu, Molina, and Katsaggelos}]{castro2024sm}
\bibinfo{author}{F.~M. Castro-Mac{\'\i}as},
  \bibinfo{author}{P.~Morales-Alvarez}, \bibinfo{author}{Y.~Wu},
  \bibinfo{author}{R.~Molina}, \bibinfo{author}{A.~Katsaggelos},
\newblock \bibinfo{title}{Sm: enhanced localization in multiple instance
  learning for medical imaging classification},
\newblock in: \bibinfo{booktitle}{The Thirty-eighth Annual Conference on Neural
  Information Processing Systems}, \bibinfo{year}{2024}.
\bibitem[{Li et~al.(2018)Li, Yao, Zhu, Li, and Huang}]{li2018graph}
\bibinfo{author}{R.~Li}, \bibinfo{author}{J.~Yao}, \bibinfo{author}{X.~Zhu},
  \bibinfo{author}{Y.~Li}, \bibinfo{author}{J.~Huang},
\newblock \bibinfo{title}{Graph cnn for survival analysis on whole slide
  pathological images},
\newblock in: \bibinfo{booktitle}{International Conference on Medical Image
  Computing and Computer-Assisted Intervention},
  \bibinfo{organization}{Springer}, \bibinfo{year}{2018}.
\bibitem[{Chen et~al.(2021)Chen, Lu, Shaban, Chen et~al.}]{chen2021whole}
\bibinfo{author}{R.~J. Chen}, \bibinfo{author}{M.~Y. Lu},
  \bibinfo{author}{M.~Shaban}, \bibinfo{author}{C.~Chen}, et~al.,
\newblock \bibinfo{title}{Whole slide images are 2d point clouds: Context-aware
  survival prediction using patch-based graph convolutional networks},
\newblock in: \bibinfo{booktitle}{International Conference on Medical Image
  Computing and Computer-Assisted Intervention},
  \bibinfo{organization}{Springer}, \bibinfo{year}{2021}.
\bibitem[{Wu et~al.(2023)Wu, Castro-Mac{\'\i}as, Morales-{\'A}lvarez, Molina,
  and Katsaggelos}]{wu2023smooth}
\bibinfo{author}{Y.~Wu}, \bibinfo{author}{F.~M. Castro-Mac{\'\i}as},
  \bibinfo{author}{P.~Morales-{\'A}lvarez}, \bibinfo{author}{R.~Molina},
  \bibinfo{author}{A.~K. Katsaggelos},
\newblock \bibinfo{title}{Smooth attention for deep multiple instance learning:
  Application to ct intracranial hemorrhage detection},
\newblock in: \bibinfo{booktitle}{International Conference on Medical Image
  Computing and Computer-Assisted Intervention}, \bibinfo{year}{2023}.
\bibitem[{Zhou et~al.(2003)Zhou, Bousquet, Lal, Weston, and
  Sch{\"o}lkopf}]{zhou2003learning}
\bibinfo{author}{D.~Zhou}, \bibinfo{author}{O.~Bousquet},
  \bibinfo{author}{T.~Lal}, \bibinfo{author}{J.~Weston},
  \bibinfo{author}{B.~Sch{\"o}lkopf},
\newblock \bibinfo{title}{Learning with local and global consistency},
\newblock \bibinfo{journal}{Advances in neural information processing systems}
  (\bibinfo{year}{2003}).
\bibitem[{Ripley(1981)}]{ripley1981spatial}
\bibinfo{author}{B.~D. Ripley},
\newblock \bibinfo{title}{Spatial statistics}  (\bibinfo{year}{1981}).
\bibitem[{Bishop and Nasrabadi(2006)}]{bishop2006pattern}
\bibinfo{author}{C.~M. Bishop}, \bibinfo{author}{N.~M. Nasrabadi},
  \bibinfo{title}{Pattern recognition and machine learning},
  volume~\bibinfo{volume}{4}, \bibinfo{publisher}{Springer},
  \bibinfo{year}{2006}.
\bibitem[{Kingma(2013)}]{kingma2013auto}
\bibinfo{author}{D.~P. Kingma},
\newblock \bibinfo{title}{Auto-encoding variational bayes},
\newblock \bibinfo{journal}{arXiv preprint arXiv:1312.6114}
  (\bibinfo{year}{2013}).
\bibitem[{Arfken et~al.(2011)Arfken, Weber, and
  Harris}]{arfken2011mathematical}
\bibinfo{author}{G.~B. Arfken}, \bibinfo{author}{H.~J. Weber},
  \bibinfo{author}{F.~E. Harris}, \bibinfo{title}{Mathematical methods for
  physicists: a comprehensive guide}, \bibinfo{publisher}{Academic press},
  \bibinfo{year}{2011}.
\bibitem[{Folland(2009)}]{folland2009fourier}
\bibinfo{author}{G.~B. Folland}, \bibinfo{title}{Fourier analysis and its
  applications}, volume~\bibinfo{volume}{4}, \bibinfo{publisher}{American
  Mathematical Soc.}, \bibinfo{year}{2009}.
\bibitem[{Higgins et~al.(2017)Higgins, Matthey, Pal, Burgess, Glorot,
  Botvinick, Mohamed, and Lerchner}]{higgins2017beta}
\bibinfo{author}{I.~Higgins}, \bibinfo{author}{L.~Matthey},
  \bibinfo{author}{A.~Pal}, \bibinfo{author}{C.~P. Burgess},
  \bibinfo{author}{X.~Glorot}, \bibinfo{author}{M.~M. Botvinick},
  \bibinfo{author}{S.~Mohamed}, \bibinfo{author}{A.~Lerchner},
\newblock \bibinfo{title}{beta-vae: Learning basic visual concepts with a
  constrained variational framework.},
\newblock \bibinfo{journal}{ICLR} \bibinfo{volume}{3} (\bibinfo{year}{2017}).
\bibitem[{Fu et~al.(2019)Fu, Li, Liu, Gao, Celikyilmaz, and
  Carin}]{fu2019cyclical}
\bibinfo{author}{H.~Fu}, \bibinfo{author}{C.~Li}, \bibinfo{author}{X.~Liu},
  \bibinfo{author}{J.~Gao}, \bibinfo{author}{A.~Celikyilmaz},
  \bibinfo{author}{L.~Carin},
\newblock \bibinfo{title}{Cyclical annealing schedule: A simple approach to
  mitigating kl vanishing},
\newblock in: \bibinfo{booktitle}{Proceedings of the 2019 Conference of the
  North American Chapter of the Association for Computational Linguistics:
  Human Language Technologies}, \bibinfo{year}{2019}.
\bibitem[{Bishop and Bishop(2023)}]{bishop2024deep}
\bibinfo{author}{C.~M. Bishop}, \bibinfo{author}{H.~Bishop},
  \bibinfo{title}{Deep Learning: Foundations and Concepts},
  \bibinfo{year}{2023}.
\bibitem[{Flanders et~al.(2020)Flanders, Prevedello, Shih, Halabi,
  Kalpathy-Cramer, Ball et~al.}]{flanders2020construction}
\bibinfo{author}{A.~E. Flanders}, \bibinfo{author}{L.~M. Prevedello},
  \bibinfo{author}{G.~Shih}, \bibinfo{author}{S.~S. Halabi},
  \bibinfo{author}{J.~Kalpathy-Cramer}, \bibinfo{author}{R.~Ball}, et~al.,
\newblock \bibinfo{title}{Construction of a machine learning dataset through
  collaboration: the rsna 2019 brain ct hemorrhage challenge},
\newblock \bibinfo{journal}{Radiology: Artificial Intelligence}
  (\bibinfo{year}{2020}).
\bibitem[{Bulten et~al.(2022)Bulten, Kartasalo, Chen, Str{\"o}m, Pinckaers,
  Nagpal, Cai, Steiner et~al.}]{bulten2022artificial}
\bibinfo{author}{W.~Bulten}, \bibinfo{author}{K.~Kartasalo},
  \bibinfo{author}{P.-H.~C. Chen}, \bibinfo{author}{P.~Str{\"o}m},
  \bibinfo{author}{H.~Pinckaers}, \bibinfo{author}{K.~Nagpal},
  \bibinfo{author}{Y.~Cai}, \bibinfo{author}{D.~F. Steiner}, et~al.,
\newblock \bibinfo{title}{Artificial intelligence for diagnosis and gleason
  grading of prostate cancer: the panda challenge},
\newblock \bibinfo{journal}{Nature medicine}  (\bibinfo{year}{2022}).
\bibitem[{Bejnordi et~al.(2017)Bejnordi, Veta, Van~Diest, Van~Ginneken,
  Karssemeijer, Litjens, Van Der~Laak et~al.}]{bejnordi2017diagnostic}
\bibinfo{author}{B.~E. Bejnordi}, \bibinfo{author}{M.~Veta},
  \bibinfo{author}{P.~J. Van~Diest}, \bibinfo{author}{B.~Van~Ginneken},
  \bibinfo{author}{N.~Karssemeijer}, \bibinfo{author}{G.~Litjens},
  \bibinfo{author}{J.~A. Van Der~Laak}, et~al.,
\newblock \bibinfo{title}{Diagnostic assessment of deep learning algorithms for
  detection of lymph node metastases in women with breast cancer},
\newblock \bibinfo{journal}{Jama}  (\bibinfo{year}{2017}).
\bibitem[{Silva-Rodriguez et~al.(2021)Silva-Rodriguez, Colomer, Dolz, and
  Naranjo}]{silva2021self}
\bibinfo{author}{J.~Silva-Rodriguez}, \bibinfo{author}{A.~Colomer},
  \bibinfo{author}{J.~Dolz}, \bibinfo{author}{V.~Naranjo},
\newblock \bibinfo{title}{Self-learning for weakly supervised gleason grading
  of local patterns},
\newblock \bibinfo{journal}{IEEE journal of biomedical and health informatics}
  \bibinfo{volume}{25} (\bibinfo{year}{2021}) \bibinfo{pages}{3094--3104}.
\bibitem[{Kang et~al.(2023)Kang, Song, Park, Yoo, and
  Pereira}]{kang2023benchmarking}
\bibinfo{author}{M.~Kang}, \bibinfo{author}{H.~Song},
  \bibinfo{author}{S.~Park}, \bibinfo{author}{D.~Yoo},
  \bibinfo{author}{S.~Pereira},
\newblock \bibinfo{title}{Benchmarking self-supervised learning on diverse
  pathology datasets},
\newblock in: \bibinfo{booktitle}{IEEE/CVF Conference on Computer Vision and
  Pattern Recognition}, \bibinfo{year}{2023}.
\bibitem[{Wang et~al.(2024)Wang, Zhao, Marostica, Yuan, Jin, Zhang, Li, Tang,
  Wang, Li et~al.}]{wang2024pathology}
\bibinfo{author}{X.~Wang}, \bibinfo{author}{J.~Zhao},
  \bibinfo{author}{E.~Marostica}, \bibinfo{author}{W.~Yuan},
  \bibinfo{author}{J.~Jin}, \bibinfo{author}{J.~Zhang},
  \bibinfo{author}{R.~Li}, \bibinfo{author}{H.~Tang},
  \bibinfo{author}{K.~Wang}, \bibinfo{author}{Y.~Li}, et~al.,
\newblock \bibinfo{title}{A pathology foundation model for cancer diagnosis and
  prognosis prediction},
\newblock \bibinfo{journal}{Nature}  (\bibinfo{year}{2024})
  \bibinfo{pages}{1--9}.
\bibitem[{Ren et~al.(2023)Ren, Zhao, He, Wu, Mai et~al.}]{ren2023iib}
\bibinfo{author}{Q.~Ren}, \bibinfo{author}{Y.~Zhao}, \bibinfo{author}{B.~He},
  \bibinfo{author}{B.~Wu}, \bibinfo{author}{S.~Mai}, et~al.,
\newblock \bibinfo{title}{Iibmil: Integrated instance-level and bag-level
  multiple instances learning with label disambiguation for pathological image
  analysis},
\newblock in: \bibinfo{booktitle}{International Conference on Medical Image
  Computing and Computer-Assisted Intervention}, \bibinfo{year}{2023}.

\end{thebibliography}






\newpage

\appendix

\advance\voffset by -0.5cm 
\advance\footskip by 0.5cm 
\enlargethispage{1.\baselineskip} 

\section{Variational inference with a Dirac delta}
\label{app:dirac_delta}

We provide a derivation of the training objective for the deterministic variant of the proposed method, see \autoref{eq:elbo_diracdelta}. The fundamental idea here is to approximate the Dirac delta using a sequence of continuous densities of compact support. For simplicity, we assume bags of constant size $N \in \Nbb$. Our derivation is based on the following result.

\noindent
\textbf{Preliminary result.}
Let $u \colon \Rbb^{N} \to \Rbb$ be a continuous function of compact support such that $\int_{\Rbb^{N}} u \left( \bx \right) \dd \bx = 1$. For $\varepsilon > 0$, we define 
\begin{equation}\label{eq:u_epsilon}
    u_{\varepsilon} \left( \bx \right) = \varepsilon^{-N} u \left( \varepsilon^{-1} \bx \right), \quad \forall \bx \in \Rbb^N.
\end{equation}
Let $\ba \in \Rbb^N$. For any continuous function $g \colon \Rbb^{N} \to \Rbb$, it holds that 
\begin{equation}\label{eq:preliminary_result}
    \lim_{\varepsilon \to 0^+}\int_{\Rbb^{N}} u_{\varepsilon} \left( \bx - \ba \right) g \left( \bx \right) \dd \bx = g \left( \ba \right). 
\end{equation}
See \cite[Theorem 7.3]{folland2009fourier} for a proof. 
This expression corresponds to the definition of the Dirac delta using a (continuous and compactly supported) approximation to the identity \cite{folland2009fourier}. 
It is usually written informally as
\begin{equation}
    \int_{\Rbb^{N}} \delta \left( \bx - \ba \right) g \left( \bx \right) \dd \bx = g \left( \ba \right).
\end{equation}

\noindent
\textbf{Derivation of \autoref{eq:elbo_diracdelta}.}
We are now ready to derive \autoref{eq:elbo_diracdelta}. 
Let $\mu \colon \Rbb^{N \times D} \to \Rbb^{N}$ a neural network parameterizing the attention values. 
Let $u_{\varepsilon}$ defined as in \autoref{eq:u_epsilon}. We consider a family $\left\{  \q_{\varepsilon} \right\}_{\varepsilon > 0}$ of variational posteriors defined as
\begin{equation}
    \q_{\varepsilon} \left( \bff \mid \bX \right) = u_{\varepsilon} \left( \bff - \mu\left( \bX \right)\right).
\end{equation}
Intuitively, $\left\{  \q_{\varepsilon} \right\}_{\varepsilon > 0}$ converges to the variational posterior defined using the Dirac delta in \autoref{eq:varposterior_diracdelta}.
Let us write the ELBO for this family,
\begin{align}
    \operatorname{ELBO}_{\varepsilon} = & \sum_{b=1}^{B} \{ \Ebb_{ \q_{\varepsilon} \left( \bff_b \mid \bX_b \right) } \left[  \log \p(Y_b \mid \bX_b, \bff_b) \right] +  \Ebb_{ \q_{\varepsilon} \left( \bff_b \mid \bX_b \right) } \left[ \log \p \left( \bff_b \right) \right] + \\
    & - \Ebb_{ \q_{\varepsilon} \left( \bff_b \mid \bX_b \right) } \left[ \log \q_{\varepsilon} \left( \bff_b \mid \bX_b \right) \right] \} 
\end{align}
We are interested in optimizing this ELBO with respect to the parameters of our neural network in the limit $\varepsilon \to 0^+$. For the last term, after a change of variable, we have
\begin{equation}
    \Ebb_{ \q_{\varepsilon} \left( \bff_b \mid \bX_b \right) } \left[ \log \q_{\varepsilon} \left( \bff_b \mid \bX_b \right) \right] \} = \int_{\Rbb^N} u_{\varepsilon}\left( \bx \right) \log u_{\varepsilon}\left( \bx \right) \dd \bx.
\end{equation}
Note that this term does not depend on the parameters of the neural network, and is well defined for $\varepsilon > 0$. 
However, its limit when $\varepsilon \to 0^+$ may not be well defined. 
This is not a problem for us, since we are only interested in those terms that depend on the parameters of the neural network, i.e., 
\begin{equation}
    \widetilde{\operatorname{ELBO}}_{\varepsilon} = \sum_{b=1}^{B} \left\{ \Ebb_{ \q_{\varepsilon} \left( \bff_b \mid \bX_b \right) } \left[  \log \p(Y_b \mid \bX_b, \bff_b) \right] +  \Ebb_{ \q_{\varepsilon} \left( \bff_b \mid \bX_b \right) } \left[ \log \p \left( \bff_b \right) \right] \right\}.
\end{equation}
Finally, we take the limit when $\varepsilon \to 0^+$ and apply the preliminary result in \autoref{eq:preliminary_result}, obtaining the desired optimization objective, see \autoref{eq:elbo_diracdelta}.

\section{Transformer encoder}
\label{app:transformer_encoder}

We provide more details about the Transformer encoder introduced in \autoref{subsec:global_interactions}. 
At the core of this module is the self-attention mechanism, which is given an input bag $\bX \in \Rbb^{N \times D}$ and outputs a new bag $\bY = \operatorname{SelfAttention} \left( \bX \right) \in \Rbb^{N \times D_v}$ transformed according to
\begin{equation}
    \operatorname{SelfAttention} \left( \bX \right) = \softmax\left( q\left( \bX \right) k\left( \bX \right)^\top \right) v\left( \bX \right),
\end{equation}
where $q, k \colon \Rbb^{\Nbb \times D} \to \Rbb^{\Nbb \times D_{qk}}$, and $v \colon \Rbb^{\Nbb \times D} \to \Rbb^{\Nbb \times D_{v}}$ are the standard queries, keys, and values in the dot product self-attention \citep{bishop2024deep}. Each layer of the Transformer encoder leverages multiple heads in the self-attention mechanism, along with skip-connections and pre-layer normalization \citep{bishop2024deep},  
\begin{gather}
    \operatorname{TransformerEnc}(\bX) = \bY^L, \\
    \bY^0 = \bX, \\
    \bZ^{l+1} = \bY^l + \operatorname{SelfAttention} \left( \operatorname{LayerNorm}\left( \bY^l \right) \right), \\
    \bY^{l+1} = \bZ^{l+1} + \operatorname{MLP}\left( \operatorname{LayerNorm}\left( \bZ^{l+1} \right) \right),
\end{gather}
where $l \in \left\{ 0, \ldots, L-1 \right\}$, and $\operatorname{MLP}$ denotes a two-layer perceptron. 
As explained in \autoref{sec:experiments}, in this work we use $D_{qk}=D_{v}=512$, $L=2$ and 8 attention heads.




\newpage
\advance\voffset by -1.5cm 
\advance\footskip by 2.1cm 


\section{Additional tables and figures}
\label{app:additional_tables_figures}

{\color{review}
\noindent
\textbf{Computational overhead.} \probsmoothatt\ incorporates local interactions into the loss function through the bag adjacency matrices. 
These are instantiated as sparse tensors, reducing memory usage and enabling efficient matrix-vector operations.
As shown in~\autoref{tab:training_time}, the overhead is comparable to that of other widely used methods. 
Relative to the baseline on top of which they are implemented, training time increases slightly on RSNA and PANDA, and more significantly on CAMELYON16, due to the larger size of its WSIs. Nonetheless, the overhead remains manageable, and our method scales effectively to large datasets such as CAMELYON16.
}

\begin{table}[h]
\centering
\begin{adjustbox}{width=0.48\textwidth}
\begin{tabular}{@{}cccc@{}}
\toprule
 & \multicolumn{3}{c}{Time per step (ms)} \\

Model & RSNA & PANDA & CAMELYON16 \\
\midrule
\abmil+\probsmooth($\Sigma=0$) & $2.769 \pm 0.280$ & $2.966 \pm 0.992$ & $15.334 \pm 13.077$ \\
\abmil+\probsmooth($\Sigma=\operatorname{Diag}$) & $3.776 \pm 0.377$ & $3.821 \pm 1.315$ & $16.008 \pm 12.469$ \\
\abmil & $0.997 \pm 0.132$ & $1.035 \pm 0.129$ & $2.094 \pm 6.053$ \\
\clam & $3.832 \pm 0.407$ & $4.896 \pm 4.035$ & $5.401 \pm 7.563$ \\
\dsmil & $1.692 \pm 0.275$ & $2.468 \pm 3.280$ & $3.051 \pm 5.238$ \\
\pathgcn & $2.356 \pm 0.215$ & $3.132 \pm 3.636$ & $5.790 \pm 12.516$ \\
\dtfdmil & $5.010 \pm 0.364$ & $6.373 \pm 4.265$ & $7.434 \pm 13.055$ \\
\bottomrule
\end{tabular}
\end{adjustbox}
\quad 
\begin{adjustbox}{width=0.48\textwidth}
	\begin{tabular}{@{}cccc@{}}
		\toprule
		& \multicolumn{3}{c}{Time per step (ms)} \\
		Model & RSNA & PANDA & CAMELYON16 \\
		\midrule
		\transformerabmil+\probsmooth($\Sigma=0$) & $3.374 \pm 0.315$ & $3.562 \pm 0.853$ & $47.609 \pm 52.472$ \\
		\transformerabmil+\probsmooth($\Sigma=\operatorname{Diag}$) & $4.49 \pm 0.452$ & $4.091 \pm 1.118$ & $49.172 \pm 53.664$ \\
		\transformerabmil & $2.182 \pm 0.196$ & $2.169 \pm 0.201$ & $3.65 \pm 0.805$ \\
		\transmil & $10.596 \pm 0.643$ & $11.153 \pm 0.745$ & $13.960 \pm 4.380$ \\
		\setmil & $24.321 \pm 1.528$ & $36.289 \pm 2.016$ & $307.732 \pm 103.096$ \\
		\gtp & $6.501 \pm 0.311$ & $8.059 \pm 0.384$ & $11.794 \pm 8.766$ \\
		\iibmil & $8.718 \pm 0.566$ & $19.087 \pm 1.722$ & $61.792 \pm 51.316$ \\
		\camil & $6.172 \pm 0.374$ & $6.766 \pm 0.883$ & $8.685 \pm 9.011$ \\
		\vmil & $12.231 \pm 0.728$ & $16.880 \pm 0.419$ & $58.387 \pm 2.148$ \\
		\bottomrule
	\end{tabular}
\end{adjustbox}
\caption{\color{review}
    Time per training step (in milliseconds). A training step is defined as a single forward-backward pass.   
}
\label{tab:training_time}
\vspace{-0.7cm}
\end{table}

\begin{table}[h]
\centering
\begin{adjustbox}{width=\textwidth}
\begin{tabular}{@{}ccccccccc@{}}
\toprule
& & \multicolumn{2}{c}{\textbf{RSNA}} & \multicolumn{2}{c}{\textbf{PANDA}} & \multicolumn{2}{c}{\textbf{CAMELYON16}} \\ \midrule
Model & $\lambda$ & AUROC $(\uparrow)$ & F1 $(\uparrow)$ & AUROC $(\uparrow)$ & F1 $(\uparrow)$ & AUROC $(\uparrow)$ & F1 $(\uparrow)$ & Rank $(\downarrow)$ \\ \midrule
\multirow{5}{*}{\makecell{\abmil+\probsmooth\\$\Sigma = 0$}} & 0.0 & $87.924_{1.179}$ & $78.317_{2.531}$ & $97.843_{0.201}$ & $95.060_{0.435}$ & $97.454_{1.074}$ & $90.756_{2.483}$ & $4.000_{1.265}$ \\
& 0.1 & $88.631_{0.864}$ & $79.479_{2.294}$ & $\mathbf{98.190}_{0.045}$ & $\mathbf{95.398}_{0.252}$ & $\mathbf{98.628}_{0.468}$ & $91.122_{1.762}$ & $2.500_{1.643}$ \\
& 0.5 & $89.384_{0.612}$ & $81.167_{1.951}$ & $97.557_{0.114}$ & $94.451_{0.299}$ & $97.459_{1.135}$ & $92.113_{1.941}$ & $3.167_{0.753}$ \\
& 1.0 & $89.950_{0.990}$ & $82.509_{2.572}$ & $97.146_{0.178}$ & $93.865_{0.254}$ & $96.888_{1.555}$ & $\mathbf{92.360}_{1.648}$ & $3.333_{1.862}$ \\
& cyclical & $\mathbf{90.189}_{0.482}$ & $\mathbf{83.168}_{1.259}$ & $97.896_{0.090}$ & $94.973_{0.140}$ & $97.633_{0.618}$ & $91.307_{1.053}$ & $\mathbf{2.000}_{0.894}$ \\
\midrule
\multirow{5}{*}{\makecell{\abmil+\probsmooth\\$\Sigma = \operatorname{Diag}$}} & 0.0 & $87.924_{1.179}$ & $78.317_{2.531}$ & $97.843_{0.201}$ & $95.060_{0.435}$ & $97.454_{1.074}$ & $90.756_{2.483}$ & $4.000_{1.265}$ \\
& 0.1 & $88.957_{0.575}$ & $80.409_{1.211}$ & $\mathbf{98.178}_{0.048}$ & $\mathbf{95.559}_{0.184}$ & $\mathbf{98.571}_{0.389}$ & $92.176_{1.347}$ & $2.167_{1.472}$ \\
& 0.5 & $90.089_{1.029}$ & $82.297_{1.947}$ & $97.664_{0.067}$ & $94.643_{0.184}$ & $97.173_{1.884}$ & $91.452_{2.825}$ & $3.667_{0.516}$ \\
& 1.0 & $\mathbf{90.666}_{0.877}$ & $82.820_{1.461}$ & $97.121_{0.138}$ & $93.836_{0.242}$ & $96.949_{1.111}$ & $91.659_{1.263}$ & $3.500_{1.761}$ \\
& cyclical & $90.231_{0.401}$ & $\mathbf{83.230}_{1.719}$ & $98.094_{0.069}$ & $95.274_{0.236}$ & $97.885_{1.259}$ & $\mathbf{92.351}_{2.385}$ & $\mathbf{1.667}_{0.516}$ \\
\midrule
\multirow{5}{*}{\makecell{\transformerabmil+\probsmooth\\$\Sigma = 0$}} & 0.0 & $91.083_{0.978}$ & $84.083_{1.962}$ & $98.014_{0.226}$ & $95.289_{0.322}$ & $97.673_{0.695}$ & $91.826_{3.638}$ & $3.500_{1.225}$ \\
& 0.1 & $91.141_{1.023}$ & $83.023_{2.429}$ & $98.046_{0.176}$ & $95.378_{0.190}$ & $98.017_{0.416}$ & $93.193_{3.426}$ & $2.833_{1.329}$ \\
& 0.5 & $91.079_{1.171}$ & $83.469_{2.720}$ & $\mathbf{98.064}_{0.135}$ & $\mathbf{95.457}_{0.115}$ & $98.304_{0.412}$ & $92.676_{0.632}$ & $2.667_{1.506}$ \\
& 1.0 & $91.072_{1.113}$ & $83.580_{2.784}$ & $98.007_{0.146}$ & $95.254_{0.305}$ & $98.068_{1.220}$ & $93.131_{1.753}$ & $3.667_{0.816}$ \\
& cyclical & $\mathbf{91.781}_{1.200}$ & $\mathbf{84.591}_{2.535}$ & $97.974_{0.156}$ & $95.213_{0.072}$ & $\mathbf{98.418}_{1.104}$ & $\mathbf{94.220}_{2.168}$ & $\mathbf{2.333}_{2.066}$ \\
\midrule
\multirow{5}{*}{\makecell{\transformerabmil+\probsmooth\\$\Sigma = \operatorname{Diag}$}} & 0.0 & $\mathbf{91.083}_{0.978}$ & $\mathbf{84.083}_{1.962}$ & $\mathbf{98.014}_{0.226}$ & $95.289_{0.322}$ & $97.673_{0.695}$ & $91.826_{3.638}$ & $2.500_{1.975}$ \\
& 0.1 & $90.812_{1.065}$ & $83.260_{2.006}$ & $97.928_{0.109}$ & $95.211_{0.216}$ & $97.934_{0.572}$ & $93.834_{1.816}$ & $3.833_{0.753}$ \\
& 0.5 & $90.808_{1.015}$ & $83.226_{2.144}$ & $97.974_{0.187}$ & $95.266_{0.194}$ & $\mathbf{98.486}_{1.147}$ & $94.657_{2.036}$ & $3.000_{1.414}$ \\
& 1.0 & $90.883_{1.043}$ & $82.964_{2.064}$ & $97.929_{0.199}$ & $95.198_{0.325}$ & $98.061_{1.352}$ & $92.653_{2.638}$ & $3.833_{1.169}$ \\
& cyclical & $90.814_{1.683}$ & $83.683_{3.669}$ & $97.988_{0.117}$ & $\mathbf{95.339}_{0.249}$ & $98.133_{0.828}$ & $\mathbf{94.732}_{3.071}$ & $\mathbf{1.833}_{0.753}$ \\
\bottomrule
\end{tabular}
\end{adjustbox}
\caption{
    Classification results (mean and standard deviation from five independent runs) using different values for the balancing hyperparameter $\lambda$. 
    The best in each group is in bold.
    $(\downarrow)$/$(\uparrow)$ means lower/higher is better.
    The optimal choice of $\lambda$ depends on the variant and the dataset.
    Setting $\lambda = \cyclical$ provides the best rank for every variant.
}
\label{tab:ablation_results}
\vspace{-4cm}
\end{table}

\newpage
\advance\footskip by -0.7cm 

\begin{figure}[h]
    \scriptsize
	\begin{center}
		\centering
		\begin{adjustbox}{width=0.7\textwidth}
			\begin{tabular}{cc}
				\includegraphics[trim={0cm 0cm 0cm 0cm},clip,width=5cm]{img/att_map-bar_horizontal.pdf} 
				&
				\includegraphics[trim={0cm 0cm 0cm 0cm},clip,width=5cm]{img/uncert_map-bar_horizontal.pdf}
			\end{tabular}
		\end{adjustbox}
		\begin{adjustbox}{width=1.0\textwidth}
			\begin{tabular}{ccccccc}
				\includegraphics[trim={0.9cm 0cm 0cm 0cm},clip,height=3cm]{img/panda/wsis/wsi-89bdb0f20bb2e4b15583af626d26ad62.pdf}
				& 
				\includegraphics[trim={0.9cm 0cm 0cm 0cm},clip,height=3cm]
				{img/panda/gt/gt-89bdb0f20bb2e4b15583af626d26ad62.pdf}
				& 
				\includegraphics[trim={0.95cm 0cm 0cm 0cm},clip,height=3cm]
				{img/panda/att_maps/att_map-bayes_smooth_abmil_diag-89bdb0f20bb2e4b15583af626d26ad62.pdf}
				& 
				\includegraphics[trim={0.95cm 0cm 0cm 0cm},clip,height=3cm]
				{img/panda/uncert_maps/uncert_map-bayes_smooth_abmil_diag-89bdb0f20bb2e4b15583af626d26ad62.pdf}
                &
                \includegraphics[trim={0.95cm 0cm 0cm 0cm},clip,height=3cm]
				{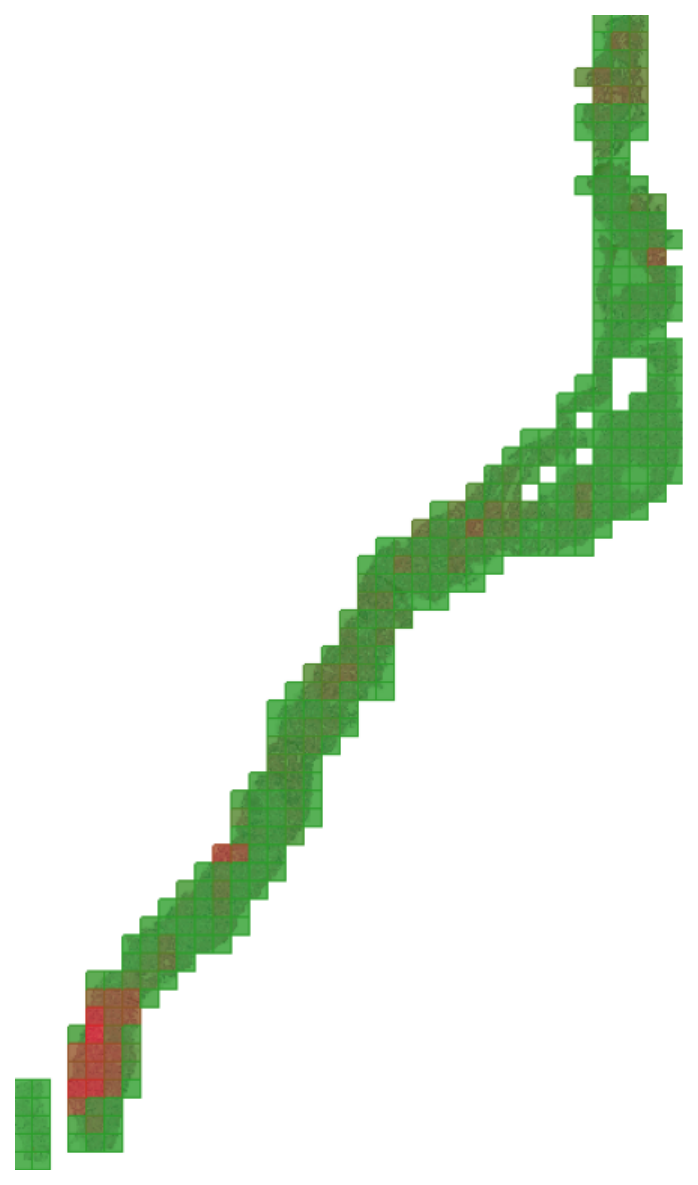}
				& 
				\includegraphics[trim={0.95cm 0cm 0cm 0cm},clip,height=3cm]
				{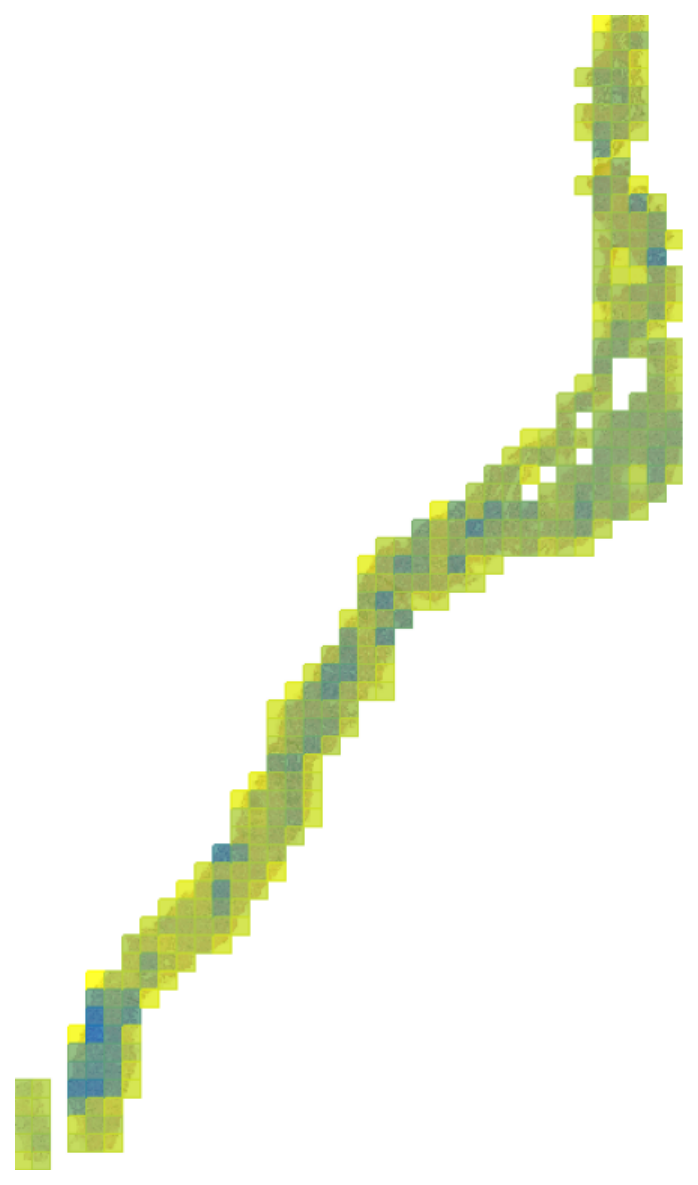}
                \\
                WSI &  Patch labels & \makecell{\abmil+PSA \\ Mean} & \makecell{\abmil+PSA \\ Variance} & \makecell{\transformerabmil+PSA \\ Mean} & \makecell{\transformerabmil+PSA \\ Variance} \\
                \includegraphics[trim={0.95cm 0cm 0cm 0cm},clip,height=3cm]
				{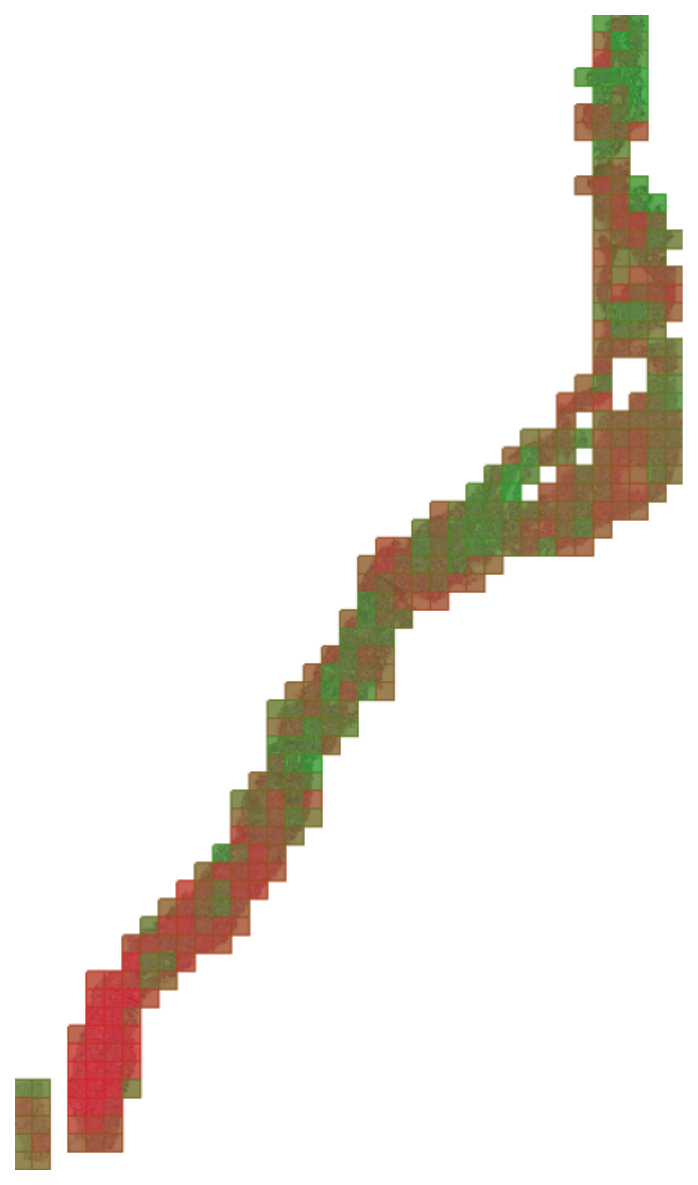}
				& 
				\includegraphics[trim={0.95cm 0cm 0cm 0cm},clip,height=3cm]
				{img/panda/att_maps/att_map-pathgcn-89bdb0f20bb2e4b15583af626d26ad62.pdf}
                &
				\includegraphics[trim={0.95cm 0cm 0cm 0cm},clip,height=3cm]{img/panda/att_maps/att_map-dftdmil-89bdb0f20bb2e4b15583af626d26ad62.pdf}
                &
                \includegraphics[trim={0.95cm 0cm 0cm 0cm},clip,height=3cm]
				{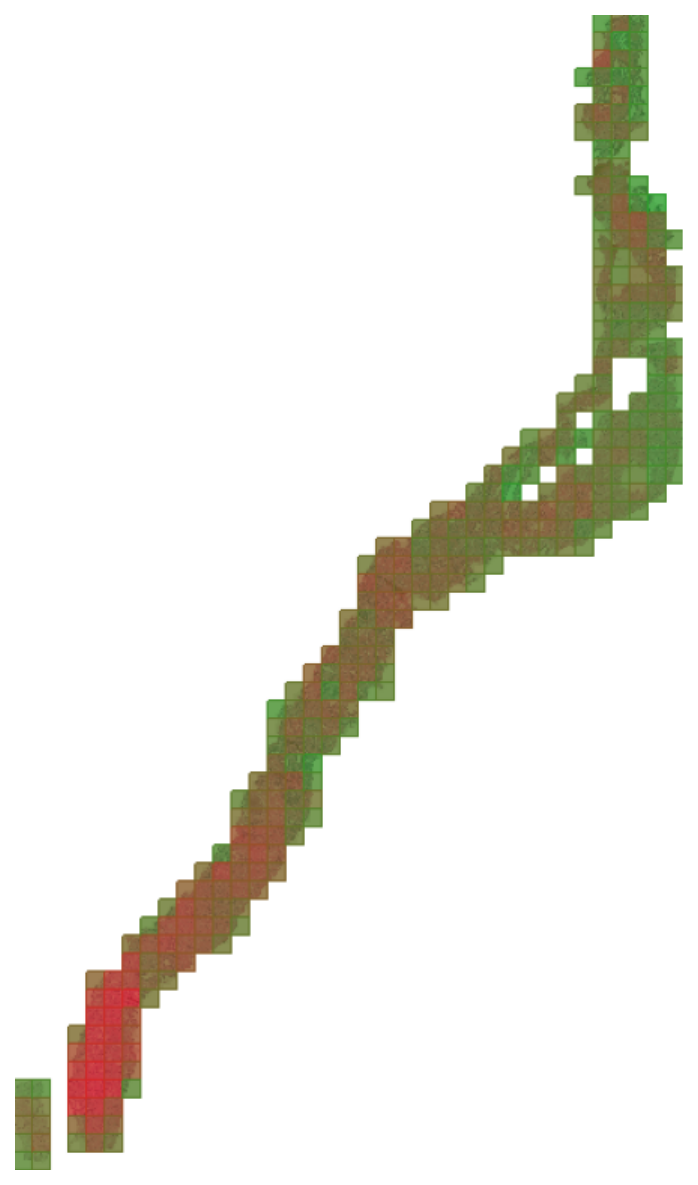}
                &
				\includegraphics[trim={0.95cm 0cm 0cm 0cm},clip,height=3cm]
				{img/panda/att_maps/att_map-camil-89bdb0f20bb2e4b15583af626d26ad62.pdf}
				& 
				\includegraphics[trim={0.95cm 0cm 0cm 0cm},clip,height=3cm]
				{img/panda/att_maps/att_map-gtp-89bdb0f20bb2e4b15583af626d26ad62.pdf}
				\\        
				\abmil & \pathgcn & \dtfdmil & \transformerabmil & \camil & \gtp \\
			\end{tabular}
		\end{adjustbox}
		\caption{
			Attention maps in a WSI from PANDA. The attention values have been normalized to ease visualization. PSA stands for \probsmoothatt. 
		}
		\label{fig:attmaps-panda-89bdb0f20bb2e4b15583af626d26ad62}
        \vspace{-0.5cm}
	\end{center}
\end{figure}



\begin{figure}[h!]
    \scriptsize
    \begin{center}
        \centering
        \begin{adjustbox}{width=0.7\textwidth}
        \begin{tabular}{cc}
             \includegraphics[trim={0cm 0cm 0cm 0cm},clip,width=5cm]{img/att_map-bar_horizontal.pdf} 
             &
             \includegraphics[trim={0cm 0cm 0cm 0cm},clip,width=5cm]{img/uncert_map-bar_horizontal.pdf}
        \end{tabular}
        \end{adjustbox}
        \begin{adjustbox}{width=1.0\textwidth}
            \begin{tabular}{ccccccc}
				\includegraphics[trim={0.0cm 0cm 0cm 0cm},clip,height=3cm]{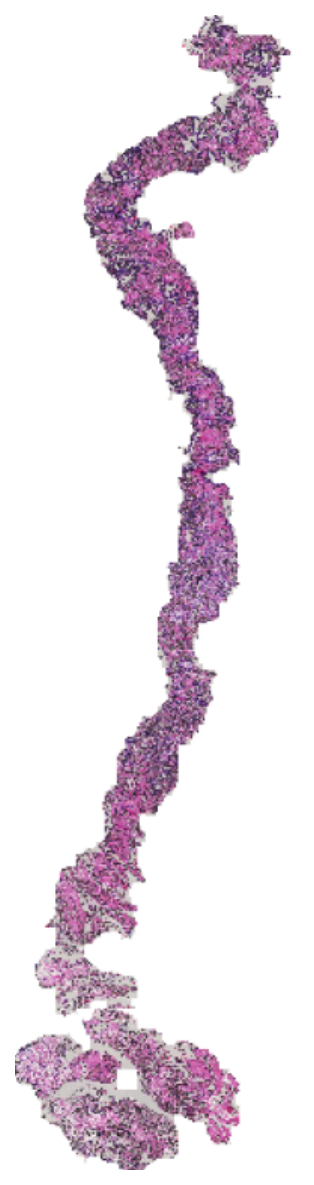}
				& 
				\includegraphics[trim={0.0cm 0cm 0cm 0cm},clip,height=3cm]{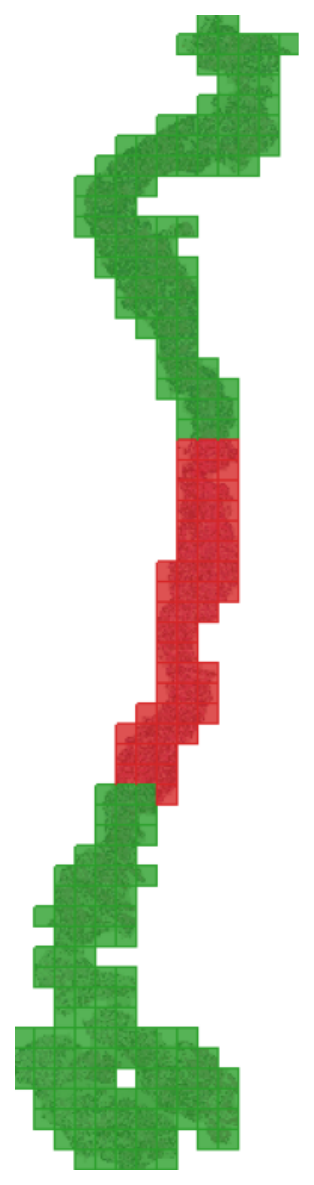}
				& 
				\includegraphics[trim={0.0cm 0cm 0cm 0cm},clip,height=3cm]
				{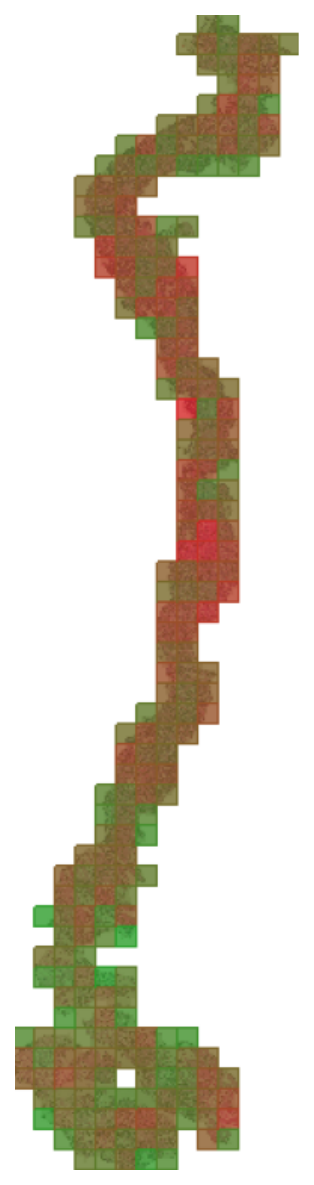}
				& 
				\includegraphics[trim={0.0cm 0cm 0cm 0cm},clip,height=3cm]
				{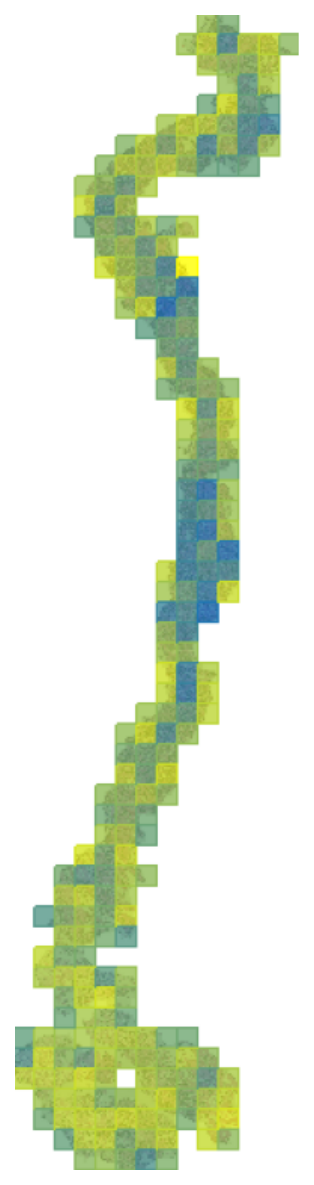}
                &
                \includegraphics[trim={0.0cm 0cm 0cm 0cm},clip,height=3cm]
				{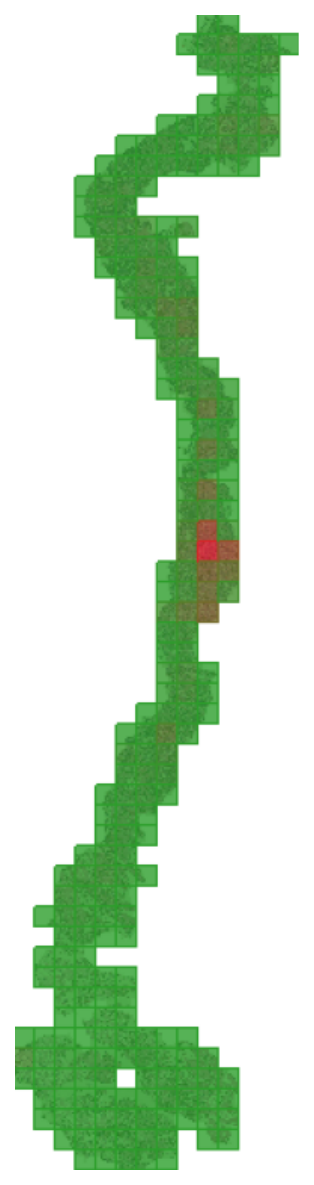}
				& 
				\includegraphics[trim={0.0cm 0cm 0cm 0cm},clip,height=3cm]
				{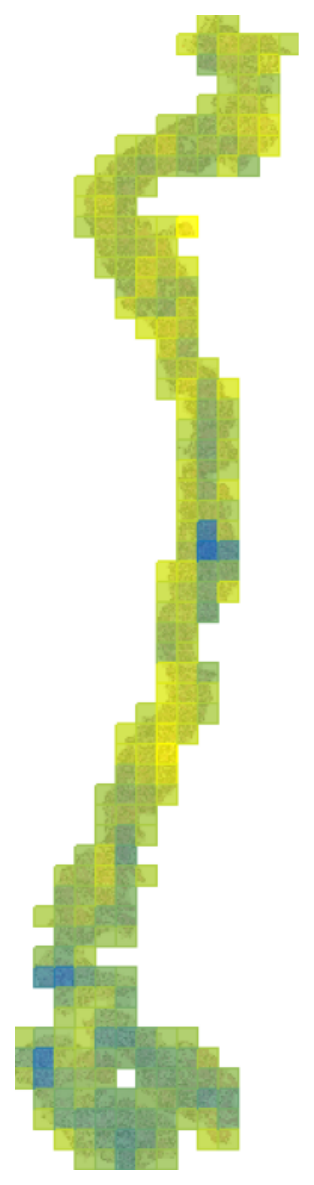}
                \\
                WSI &  Patch labels & \makecell{\abmil+PSA \\ Mean} & \makecell{\abmil+PSA \\ Variance} & \makecell{\transformerabmil+PSA \\ Mean} & \makecell{\transformerabmil+PSA \\ Variance} \\
                \includegraphics[trim={0.0cm 0cm 0cm 0cm},clip,height=3cm]
				{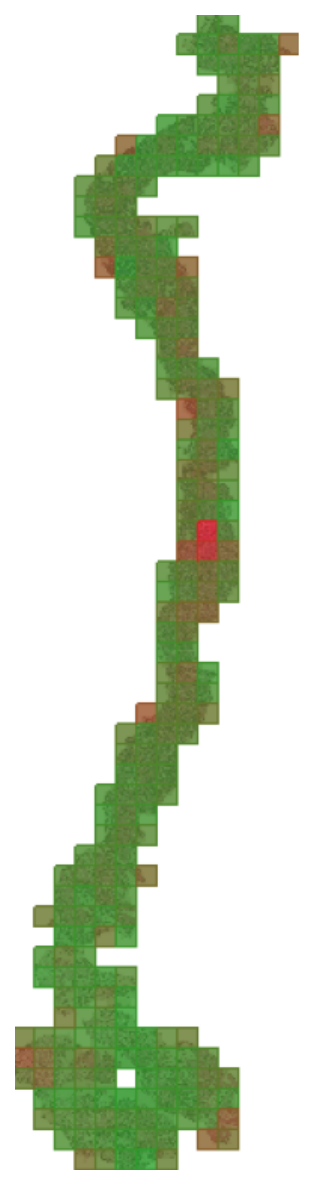}
				& 
				\includegraphics[trim={0.0cm 0cm 0cm 0cm},clip,height=3cm]
				{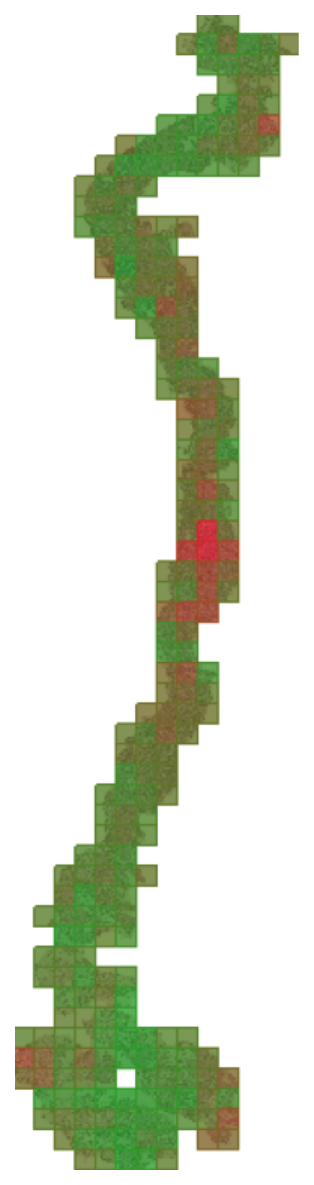}
                &
				\includegraphics[trim={0.0cm 0cm 0cm 0cm},clip,height=3cm]{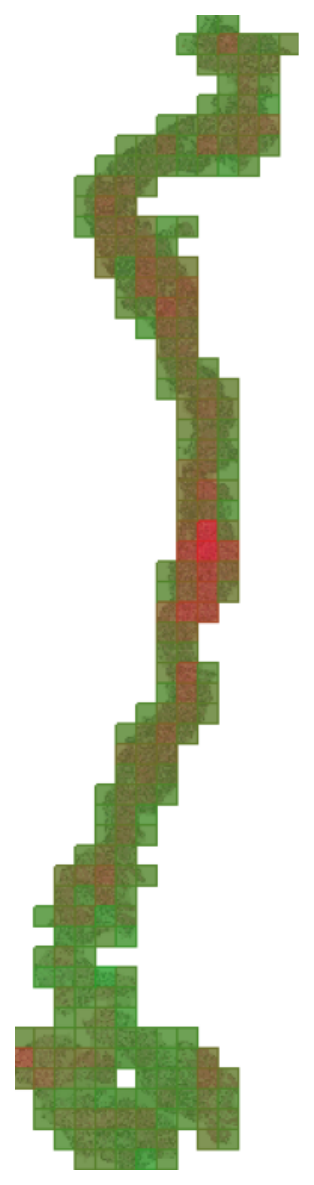}
                &
                \includegraphics[trim={0.0cm 0cm 0cm 0cm},clip,height=3cm]
				{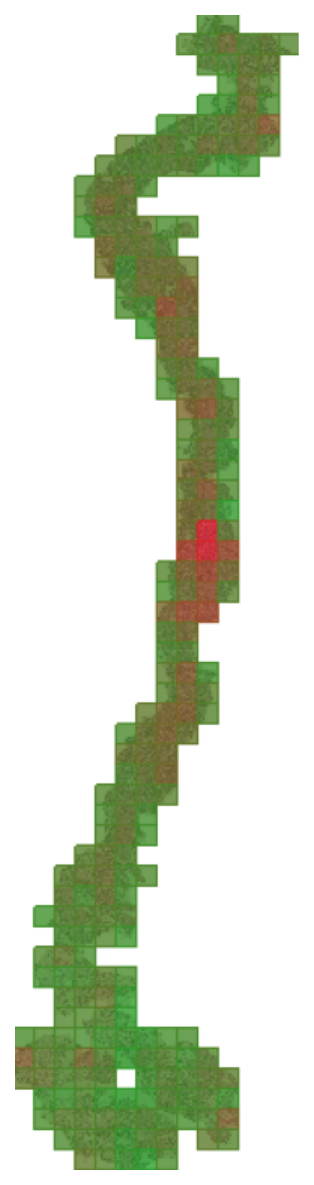}
                &
				\includegraphics[trim={0.0cm 0cm 0cm 0cm},clip,height=3cm]
				{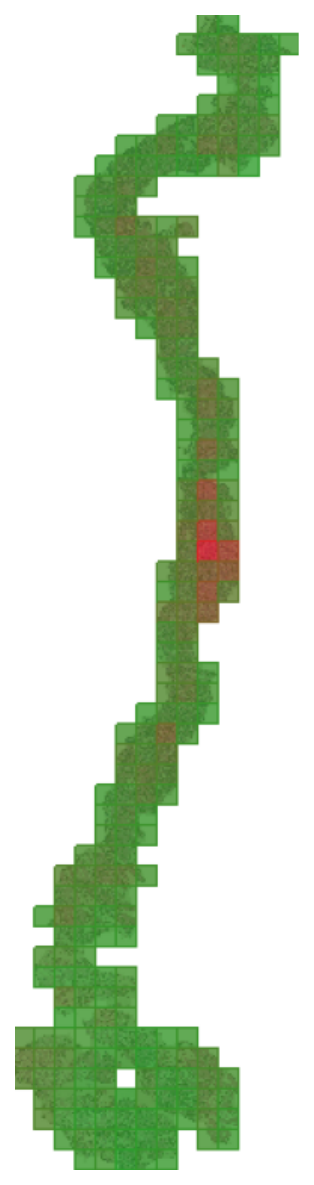}
				& 
				\includegraphics[trim={0.0cm 0cm 0cm 0cm},clip,height=3cm]
				{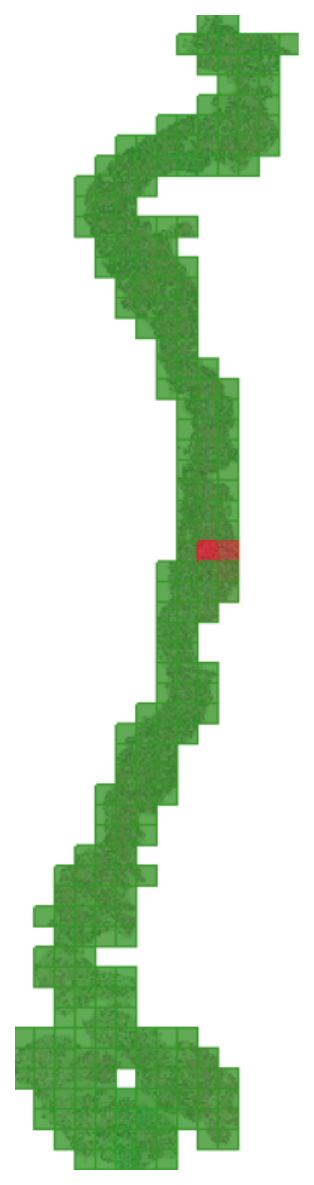}
				\\        
				\abmil & \pathgcn & \dtfdmil & \transformerabmil & \camil & \gtp \\
			\end{tabular}
        \end{adjustbox}
        \caption{Attention maps in a WSI from PANDA. The attention values have been normalized to ease visualization. PSA stands for \probsmoothatt.}
        \label{fig:attmaps-panda-4d048dd585259614c27e11431ec2c860}
        \vspace{-1cm}
    \end{center}
\end{figure}

\begin{figure}[h]
    \scriptsize
	\begin{center}
		\centering
		\begin{adjustbox}{width=0.6\textwidth}
			\begin{tabular}{cc}
				\includegraphics[trim={0cm 0cm 0cm 0cm},clip,width=5cm]{img/att_map-bar_horizontal.pdf} 
				&
				\includegraphics[trim={0cm 0cm 0cm 0cm},clip,width=5cm]{img/uncert_map-bar_horizontal.pdf}
			\end{tabular}
		\end{adjustbox}
		\begin{adjustbox}{width=1.0\textwidth}
			\begin{tabular}{ccccccc}
				\includegraphics[trim={0.0cm 0cm 0cm 0cm},clip,height=3cm]{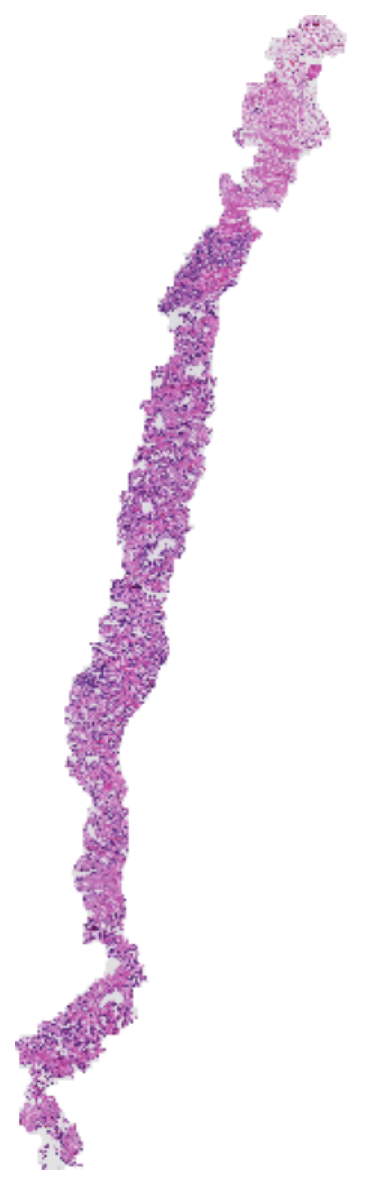}
				& 
				\includegraphics[trim={0.0cm 0cm 0cm 0cm},clip,height=3cm]{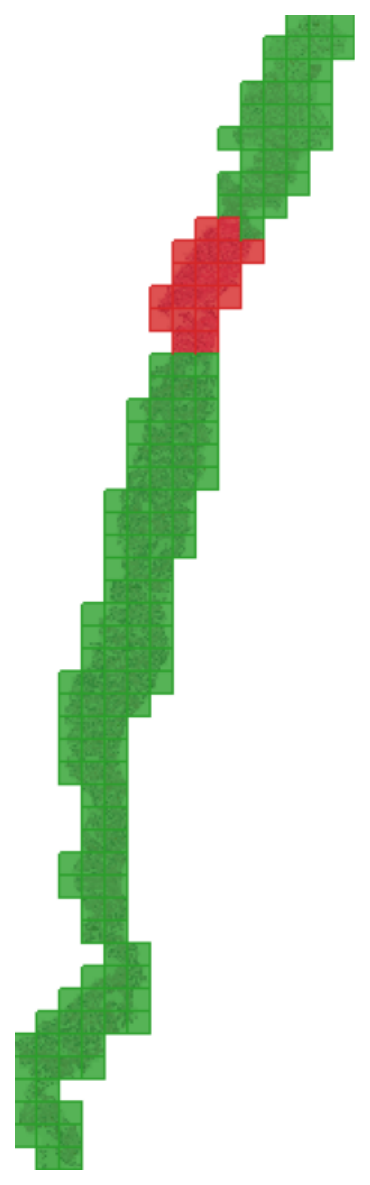}
				& 
				\includegraphics[trim={0.0cm 0cm 0cm 0cm},clip,height=3cm]
				{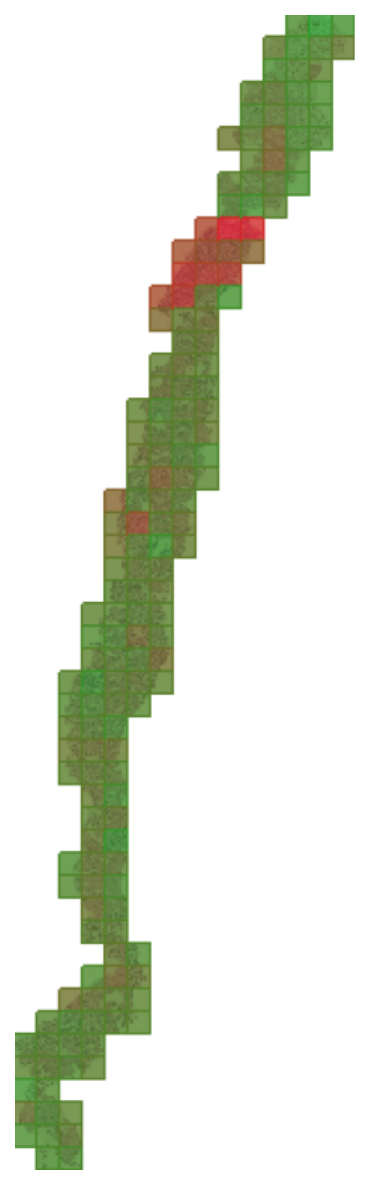}
				& 
				\includegraphics[trim={0.0cm 0cm 0cm 0cm},clip,height=3cm]
				{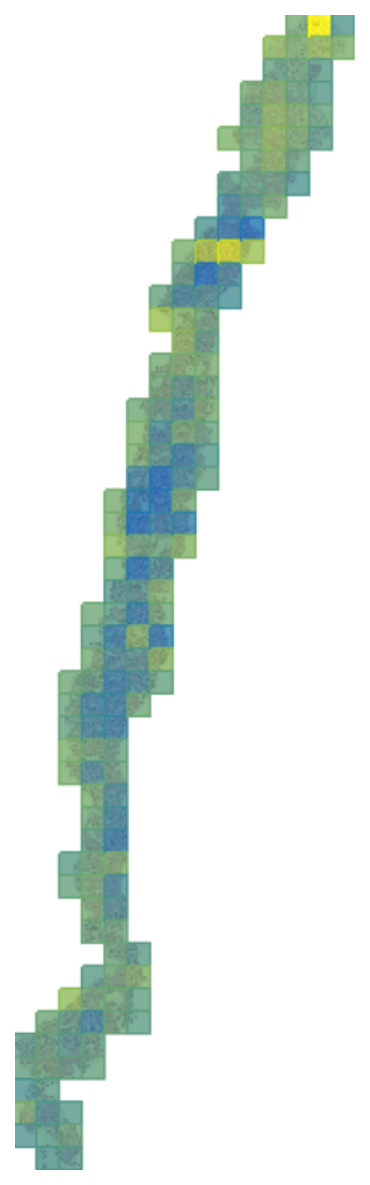}
                &
                \includegraphics[trim={0.0cm 0cm 0cm 0cm},clip,height=3cm]
				{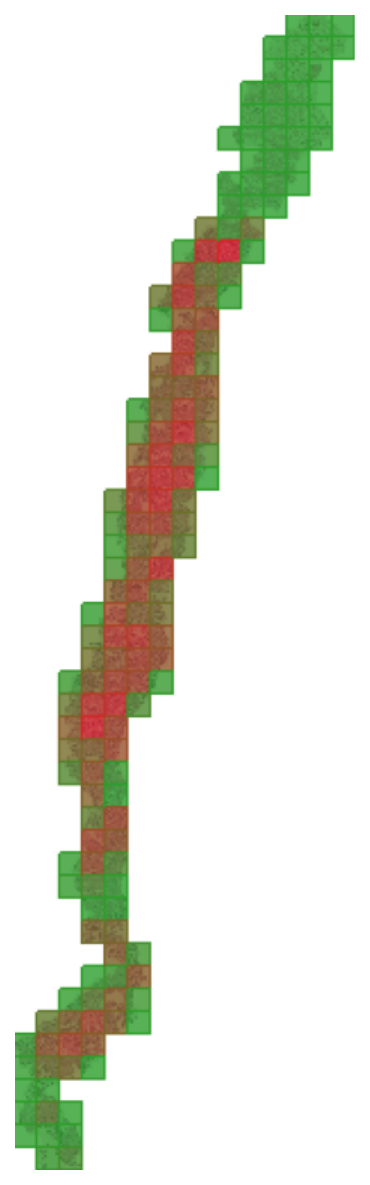}
				& 
				\includegraphics[trim={0.0cm 0cm 0cm 0cm},clip,height=3cm]
				{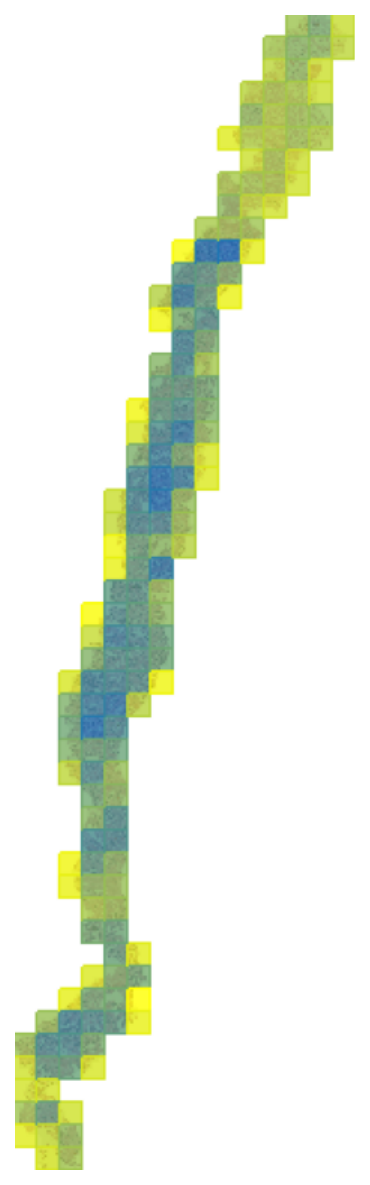}
                \\
                WSI &  Patch labels & \makecell{\abmil+PSA \\ Mean} & \makecell{\abmil+PSA \\ Variance} & \makecell{\transformerabmil+PSA \\ Mean} & \makecell{\transformerabmil+PSA \\ Variance} \\
                \includegraphics[trim={0.0cm 0cm 0cm 0cm},clip,height=3cm]
				{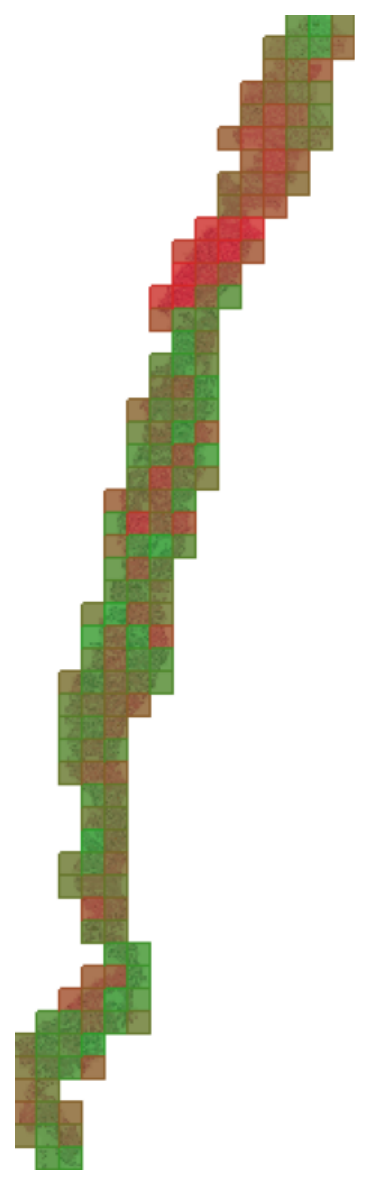}
				& 
				\includegraphics[trim={0.0cm 0cm 0cm 0cm},clip,height=3cm]
				{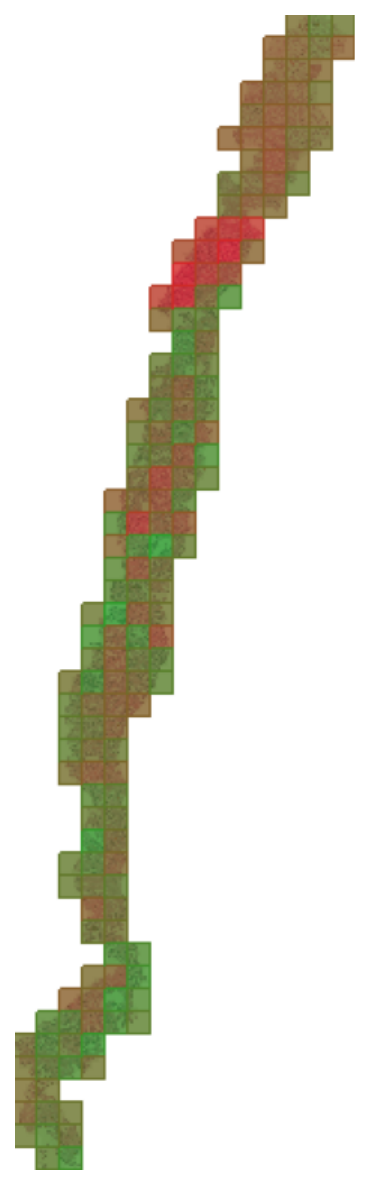}
                &
				\includegraphics[trim={0.0cm 0cm 0cm 0cm},clip,height=3cm]{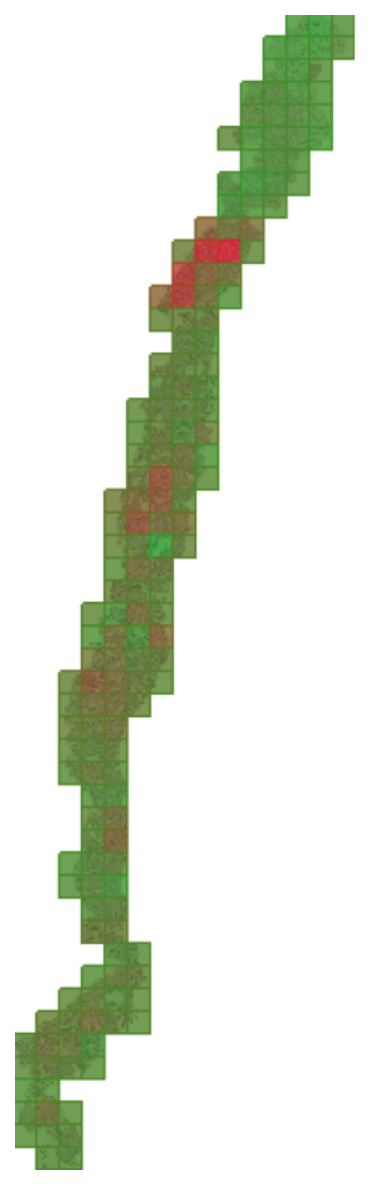}
                &
                \includegraphics[trim={0.0cm 0cm 0cm 0cm},clip,height=3cm]
				{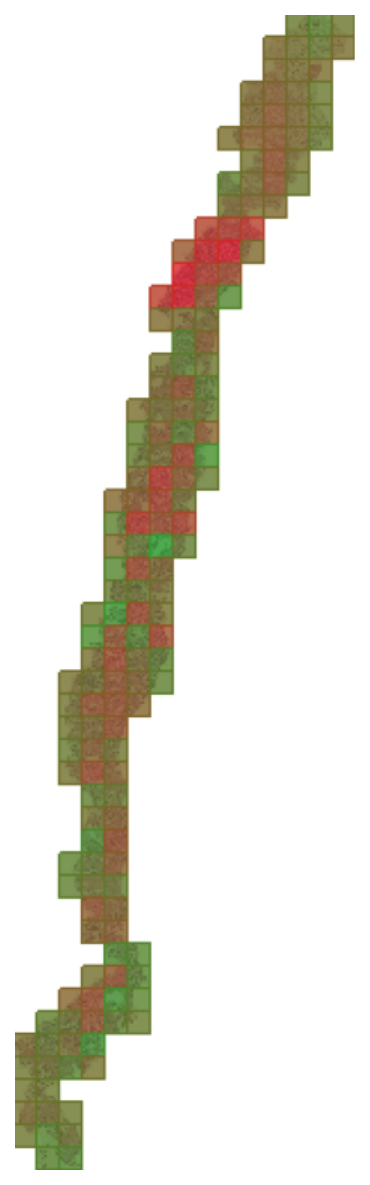}
                &
				\includegraphics[trim={0.0cm 0cm 0cm 0cm},clip,height=3cm]
				{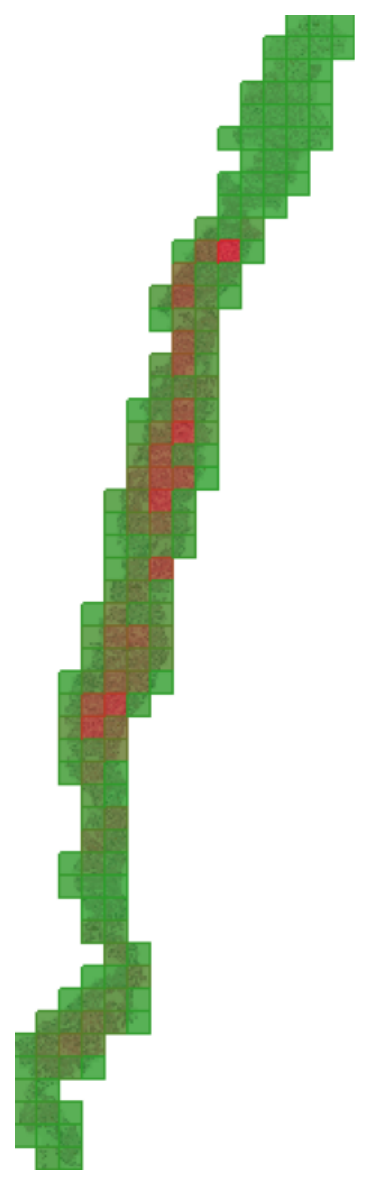}
				& 
				\includegraphics[trim={0.0cm 0cm 0cm 0cm},clip,height=3cm]
				{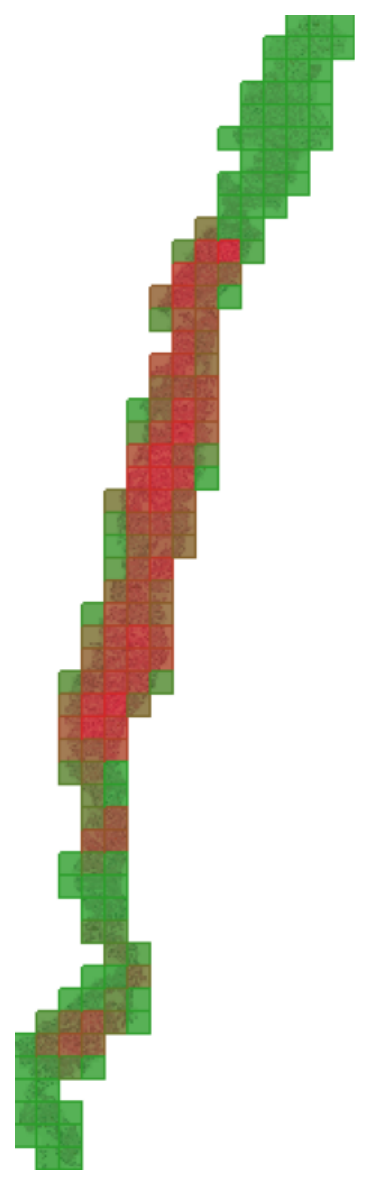}
				\\        
				\abmil & \pathgcn & \dtfdmil & \transformerabmil & \camil & \gtp \\
			\end{tabular}
		\end{adjustbox}
		\caption{Attention maps in a WSI from PANDA. The attention values have been normalized to ease visualization. PSA stands for \probsmoothatt.}
		\label{fig:attmaps-panda-a04310d441e8d2c7a5066627baeec9b6}
        \vspace{-1cm}
	\end{center}
\end{figure}

\begin{figure}[h]
    \renewcommand{\arraystretch}{3} 
	\begin{center}
		\begin{adjustbox}{width=\textwidth}
			\centering
			\begin{tblr}{
					colspec = {X[c,h,2.5cm]X[c]},
					cells   = {font = \fontsize{8pt}{8pt}\selectfont},
                    rows={rowsep=0.2pt}
				}
				& \makecell{\includegraphics[trim={0cm 0cm 0cm 0cm},clip,width=4.5cm]{img/att_map-bar_horizontal.pdf} \hfill \includegraphics[trim={0cm 0cm 0cm 0cm},clip,width=4.5cm]{img/uncert_map-bar_horizontal.pdf}} \\
				CT scan & \includegraphics[trim={0cm 0cm 0cm 0cm},clip,width=0.75\textwidth]
				{img/rsna/scans/scan-ID_fee9e25ce0.pdf} \\
				Slice labels & \includegraphics[trim={0cm 0cm 0cm 0cm},clip,width=0.75\textwidth]
				{img/rsna/gt/gt-ID_fee9e25ce0.pdf} \\
				\makecell{\abmil+PSA \\ Mean} & \includegraphics[trim={0cm 0cm 0cm 0cm},clip,width=0.75\textwidth]{img/rsna/att_maps/att_map-bayes_smooth_abmil_diag-ID_fee9e25ce0.pdf} \\
				\makecell{\abmil+PSA \\ Variance} & \includegraphics[trim={0cm 0cm 0cm 0cm},clip,width=0.75\textwidth]{img/rsna/uncert_maps/uncert_map-bayes_smooth_abmil_diag-ID_fee9e25ce0.pdf} \\
                \makecell{\transformerabmil+PSA \\ Mean} & \includegraphics[trim={0cm 0cm 0cm 0cm},clip,width=0.75\textwidth]{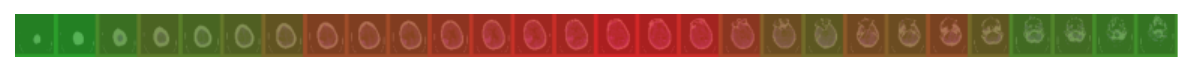} \\
				\makecell{\transformerabmil+PSA \\ Variance} & \includegraphics[trim={0cm 0cm 0cm 0cm},clip,width=0.75\textwidth]{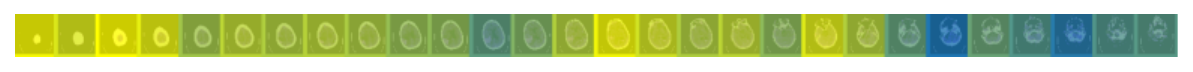} \\
				\abmil & \includegraphics[trim={0cm 0cm 0cm 0cm},clip,width=0.75\textwidth]{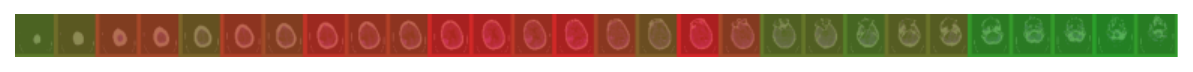} \\
				\pathgcn & \includegraphics[trim={0cm 0cm 0cm 0cm},clip,width=0.75\textwidth]{img/rsna/att_maps/att_map-pathgcn-ID_fee9e25ce0.pdf} \\
				\dtfdmil & \includegraphics[trim={0cm 0cm 0cm 0cm},clip,width=0.75\textwidth]{img/rsna/att_maps/att_map-dftdmil-ID_fee9e25ce0.pdf} \\
				\transformerabmil & \includegraphics[trim={0cm 0cm 0cm 0cm},clip,width=0.75\textwidth]{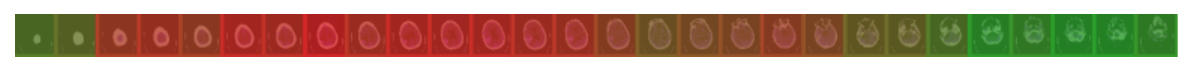} \\
                \gtp & \includegraphics[trim={0cm 0cm 0cm 0cm},clip,width=0.75\textwidth]{img/rsna/att_maps/att_map-gtp-ID_fee9e25ce0.pdf} \\
				\camil & \includegraphics[trim={0cm 0cm 0cm 0cm},clip,width=0.75\textwidth]{img/rsna/att_maps/att_map-camil-ID_fee9e25ce0.pdf} \\
			\end{tblr}
		\end{adjustbox}
        \vspace{-0.4cm}
		\caption{
			Attention maps in a CT scan from RSNA. The attention values have been normalized to ease visualization. PSA stands for \probsmoothatt. 
		}
		\label{fig:attmaps-rsna-ID_fee9e25ce0}
        \vspace{-3cm}
	\end{center}
\end{figure}

\begin{figure}[h]
    \begin{center}
        \begin{adjustbox}{width=\textwidth}
			\centering
			\begin{tblr}{
					colspec = {X[c,h,2.5cm]X[c]},
					cells   = {font = \fontsize{8pt}{8pt}\selectfont},
                    rows={rowsep=0.2pt}
				}
				& \makecell{\includegraphics[trim={0cm 0cm 0cm 0cm},clip,width=4.5cm]{img/att_map-bar_horizontal.pdf} \hfill \includegraphics[trim={0cm 0cm 0cm 0cm},clip,width=4.5cm]{img/uncert_map-bar_horizontal.pdf}} \\
				CT scan & \includegraphics[trim={0cm 0cm 0cm 0cm},clip,width=0.75\textwidth]
				{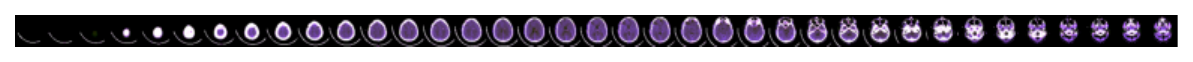} \\
				Slice labels & \includegraphics[trim={0cm 0cm 0cm 0cm},clip,width=0.75\textwidth]
				{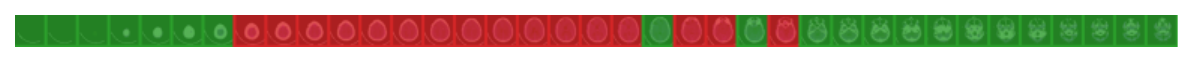} \\
				\makecell{\abmil+PSA \\ Mean} & \includegraphics[trim={0cm 0cm 0cm 0cm},clip,width=0.75\textwidth]{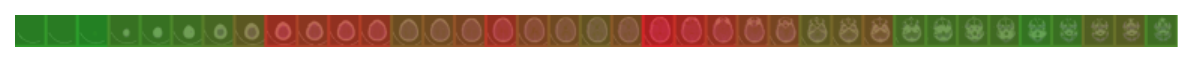} \\
				\makecell{\abmil+PSA \\ Variance} & \includegraphics[trim={0cm 0cm 0cm 0cm},clip,width=0.75\textwidth]{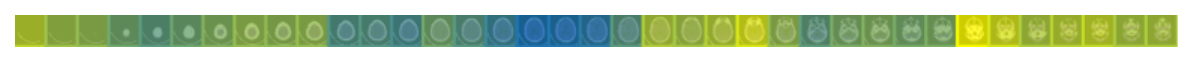} \\
                \makecell{\transformerabmil+PSA \\ Mean} & \includegraphics[trim={0cm 0cm 0cm 0cm},clip,width=0.75\textwidth]{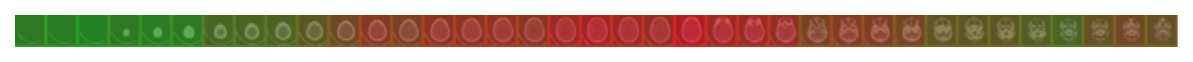} \\
				\makecell{\transformerabmil+PSA \\ Variance} & \includegraphics[trim={0cm 0cm 0cm 0cm},clip,width=0.75\textwidth]{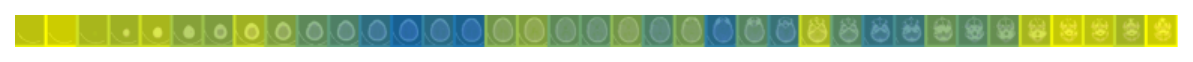} \\
				\abmil & \includegraphics[trim={0cm 0cm 0cm 0cm},clip,width=0.75\textwidth]{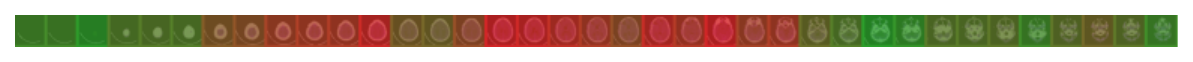} \\
				\pathgcn & \includegraphics[trim={0cm 0cm 0cm 0cm},clip,width=0.75\textwidth]{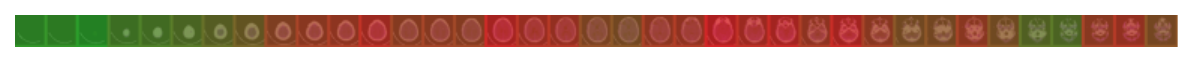} \\
				\dtfdmil & \includegraphics[trim={0cm 0cm 0cm 0cm},clip,width=0.75\textwidth]{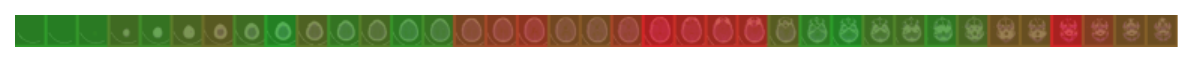} \\
				\transformerabmil & \includegraphics[trim={0cm 0cm 0cm 0cm},clip,width=0.75\textwidth]{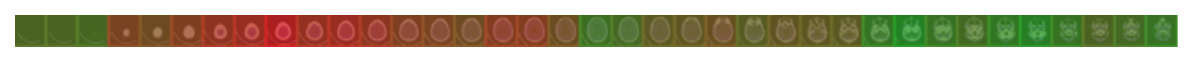} \\
                \gtp & \includegraphics[trim={0cm 0cm 0cm 0cm},clip,width=0.75\textwidth]{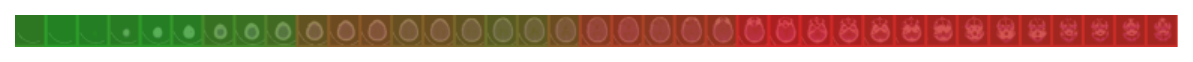} \\
				\camil & \includegraphics[trim={0cm 0cm 0cm 0cm},clip,width=0.75\textwidth]{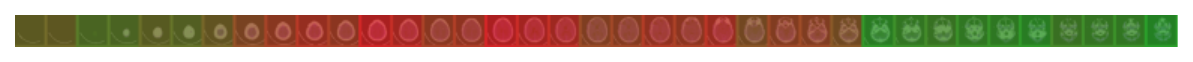} \\
			\end{tblr}
		\end{adjustbox}
        \vspace{-0.4cm}
        \caption{Attention maps in a CT scan from RSNA. The attention values have been normalized to ease visualization. PSA stands for \probsmoothatt. }
        \label{fig:attmaps-rsna-ID_fdf596e74f}
        \vspace{-2.7cm}
    \end{center}
\end{figure}

\begin{figure}[h]
	\begin{center}
		\begin{adjustbox}{width=\textwidth}
			\centering
			\begin{tblr}{
					colspec = {X[c,h,2.5cm]X[c]},
					cells   = {font = \fontsize{8pt}{8pt}\selectfont},
                    rows={rowsep=0.2pt}
				}
				& \makecell{\includegraphics[trim={0cm 0cm 0cm 0cm},clip,width=4.5cm]{img/att_map-bar_horizontal.pdf} \hfill \includegraphics[trim={0cm 0cm 0cm 0cm},clip,width=4.5cm]{img/uncert_map-bar_horizontal.pdf}} \\
				CT scan & \includegraphics[trim={0cm 0cm 0cm 0cm},clip,width=0.75\textwidth]
				{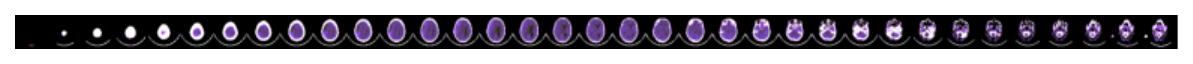} \\
				Slice labels & \includegraphics[trim={0cm 0cm 0cm 0cm},clip,width=0.75\textwidth]
				{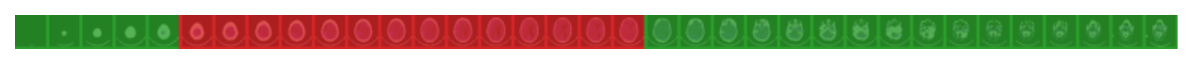} \\
				\makecell{\abmil+PSA \\ Mean} & \includegraphics[trim={0cm 0cm 0cm 0cm},clip,width=0.75\textwidth]{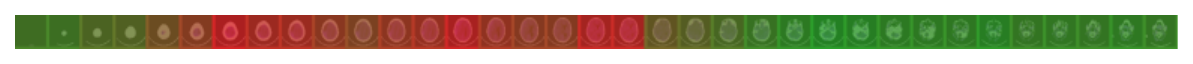} \\
				\makecell{\abmil+PSA \\ Variance} & \includegraphics[trim={0cm 0cm 0cm 0cm},clip,width=0.75\textwidth]{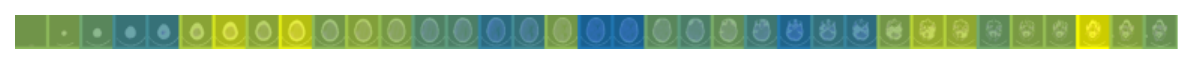} \\
                \makecell{\transformerabmil+PSA \\ Mean} & \includegraphics[trim={0cm 0cm 0cm 0cm},clip,width=0.75\textwidth]{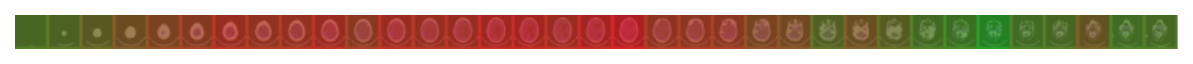} \\
				\makecell{\transformerabmil+PSA \\ Variance} & \includegraphics[trim={0cm 0cm 0cm 0cm},clip,width=0.75\textwidth]{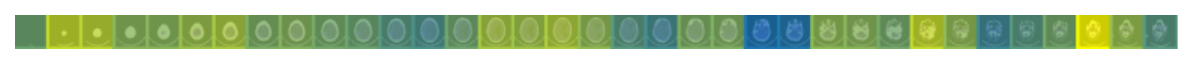} \\
				\abmil & \includegraphics[trim={0cm 0cm 0cm 0cm},clip,width=0.75\textwidth]{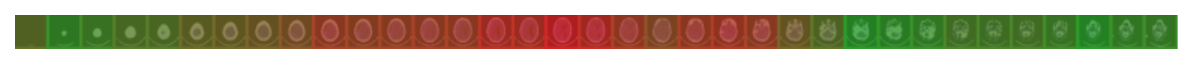} \\
				\pathgcn & \includegraphics[trim={0cm 0cm 0cm 0cm},clip,width=0.75\textwidth]{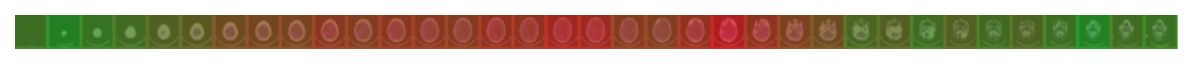} \\
				\dtfdmil & \includegraphics[trim={0cm 0cm 0cm 0cm},clip,width=0.75\textwidth]{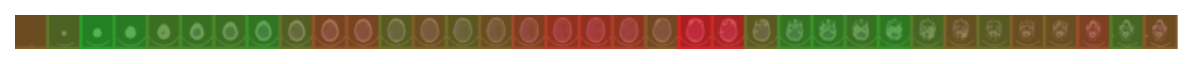} \\
				\transformerabmil & \includegraphics[trim={0cm 0cm 0cm 0cm},clip,width=0.75\textwidth]{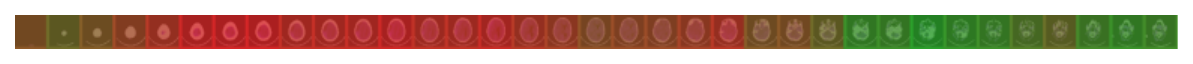} \\
                \gtp & \includegraphics[trim={0cm 0cm 0cm 0cm},clip,width=0.75\textwidth]{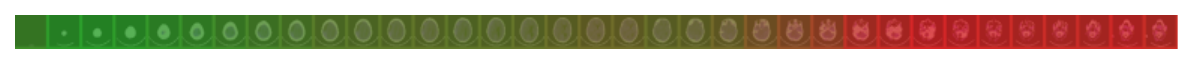} \\
				\camil & \includegraphics[trim={0cm 0cm 0cm 0cm},clip,width=0.75\textwidth]{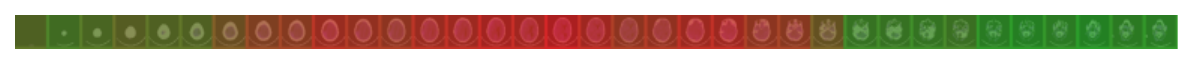} \\
			\end{tblr}
		\end{adjustbox}
        \vspace{-0.4cm}
		\caption{Attention maps in a CT scan from RSNA. The attention values have been normalized to ease visualization. PSA stands for \probsmoothatt.}
		\label{fig:attmaps-rsna-ID_ffd377876f}
        \vspace{-4cm}
	\end{center}
\end{figure}


\begin{figure}[t!]
    \Large
	\begin{center}
		\centering
		\begin{adjustbox}{width=0.7\textwidth}
			\begin{tabular}{cc}
				\includegraphics[trim={0cm 0cm 0cm 0cm},clip,width=5cm]{img/att_map-bar_horizontal.pdf} 
				&
				\includegraphics[trim={0cm 0cm 0cm 0cm},clip,width=5cm]{img/uncert_map-bar_horizontal.pdf}
			\end{tabular}
		\end{adjustbox}
		\begin{adjustbox}{width=1.0\textwidth}
			\begin{tabular}{ccccccc}
				\includegraphics[trim={0.0cm 0cm 0cm 0cm},clip,height=3cm]{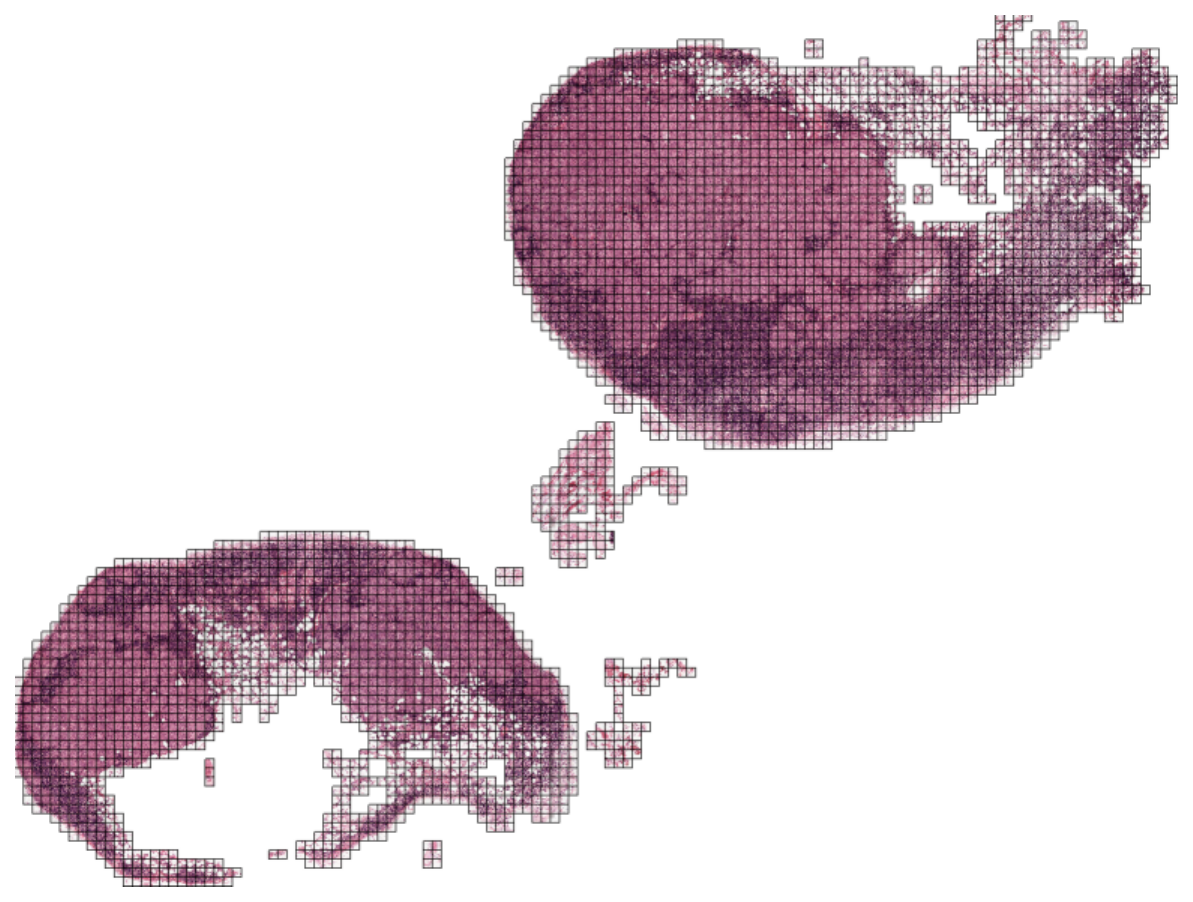}
				& 
				\includegraphics[trim={0.0cm 0cm 0cm 0cm},clip,height=3cm]{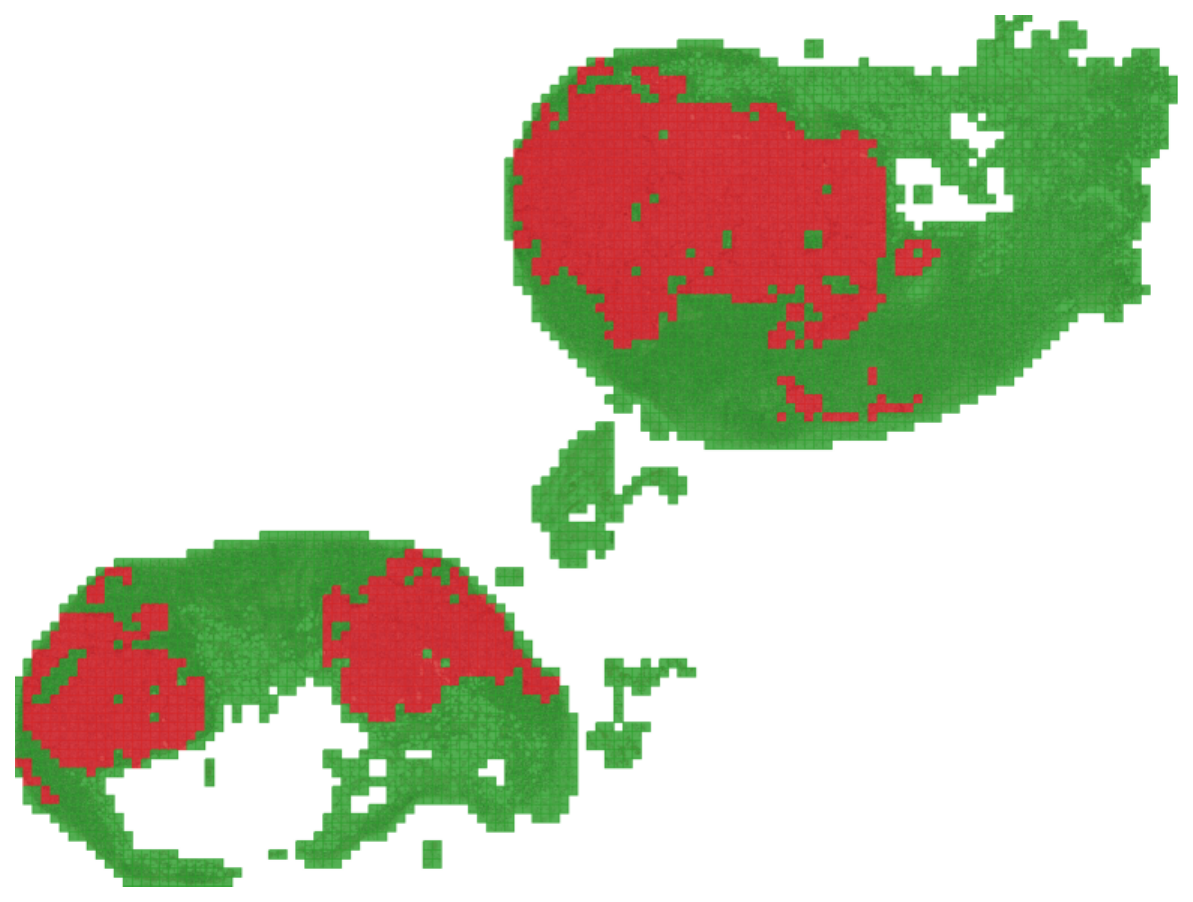}
				& 
				\includegraphics[trim={0.0cm 0cm 0cm 0cm},clip,height=3cm]
				{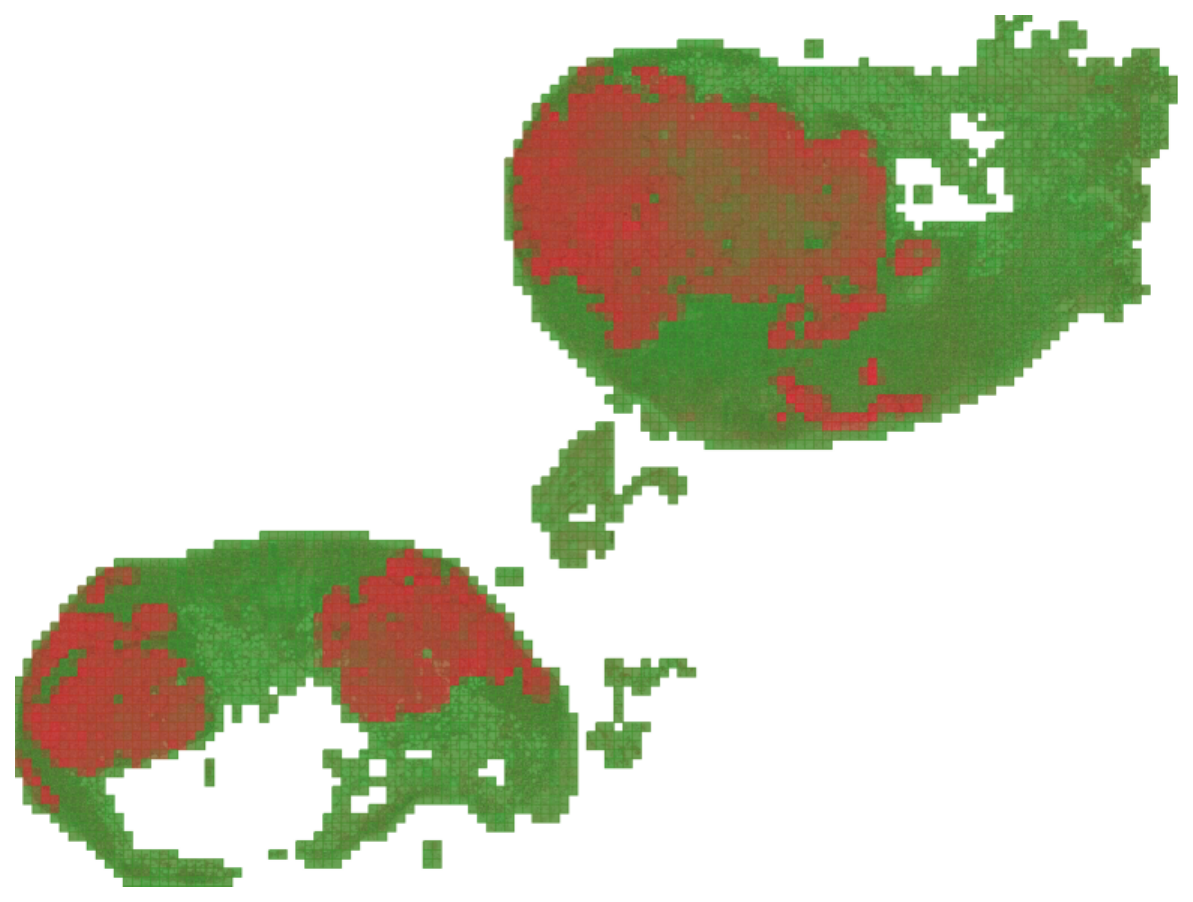}
				& 
				\includegraphics[trim={0.0cm 0cm 0cm 0cm},clip,height=3cm]
				{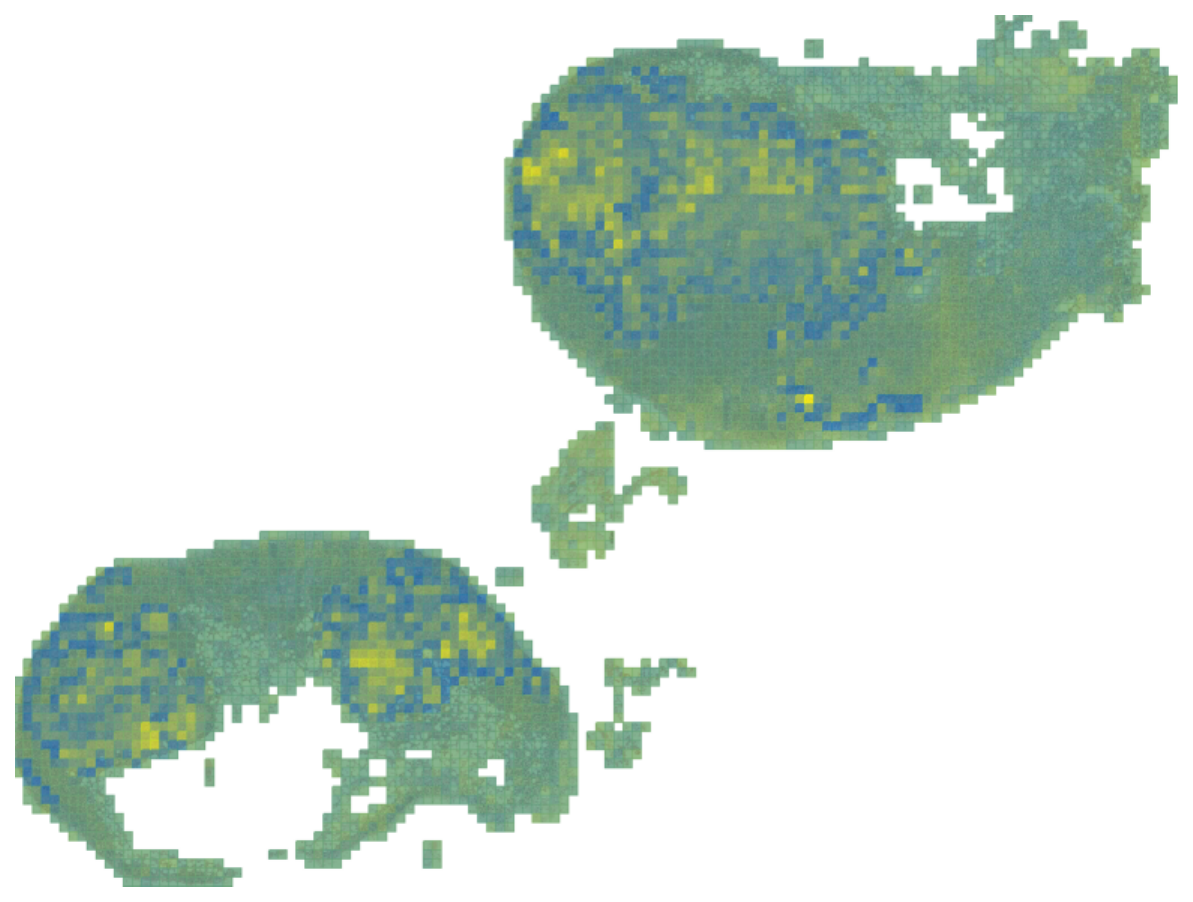}
                &
                \includegraphics[trim={0.0cm 0cm 0cm 0cm},clip,height=3cm]
				{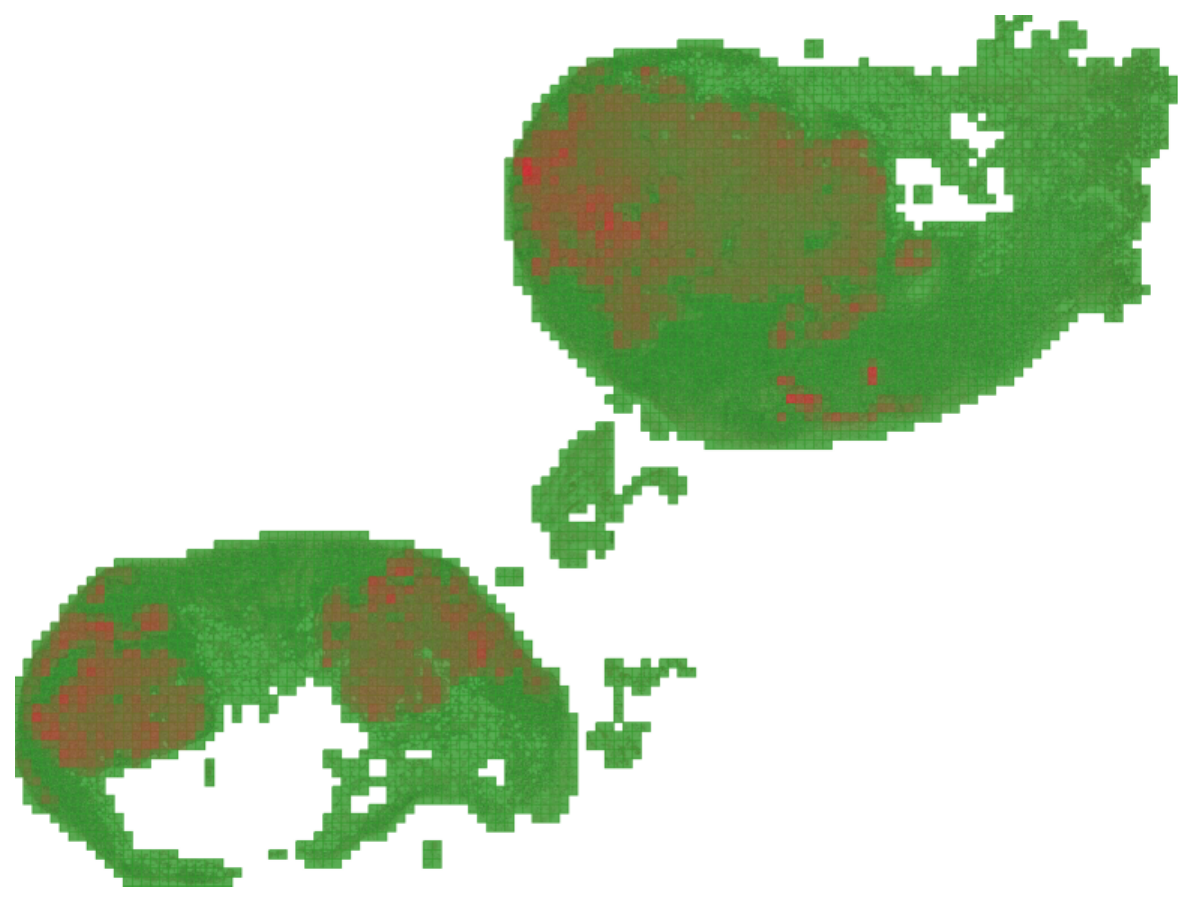}
				& 
				\includegraphics[trim={0.0cm 0cm 0cm 0cm},clip,height=3cm]
				{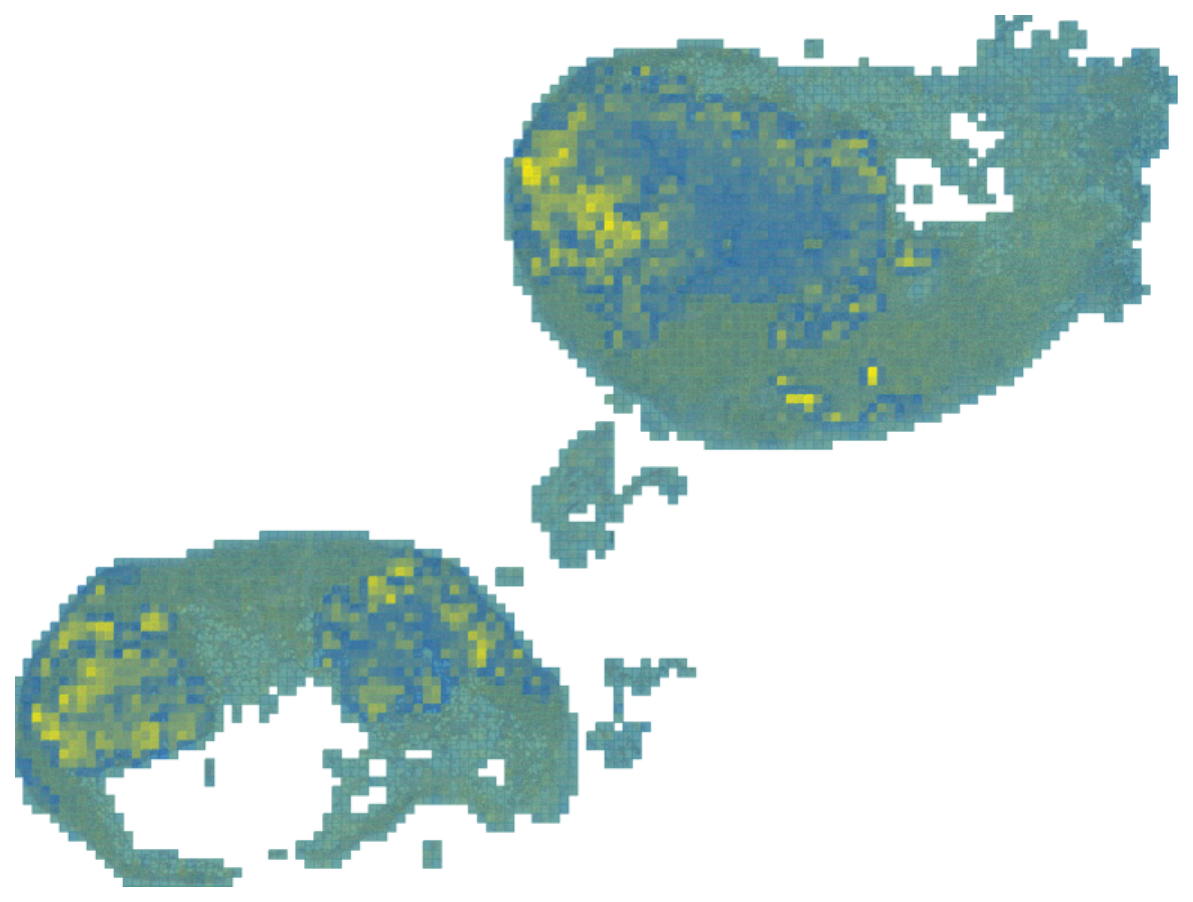}
                \\
                WSI &  Patch labels & \makecell{\abmil+PSA \\ Mean} & \makecell{\abmil+PSA \\ Variance} & \makecell{\transformerabmil+PSA \\ Mean} & \makecell{\transformerabmil+PSA \\ Variance} \\
                \includegraphics[trim={0.0cm 0cm 0cm 0cm},clip,height=3cm]
				{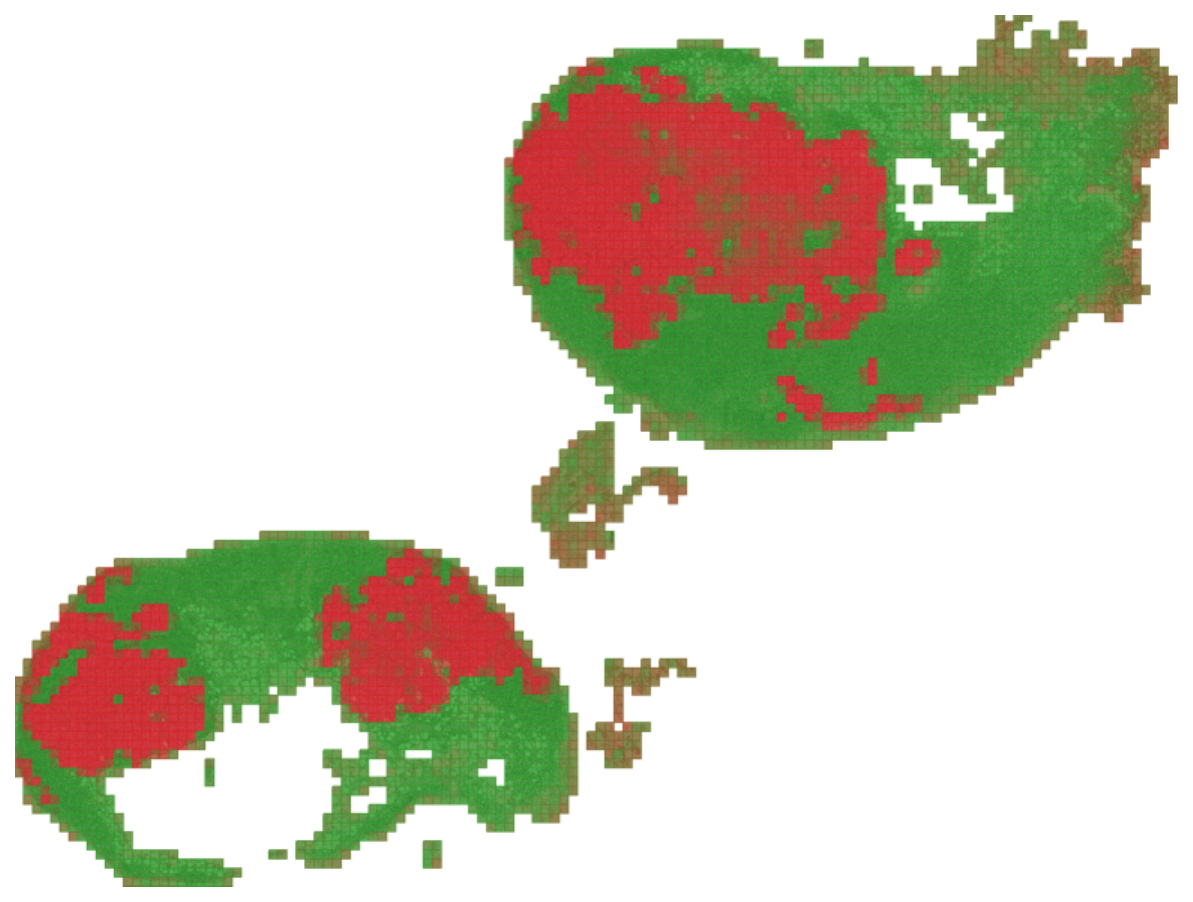}
				& 
				\includegraphics[trim={0.0cm 0cm 0cm 0cm},clip,height=3cm]
				{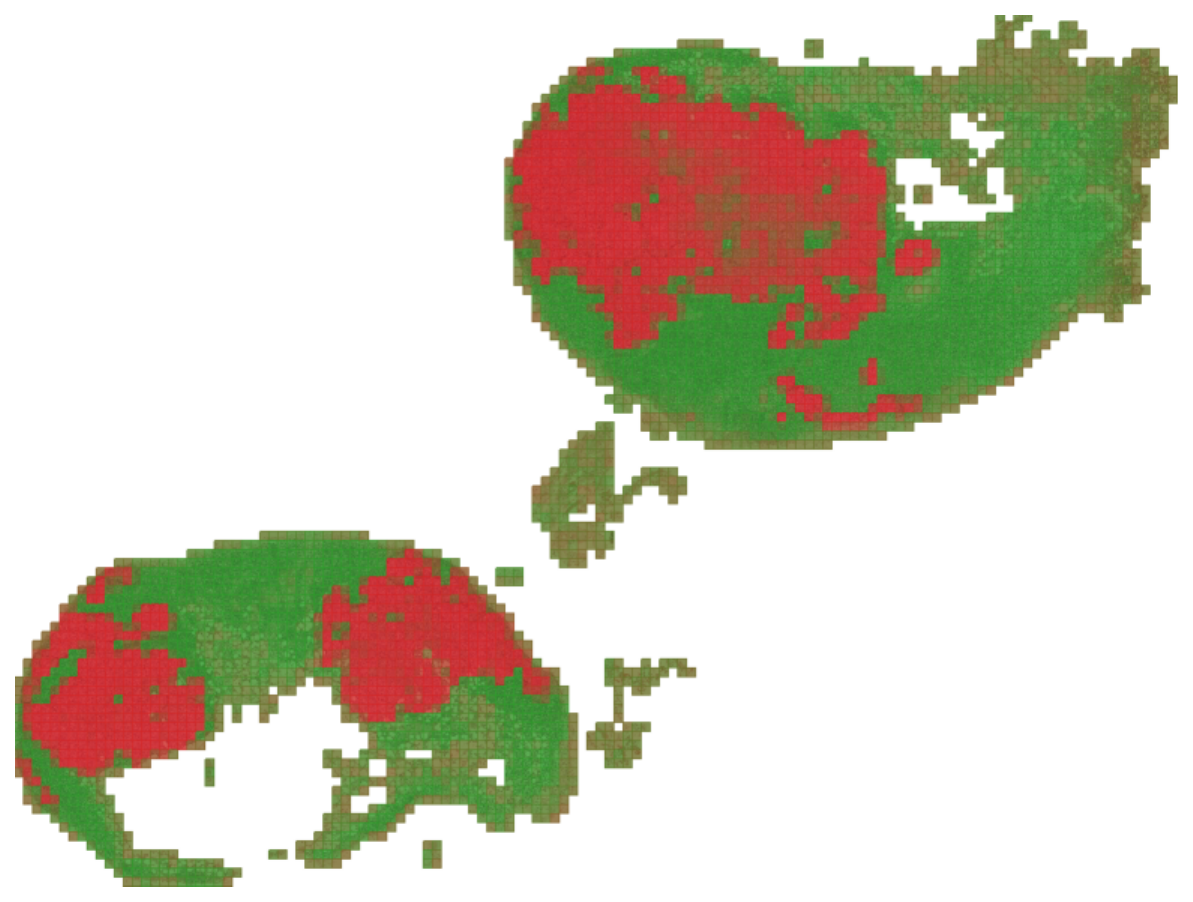}
                &
				\includegraphics[trim={0.0cm 0cm 0cm 0cm},clip,height=3cm]{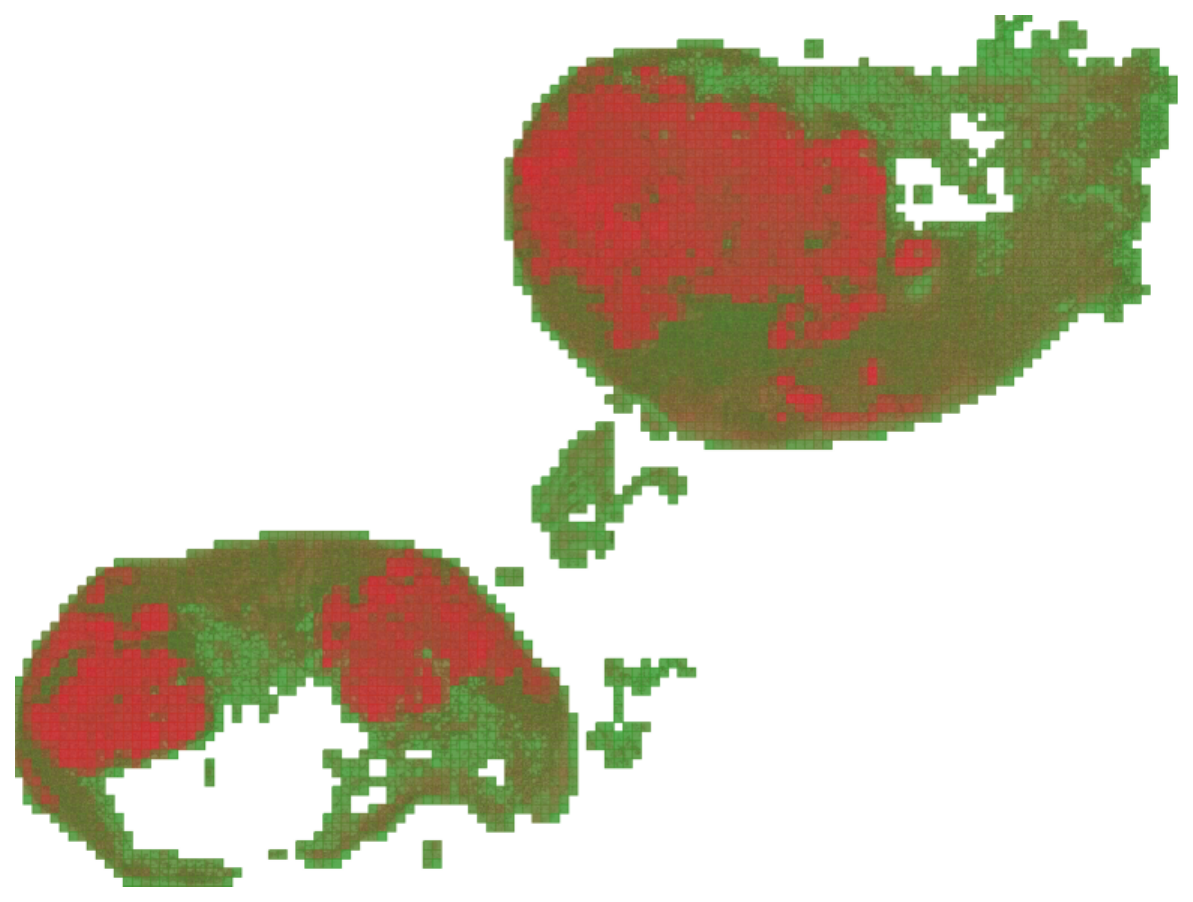}
                &
                \includegraphics[trim={0.0cm 0cm 0cm 0cm},clip,height=3cm]
				{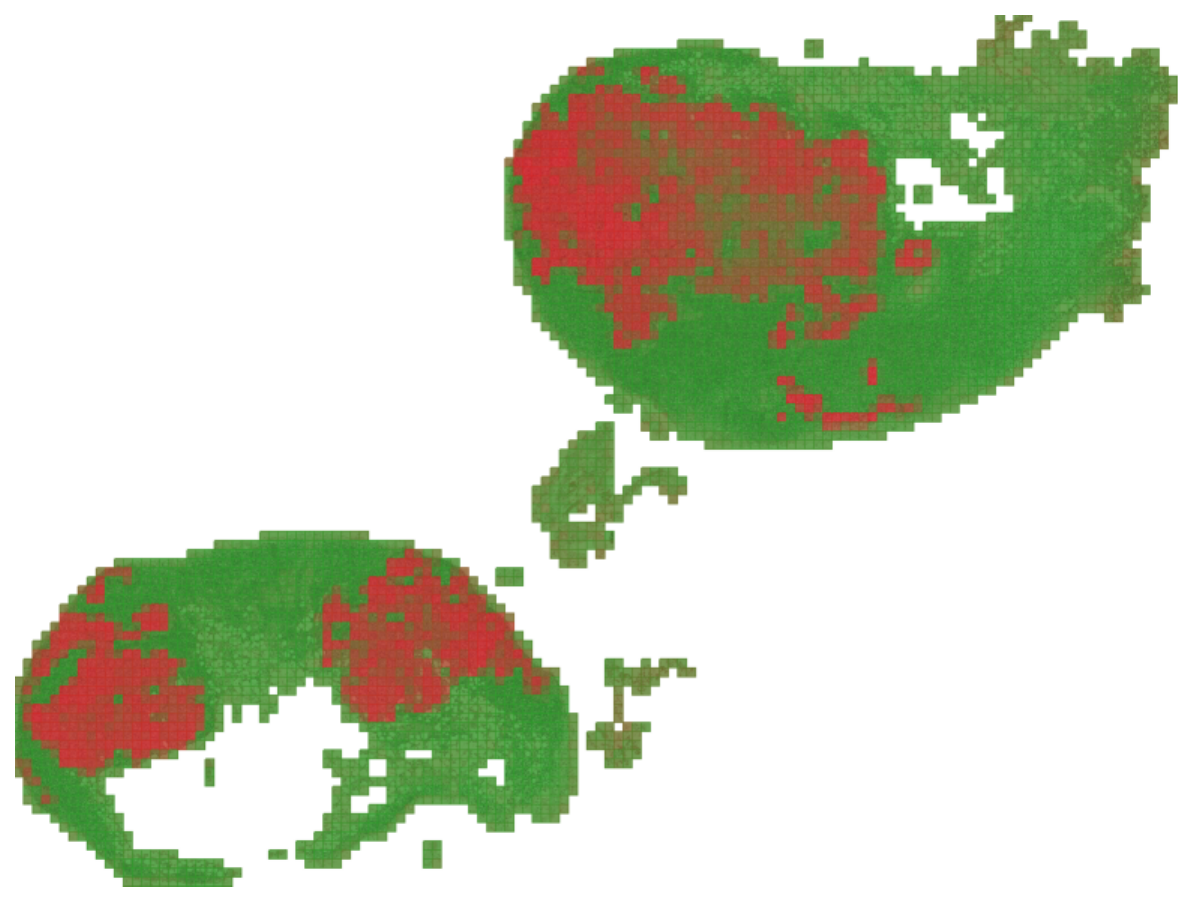}
                &
				\includegraphics[trim={0.0cm 0cm 0cm 0cm},clip,height=3cm]
				{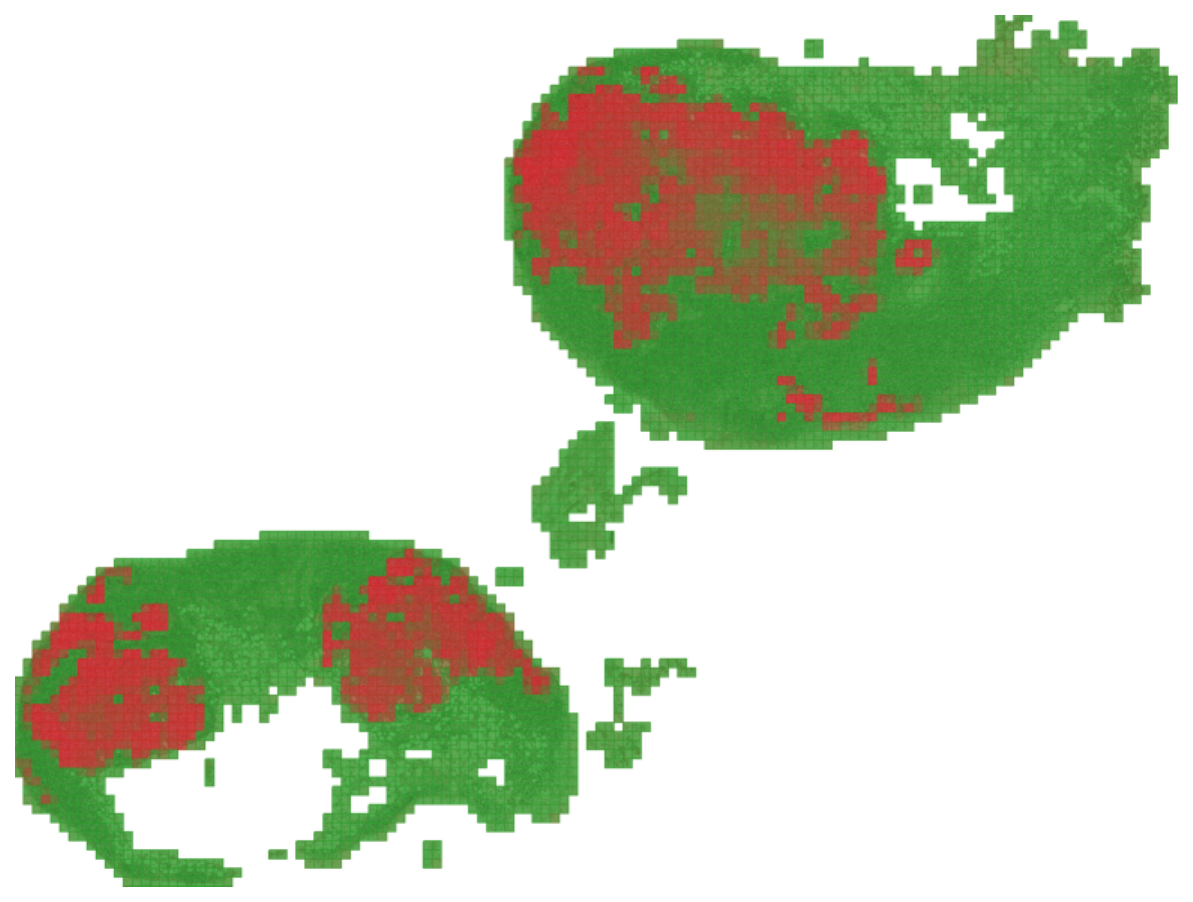}
				& 
				\includegraphics[trim={0.0cm 0cm 0cm 0cm},clip,height=3cm]
				{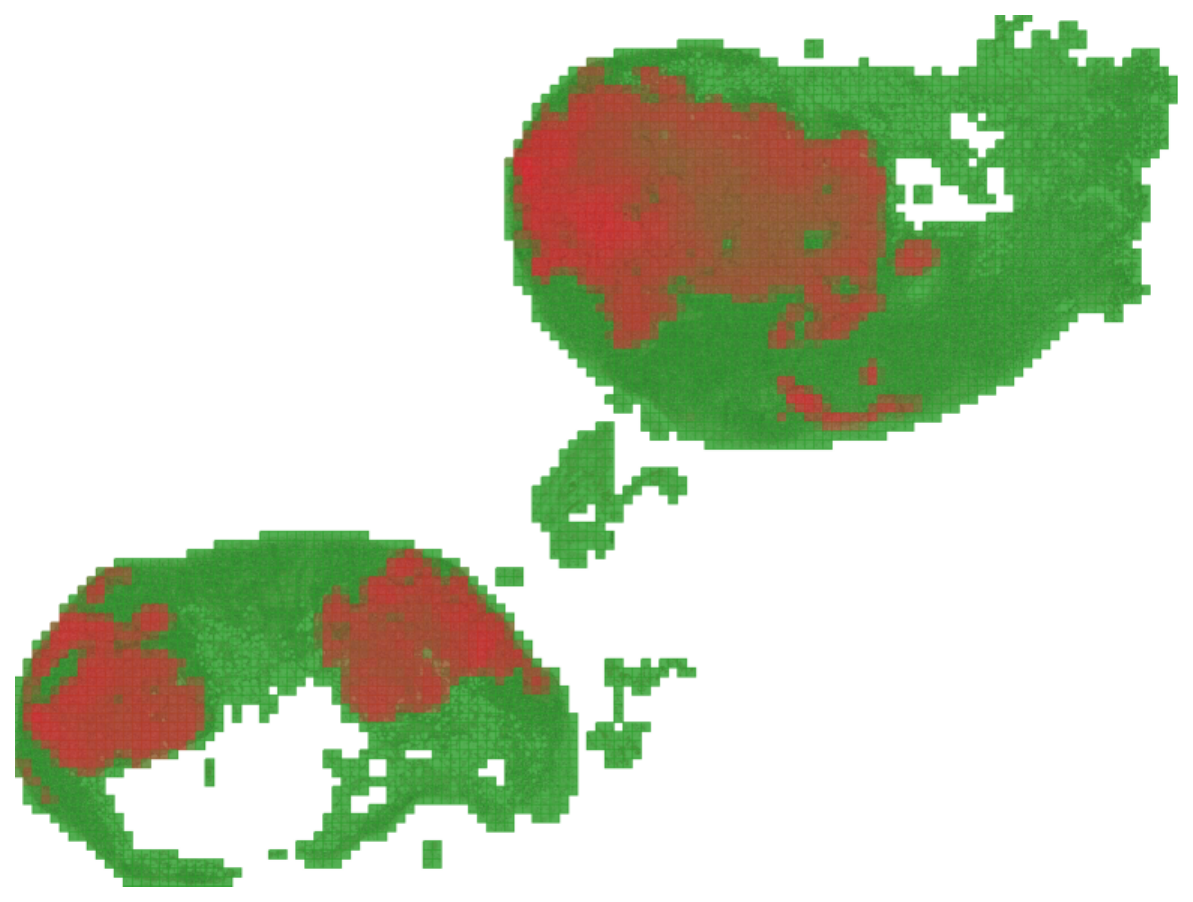}
				\\        
				\abmil & \pathgcn & \dtfdmil & \transformerabmil & \camil & \gtp \\
			\end{tabular}
		\end{adjustbox}
		\caption{Attention maps in a WSI from CAMELYON16. The attention values have been normalized to ease visualization. PSA stands for \probsmoothatt.}
		\label{fig:attmaps-camelyon-test_040}
        \vspace{-2cm}
	\end{center}
\end{figure}

\begin{figure}[t!]
    \Large
    \begin{center}
        \centering
        \begin{adjustbox}{width=0.7\textwidth}
        \begin{tabular}{cc}
             \includegraphics[trim={0cm 0cm 0cm 0cm},clip,width=5cm]{img/att_map-bar_horizontal.pdf} 
             &
             \includegraphics[trim={0cm 0cm 0cm 0cm},clip,width=5cm]{img/uncert_map-bar_horizontal.pdf}
        \end{tabular}
        \end{adjustbox}
        \begin{adjustbox}{width=1.0\textwidth}
			\begin{tabular}{ccccccc}
				\includegraphics[trim={0.0cm 0cm 0cm 0cm},clip,height=3cm]{img/camelyon16/wsis/wsi-test_105.pdf}
				& 
				\includegraphics[trim={0.0cm 0cm 0cm 0cm},clip,height=3cm]{img/camelyon16/gt/gt-test_105.pdf}
				& 
				\includegraphics[trim={0.0cm 0cm 0cm 0cm},clip,height=3cm]
				{img/camelyon16/att_maps/att_map-bayes_smooth_abmil_diag-test_105.pdf}
				& 
				\includegraphics[trim={0.0cm 0cm 0cm 0cm},clip,height=3cm]
				{img/camelyon16/uncert_maps/uncert_map-bayes_smooth_abmil_diag-test_105.pdf}
                &
                \includegraphics[trim={0.0cm 0cm 0cm 0cm},clip,height=3cm]
				{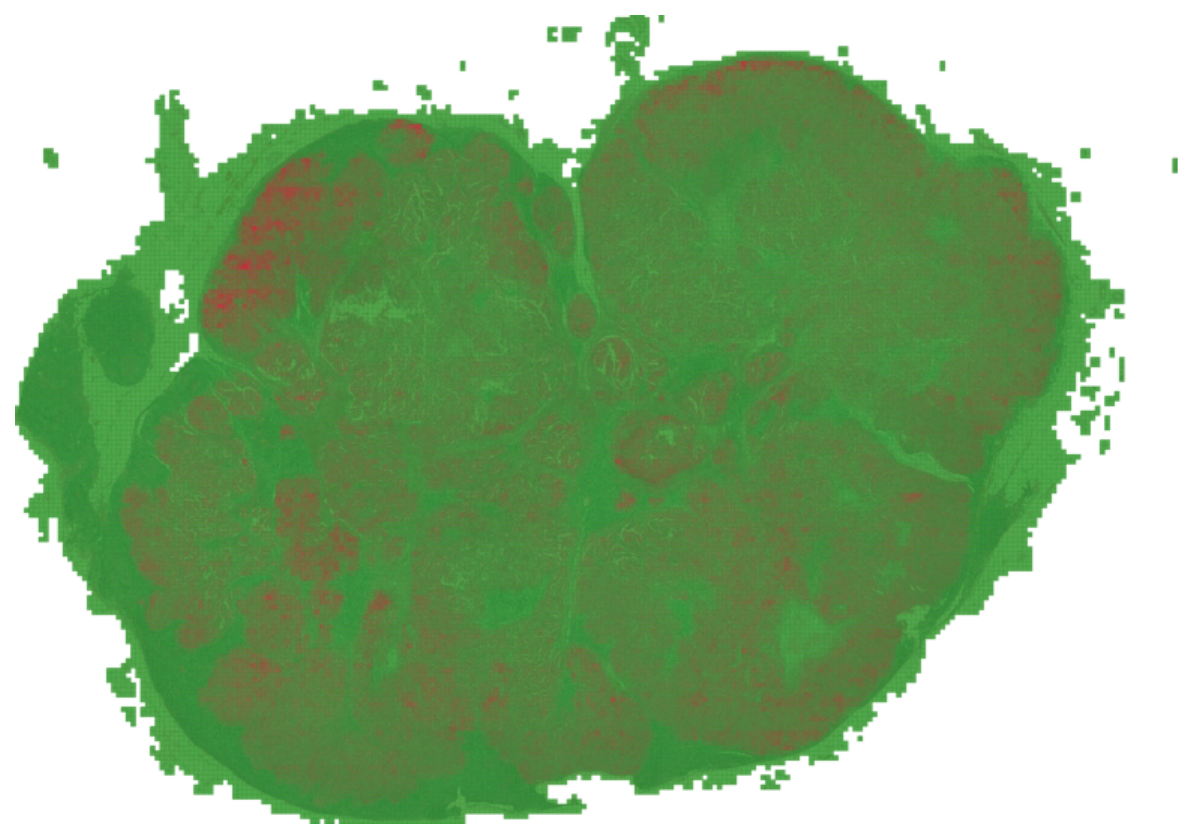}
				& 
				\includegraphics[trim={0.0cm 0cm 0cm 0cm},clip,height=3cm]
				{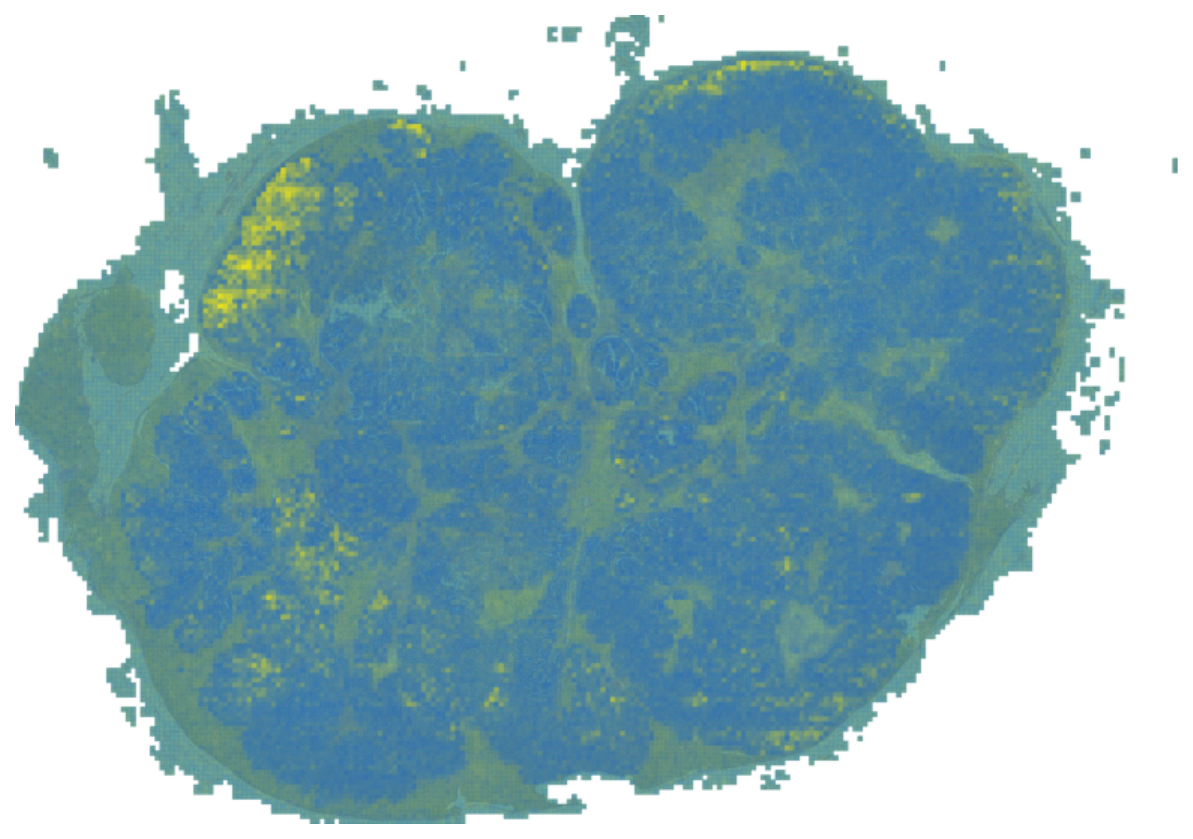}
                \\
                WSI &  Patch labels & \makecell{\abmil+PSA \\ Mean} & \makecell{\abmil+PSA \\ Variance} & \makecell{\transformerabmil+PSA \\ Mean} & \makecell{\transformerabmil+PSA \\ Variance} \\
                \includegraphics[trim={0.0cm 0cm 0cm 0cm},clip,height=3cm]
				{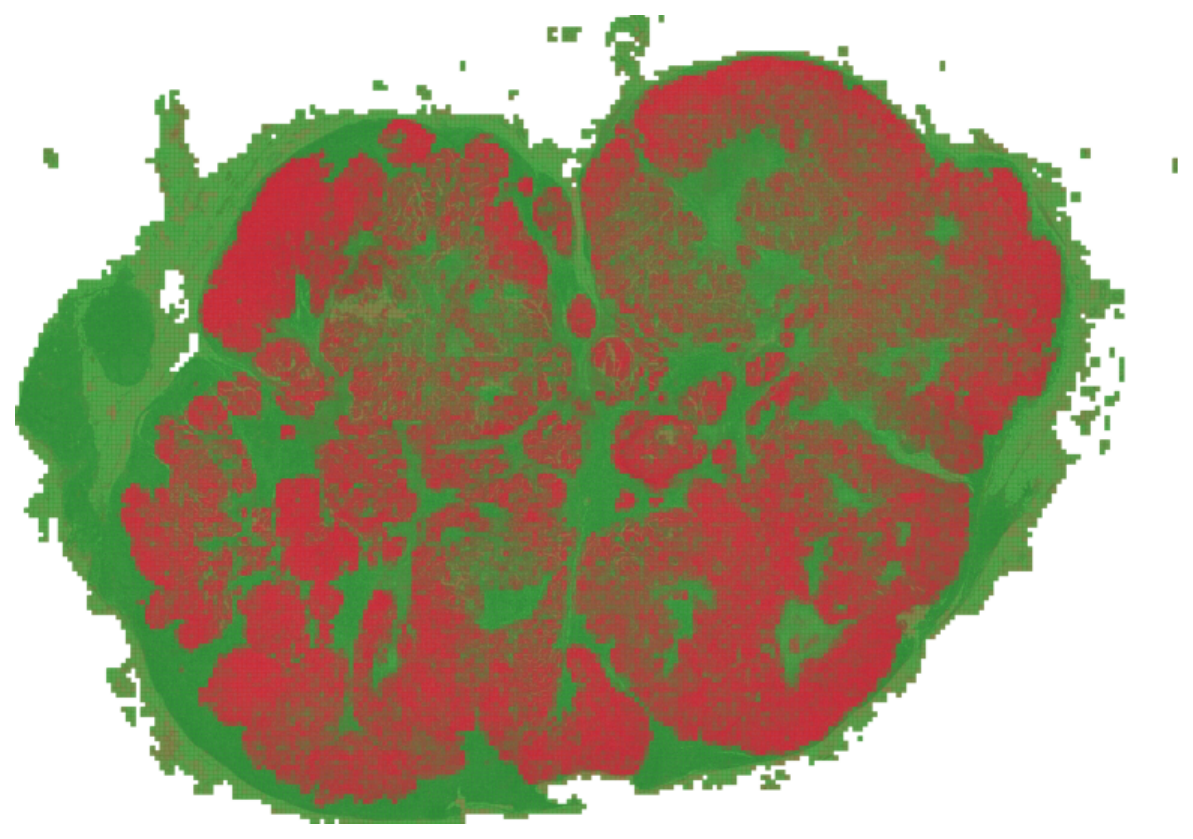}
				& 
				\includegraphics[trim={0.0cm 0cm 0cm 0cm},clip,height=3cm]
				{img/camelyon16/att_maps/att_map-pathgcn-test_105.pdf}
                &
				\includegraphics[trim={0.0cm 0cm 0cm 0cm},clip,height=3cm]{img/camelyon16/att_maps/att_map-dftdmil-test_105.pdf}
                &
                \includegraphics[trim={0.0cm 0cm 0cm 0cm},clip,height=3cm]
				{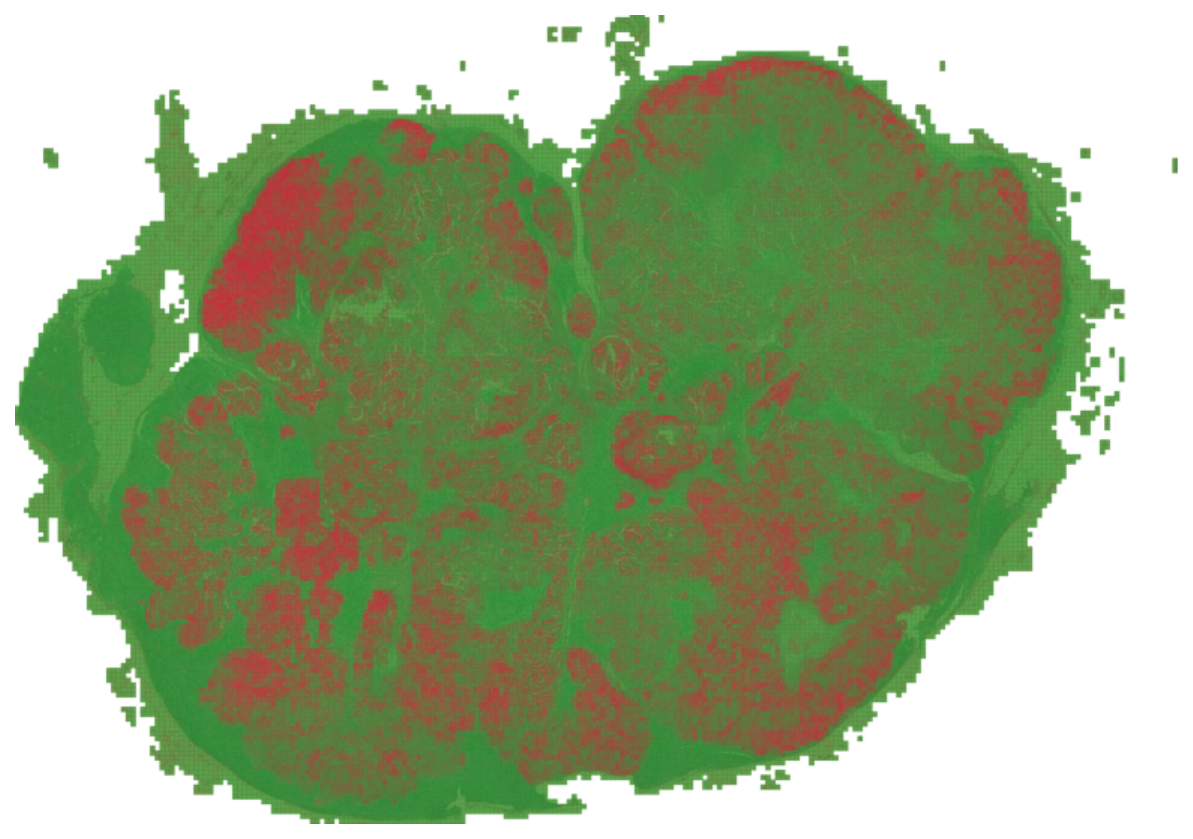}
                &
				\includegraphics[trim={0.0cm 0cm 0cm 0cm},clip,height=3cm]
				{img/camelyon16/att_maps/att_map-camil-test_105.pdf}
				& 
				\includegraphics[trim={0.0cm 0cm 0cm 0cm},clip,height=3cm]
				{img/camelyon16/att_maps/att_map-gtp-test_105.pdf}
				\\        
				\abmil & \pathgcn & \dtfdmil & \transformerabmil & \camil & \gtp \\
			\end{tabular}
		\end{adjustbox}
        \caption{Attention maps in a WSI from CAMELYON16. The attention values have been normalized to ease visualization. PSA stands for \probsmoothatt.}
        \label{fig:attmaps-camelyon-test_105}
        \vspace{-2cm}
    \end{center}
\end{figure}

\begin{figure}[t!]
    \large
	\begin{center}
		\centering
		\begin{adjustbox}{width=0.7\textwidth}
			\begin{tabular}{cc}
				\includegraphics[trim={0cm 0cm 0cm 0cm},clip,width=5cm]{img/att_map-bar_horizontal.pdf} 
				&
				\includegraphics[trim={0cm 0cm 0cm 0cm},clip,width=5cm]{img/uncert_map-bar_horizontal.pdf}
			\end{tabular}
		\end{adjustbox}
		\begin{adjustbox}{width=1.0\textwidth}
			\begin{tabular}{ccccccc}
				\includegraphics[trim={0.0cm 0cm 0cm 0cm},clip,height=3cm]{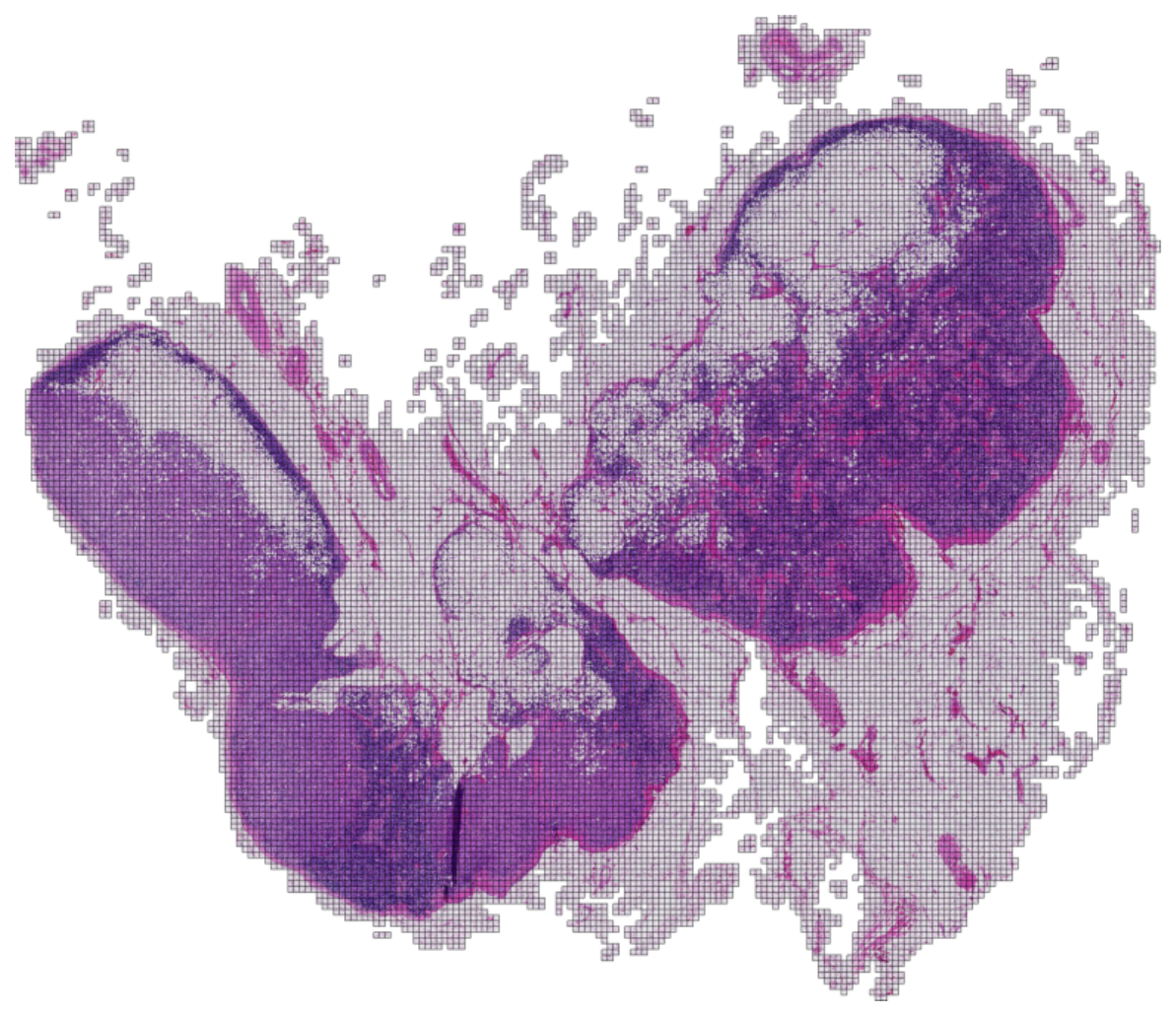}
				& 
				\includegraphics[trim={0.0cm 0cm 0cm 0cm},clip,height=3cm]{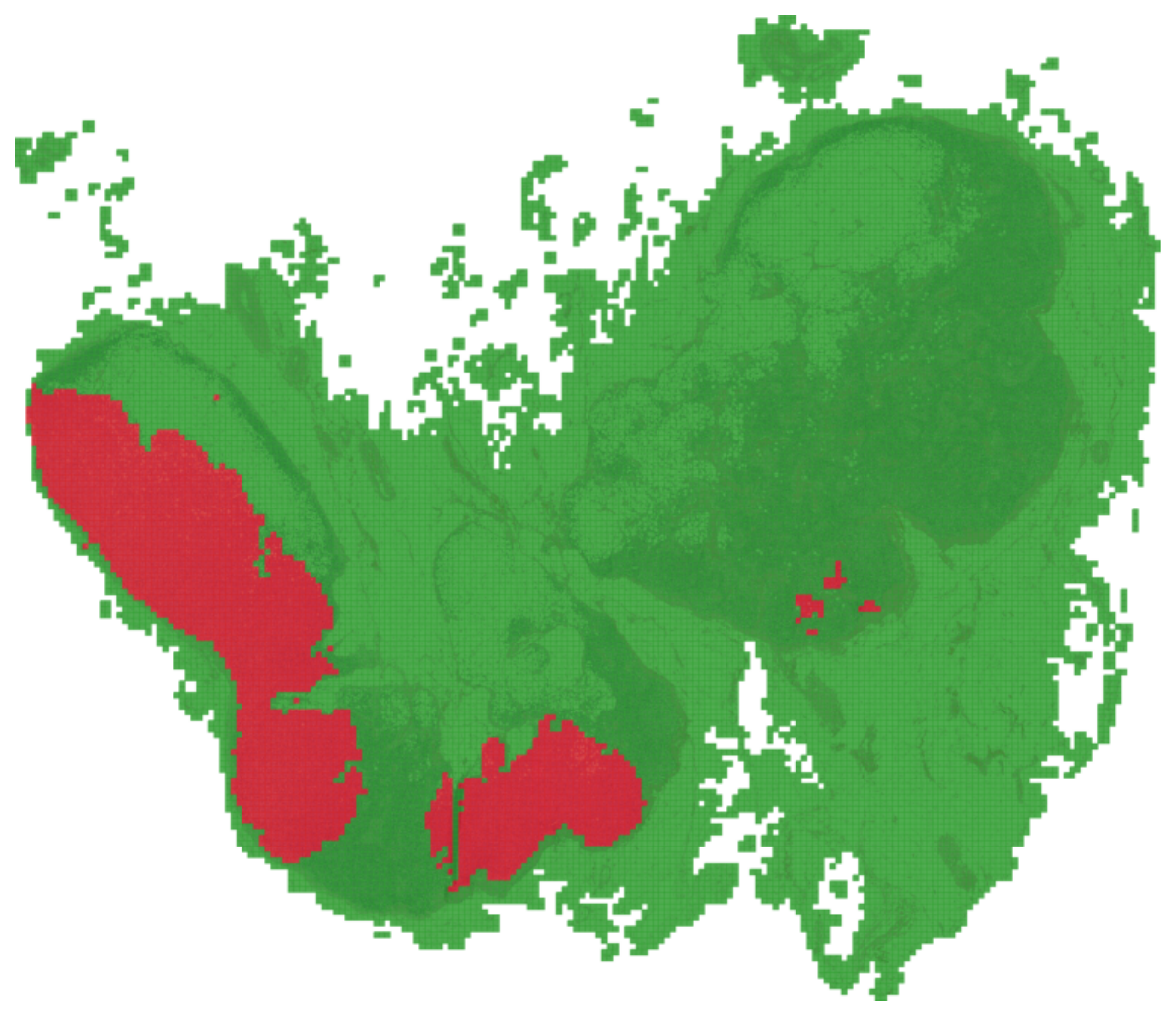}
				& 
				\includegraphics[trim={0.0cm 0cm 0cm 0cm},clip,height=3cm]
				{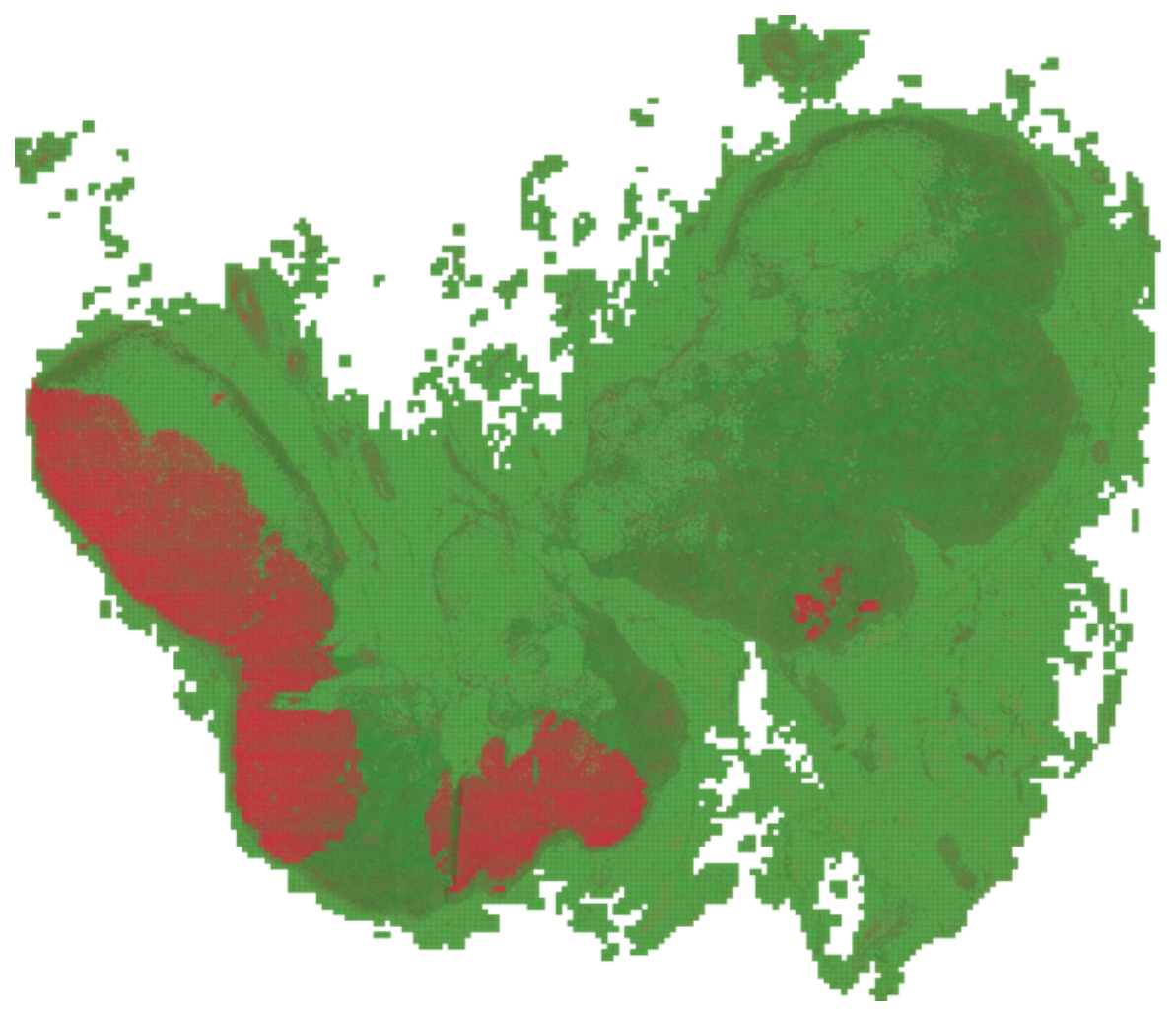}
				& 
				\includegraphics[trim={0.0cm 0cm 0cm 0cm},clip,height=3cm]
				{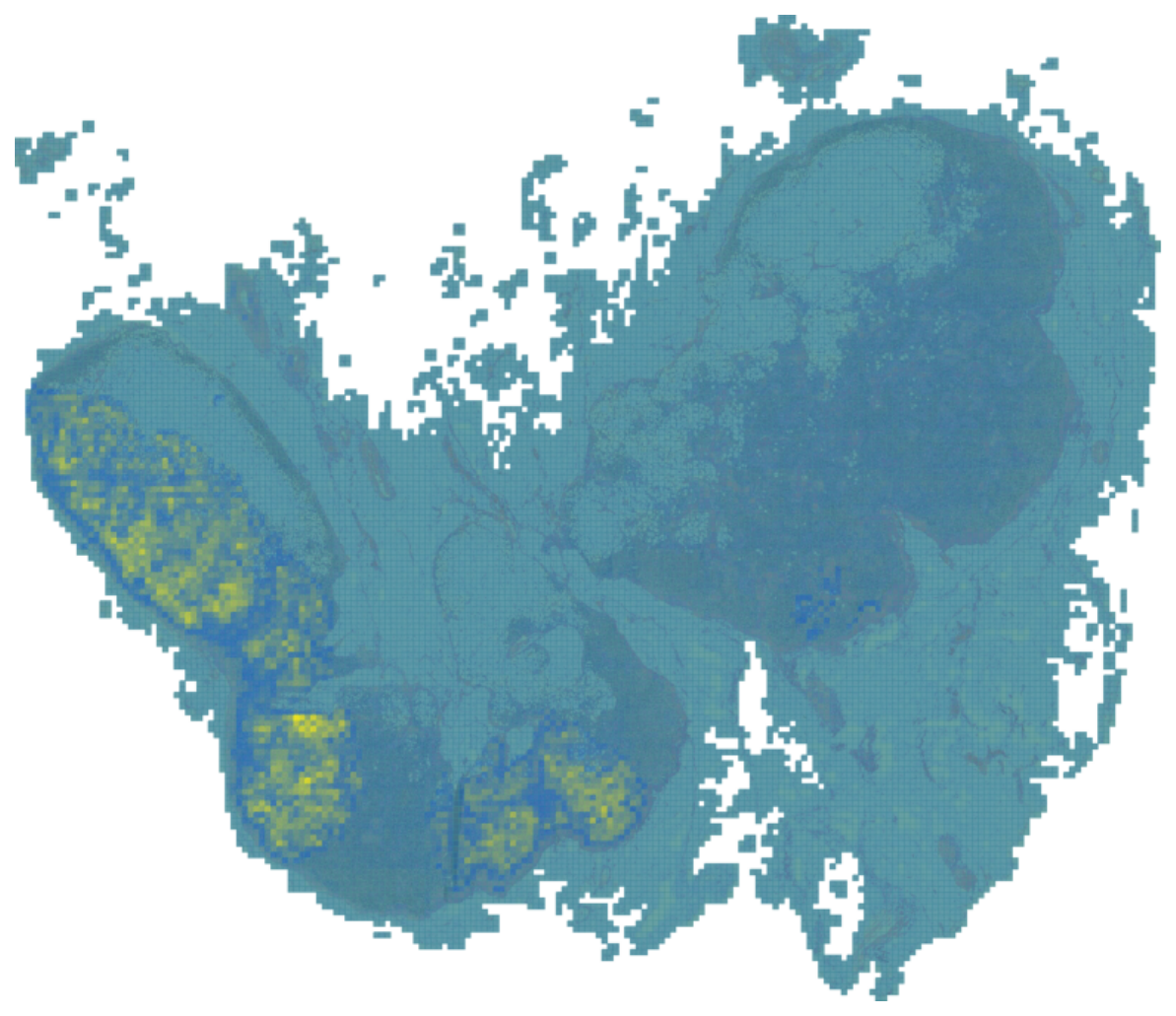}
                &
                \includegraphics[trim={0.0cm 0cm 0cm 0cm},clip,height=3cm]
				{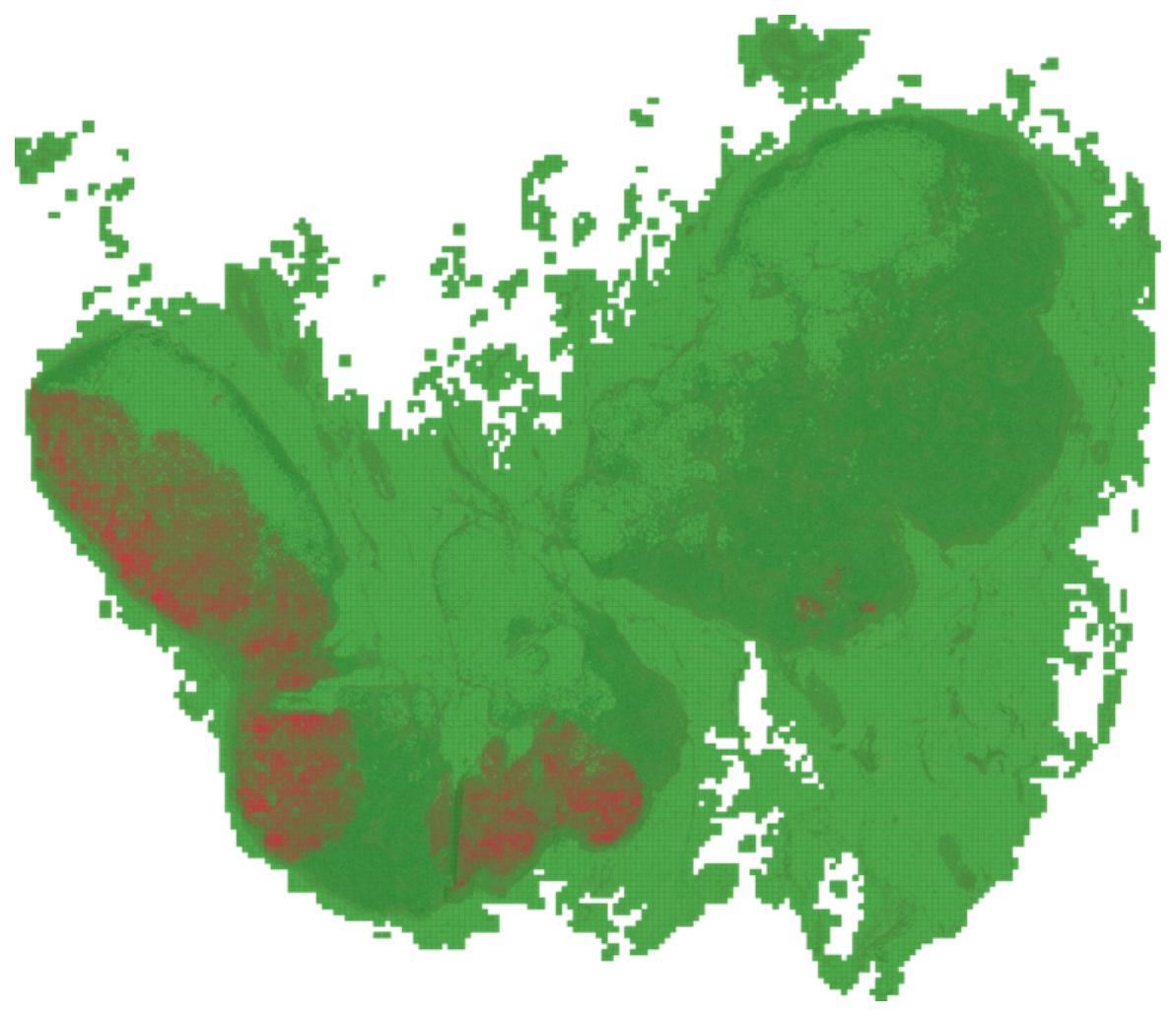}
				& 
				\includegraphics[trim={0.0cm 0cm 0cm 0cm},clip,height=3cm]
				{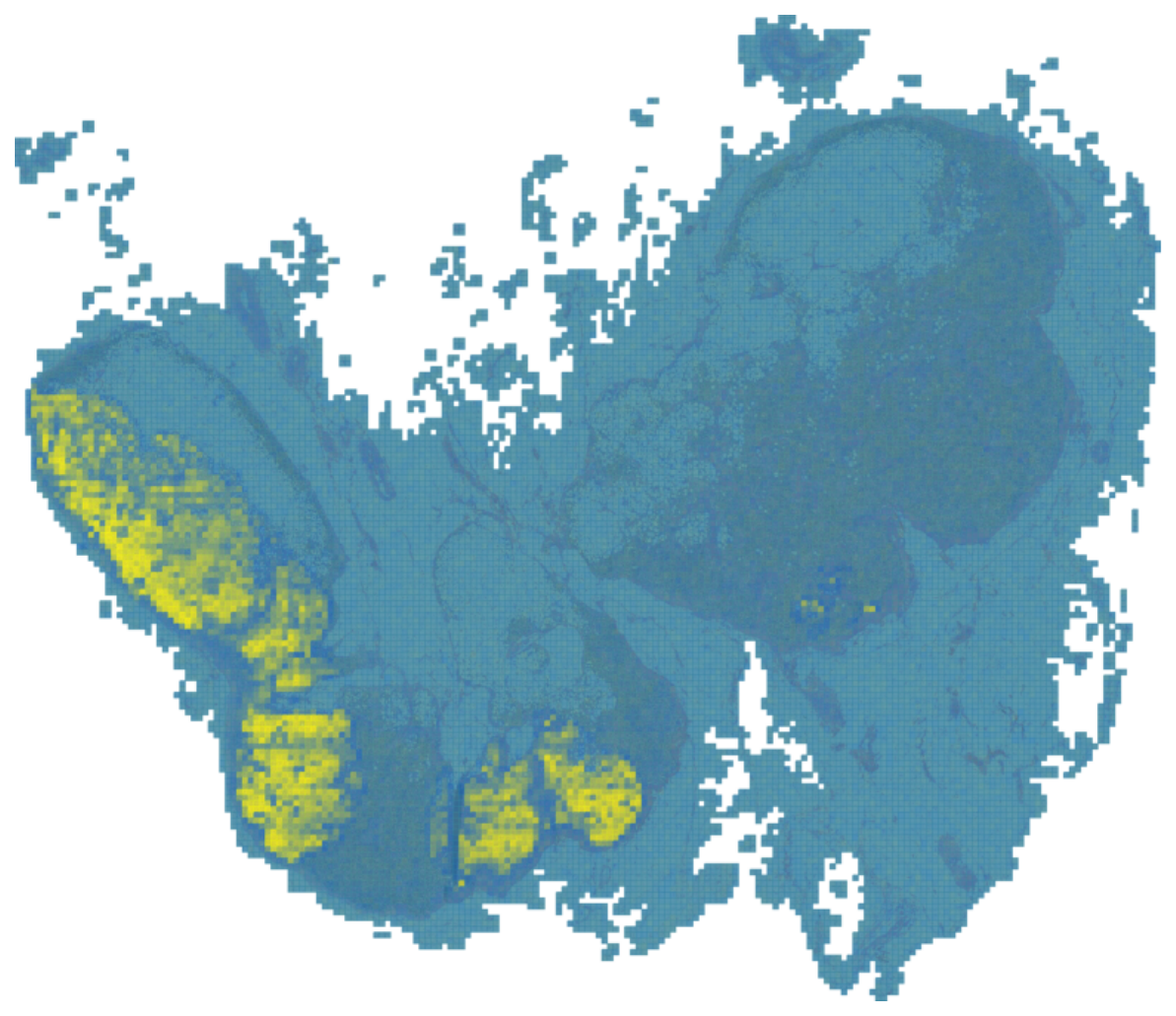}
                \\
                WSI &  Patch labels & \makecell{\abmil+PSA \\ Mean} & \makecell{\abmil+PSA \\ Variance} & \makecell{\transformerabmil+PSA \\ Mean} & \makecell{\transformerabmil+PSA \\ Variance} \\
                \includegraphics[trim={0.0cm 0cm 0cm 0cm},clip,height=3cm]
				{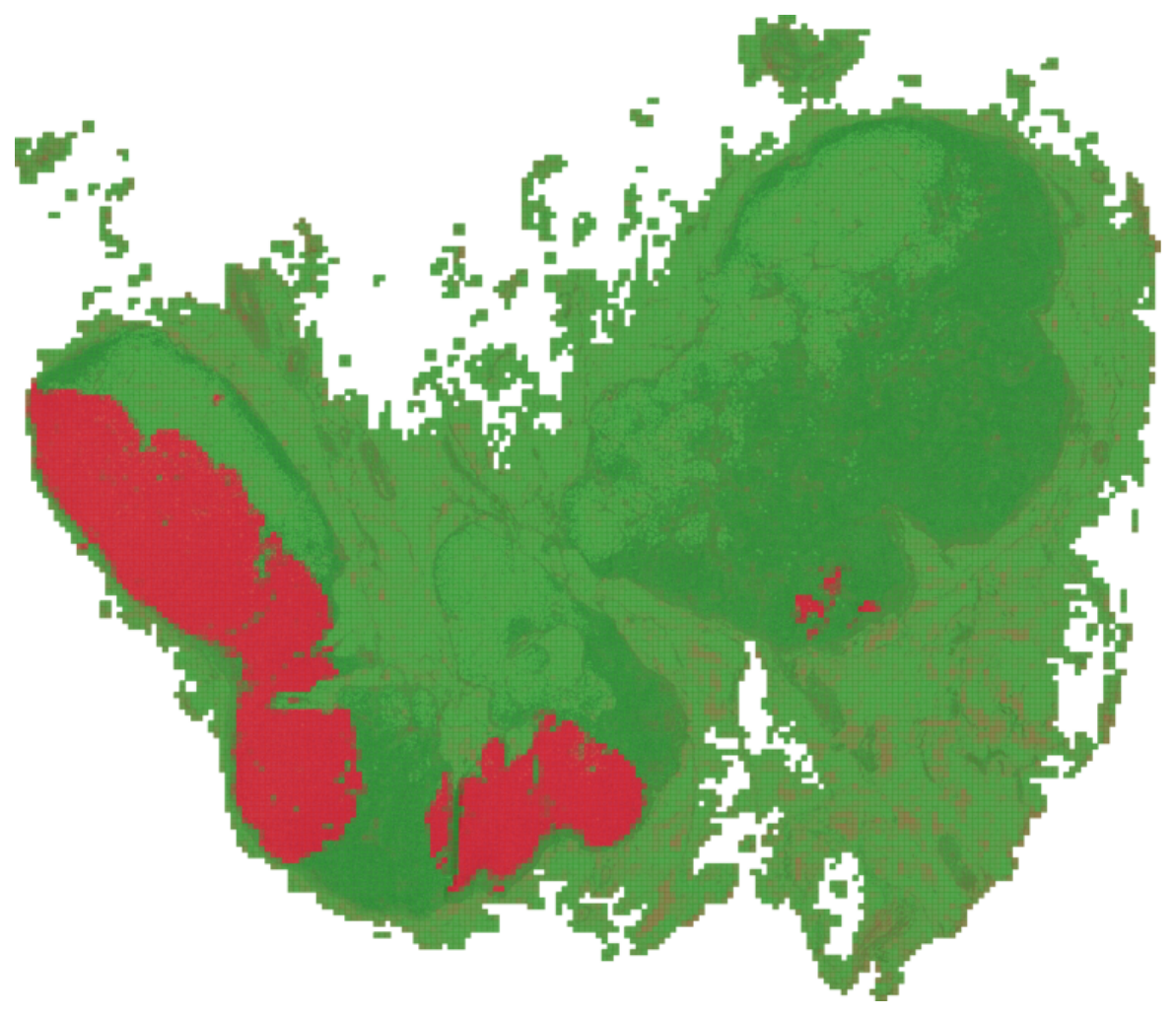}
				& 
				\includegraphics[trim={0.0cm 0cm 0cm 0cm},clip,height=3cm]
				{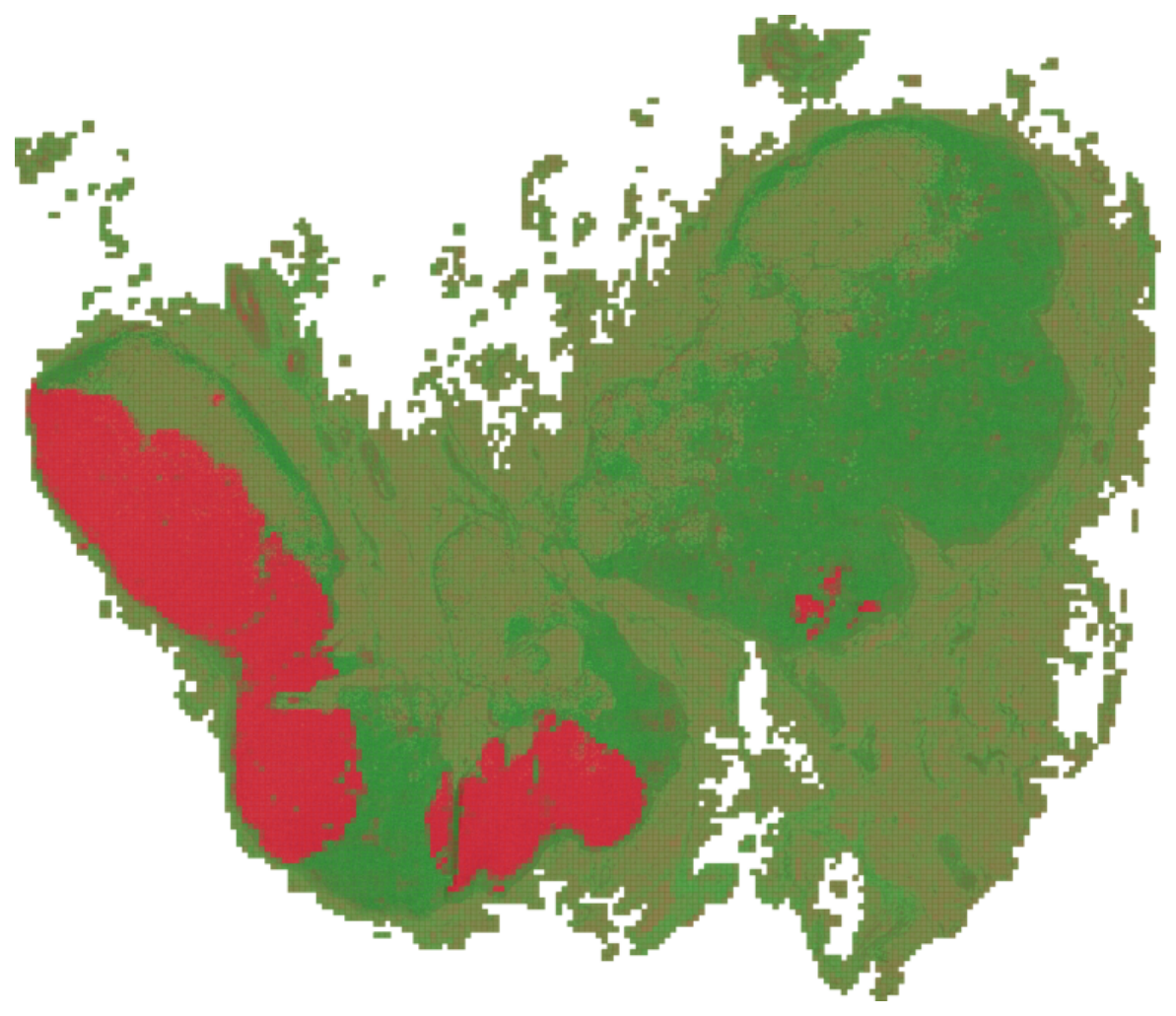}
                &
				\includegraphics[trim={0.0cm 0cm 0cm 0cm},clip,height=3cm]{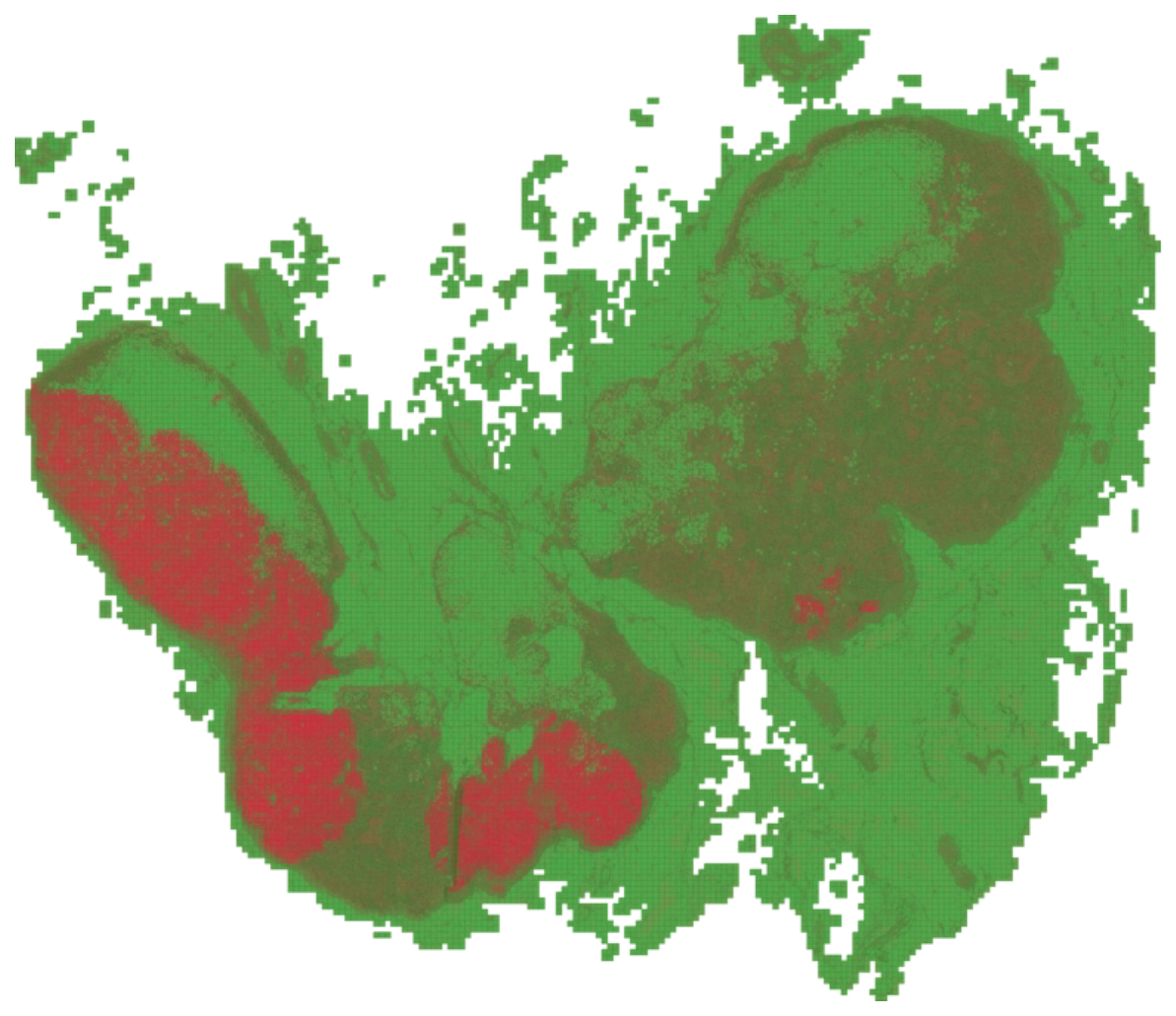}
                &
                \includegraphics[trim={0.0cm 0cm 0cm 0cm},clip,height=3cm]
				{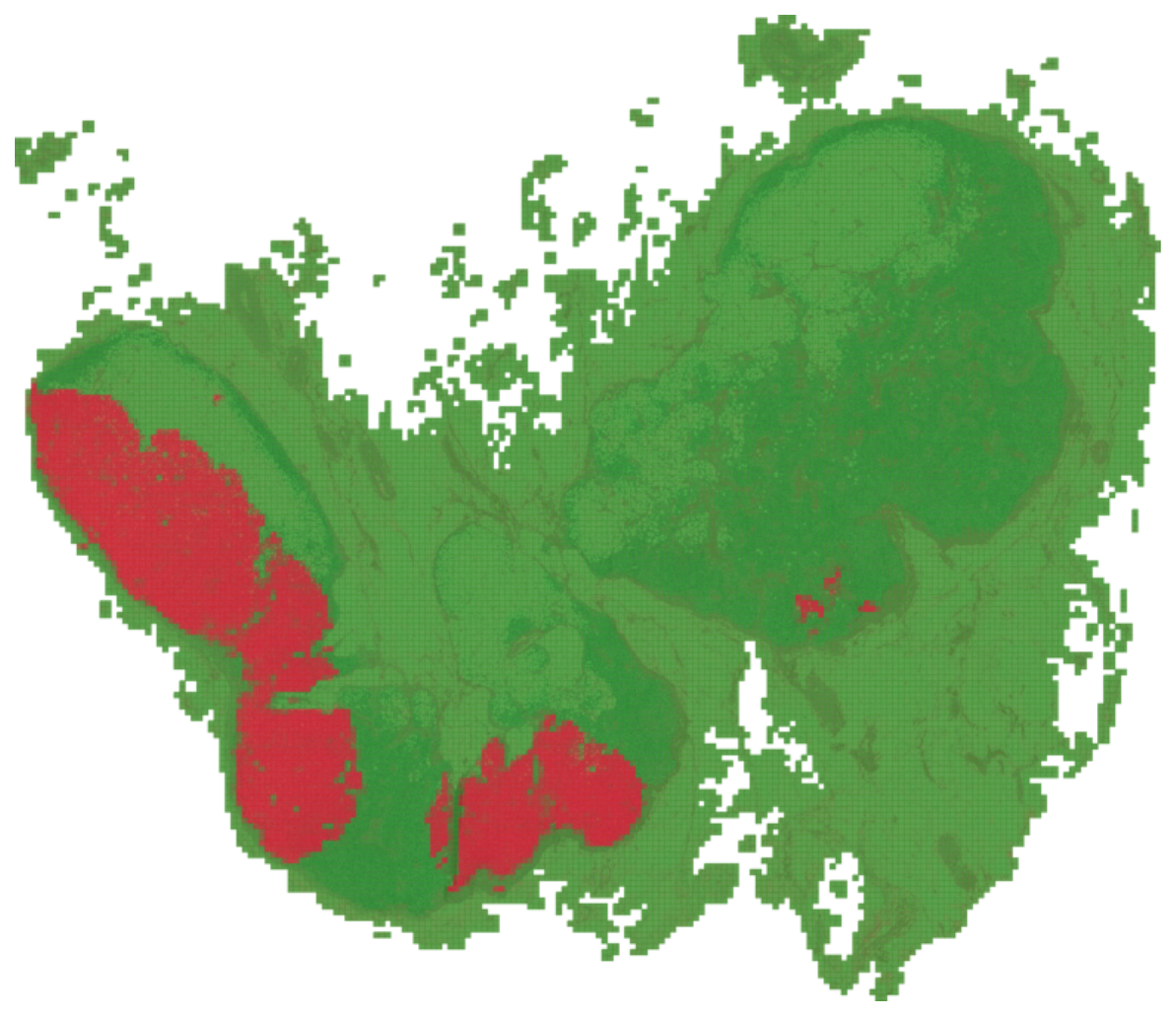}
                &
				\includegraphics[trim={0.0cm 0cm 0cm 0cm},clip,height=3cm]
				{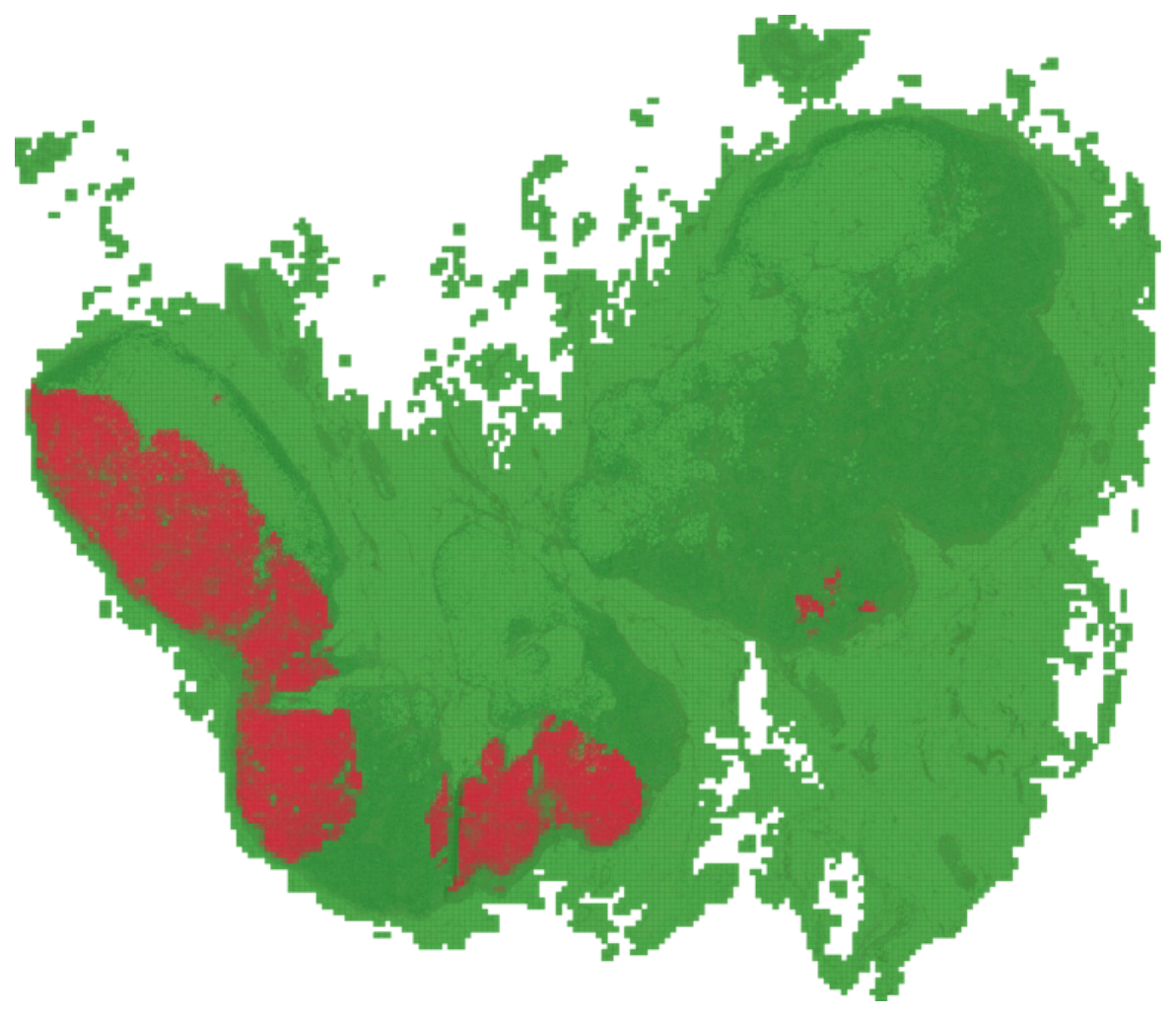}
				& 
				\includegraphics[trim={0.0cm 0cm 0cm 0cm},clip,height=3cm]
				{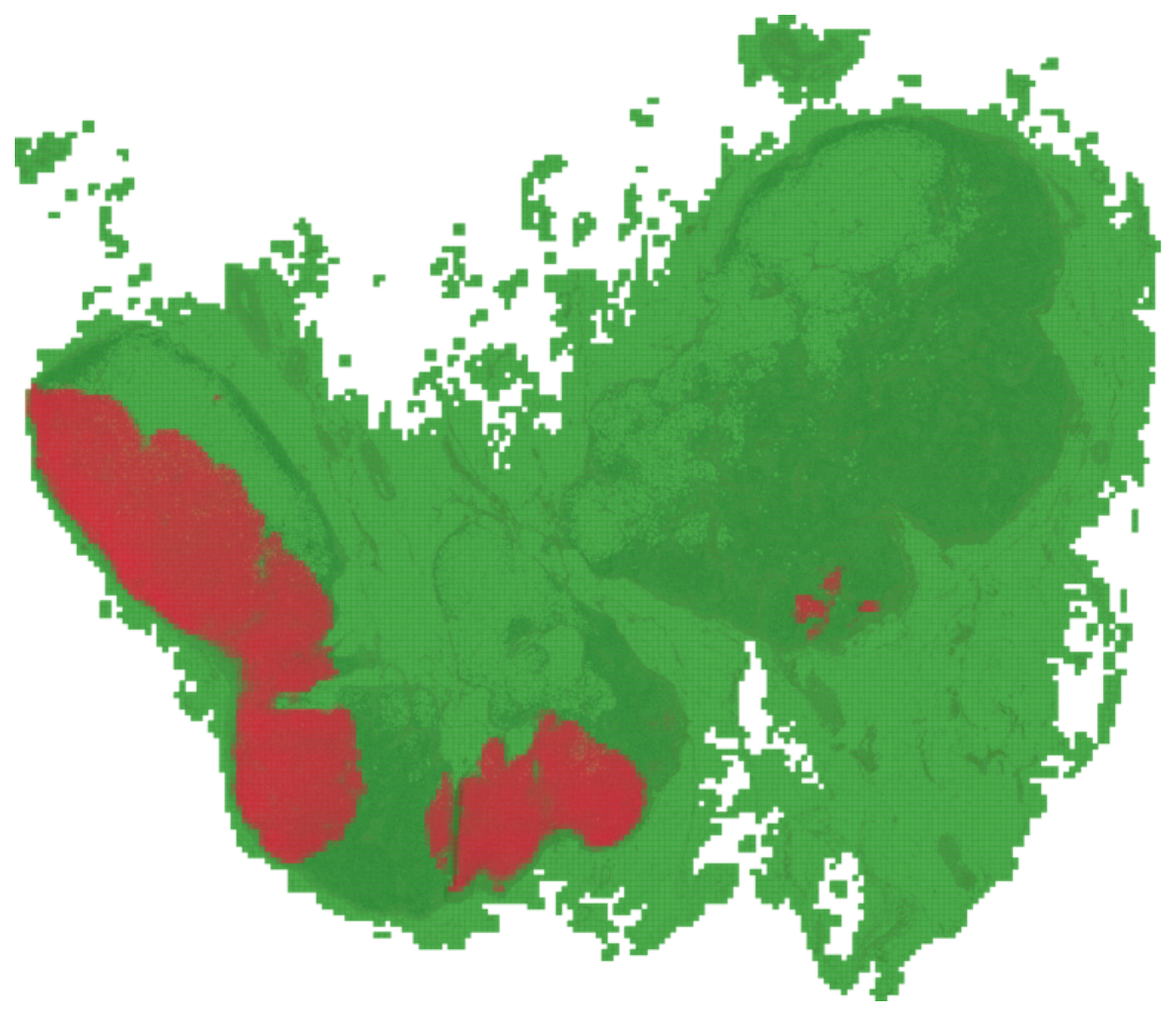}
				\\        
				\abmil & \pathgcn & \dtfdmil & \transformerabmil & \camil & \gtp \\
			\end{tabular}
		\end{adjustbox}
		\caption{Attention maps in a WSI from CAMELYON16. The attention values have been normalized to ease visualization. PSA stands for \probsmoothatt.}
		\label{fig:attmaps-camelyon-test_016}
        \vspace{-2cm}
	\end{center}
\end{figure}

\end{document}